\documentclass[jair,twoside,11pt,theapa]{article}
\usepackage{jair, theapa, rawfonts}

\ShortHeadings{Zone pAth Construction (ZAC) based Approaches for Effective Real-Time Ridesharing}
{Lowalekar, Varakantham, \& Jaillet}
\usepackage{comment}
\usepackage[utf8]{inputenc}
\usepackage{textcomp}
\usepackage{mathtools}
\usepackage{algorithm}
\usepackage{multirow}
\usepackage{epsfig}
\usepackage{epstopdf}
\usepackage{amsmath}
\usepackage{amssymb}
\usepackage{subfig}
\usepackage[noend]{algorithmic}
\DeclarePairedDelimiter\ceil{\lceil}{\rceil}
\DeclarePairedDelimiter\floor{\lfloor}{\rfloor}

\newtheorem{obsv}{Observation}
\newtheorem{defn}{Definition}

\newtheorem{Example}{Example}

\newcommand{\squishlist}{
 \begin{list}{$\bullet$}
  { \setlength{\itemsep}{0pt}
     \setlength{\parsep}{2pt}
     \setlength{\topsep}{2pt}
     \setlength{\partopsep}{0pt}
     \setlength{\leftmargin}{1em}
     \setlength{\labelwidth}{1em}
     \setlength{\labelsep}{0.5em} } }
     \newcommand{\squishend}{
  \end{list}  }

\begin{document}

\title{Zone pAth Construction (ZAC) based Approaches for Effective Real-Time Ridesharing~\thanks{This paper is an extension of our earlier paper at ICAPS 2019~\cite{lowalekar2019zac}. We have extended the paper in the following ways: (a) A \textbf{non-myopic approach} using zone paths for generating assignment of vehicles to request combinations by approximating the future effect of an assignment; (b)\textbf{Benders decomposition} method to efficiently solve the resultant non-myopic optimization formulation (after including the future effect of an assignment) in real-time; (c) A detailed experimental comparison of our non-myopic approach against leading myopic and non-myopic approaches on real-world and synthetic datasets. }}

\author{\name Meghna Lowalekar \email meghnal.2015@phdcs.smu.edu.sg \\
       \name Pradeep Varakantham \email pradeepv@smu.edu.sg \\
       \addr School of Information Systems,\\
	 Singapore Management University, Singapore\\
       \AND
       \name Patrick Jaillet \email jaillet@mit.edu \\
       \addr Department of Electrical Engineering and Computer Science,\\
       Massachusetts Institute of Technology, USA}

% For research notes, remove the comment character in the line below.	
% \researchnote

\maketitle

\begin{abstract}
Real-time ridesharing systems such as UberPool, Lyft Line, GrabShare have become hugely popular as they reduce the costs for customers, improve per trip revenue for drivers and reduce traffic on the roads by grouping customers with similar itineraries. The key challenge in these systems is to group the ``right" requests to travel together in the ``right" available vehicles in real-time, so that the objective (e.g., requests served, revenue or delay) is optimized. This challenge has been addressed in existing work by: (i) generating as many relevant feasible (with respect to the available delay for customers) combinations of requests as possible in real-time; and then (ii) optimizing assignment of the feasible request combinations to vehicles. Since the number of request combinations increases exponentially with the increase in vehicle capacity and number of requests, unfortunately, such approaches have to employ ad hoc heuristics to identify a subset of request combinations for assignment. 

Our key contribution is in developing approaches that employ zone (abstraction of individual locations) paths instead of request combinations. Zone paths allow for generation of significantly more ``relevant" combinations (in comparison to ad hoc heuristics) in real-time than competing approaches due to two reasons: (i) Each zone path can typically represent multiple request combinations; (ii) Zone paths are generated using a combination of offline and online methods. Specifically, we contribute both myopic (ridesharing assignment focussed on current requests only) and non-myopic (ridesharing assignment considers impact on expected future requests) approaches that employ zone paths. In our experimental results, we demonstrate that our myopic approach outperforms (with respect to both objective and runtime) the current best myopic approach for ridesharing on both real-world and synthetic datasets. We also show that our non-myopic approach obtains 14.7\% improvement over existing myopic approach. Our non-myopic approach gets improvements of up to 12.48\% over a recent non-myopic approach, NeurADP. Even when NeurADP is allowed to optimize learning over test settings, results largely remain comparable except in a couple of cases, where NeurADP performs better.

\end{abstract}
\section{Introduction}
\label{sect:intro}
Real-time taxi sharing platforms, such as UberPool, Lyft Line and GrabShare, etc. and on demand shuttle services such as Shotl, Beeline and GrabShuttle, etc. have become hugely popular in recent years due to reduced costs for the customers and improved per trip revenue for drivers. In addition, they also help in reducing the traffic congestion as they allow customers with similar itineraries to share a vehicle. Other shared mobility services, such as car sharing, courier services, scooter sharing, bikesharing, etc. also have a similar underlying problem and the approaches (with minor extensions) presented in this paper can be used for those problems as well.

The ridesharing problem~\cite{alonso2017demand,ma2013t} is related to the vehicle routing~\cite{ritzinger2016survey} and multi-vehicle pick-up and delivery problems~\cite{parragh2008survey,yang2004real}, where customer demand should be picked up from their origin locations and dropped at their destination locations while satisfying vehicle capacity and delay constraints. Earlier work on these problems has focussed on traditional integer programming approaches which are limited to small scale problems of 8 vehicles and 96 requests~\cite{ropke2007models,Ropke2009}. More importantly, these ridesharing problems are typically offline and do not require real-time assignment. 
\begin{figure}[htbp]
 \centering
\subfloat{\label{fig:figrl}\includegraphics[width=0.99\textwidth,height=3.75in]{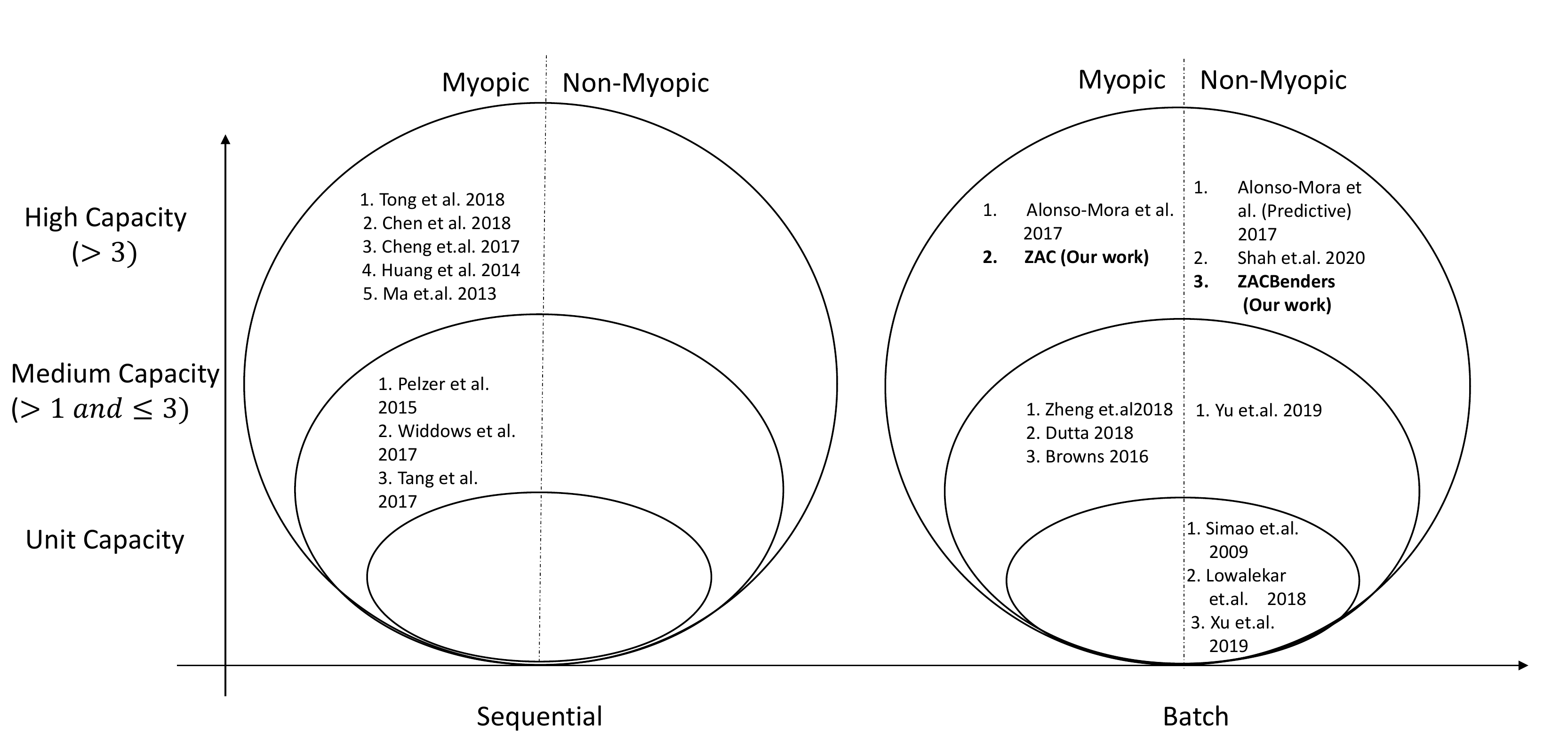}}
 \caption{Related Work}
 \label{fig:relwork}
\end{figure}

In recent times, due to taxi sharing platforms, the focus has shifted to real-time taxi ridesharing problems. Many heuristic approaches have been proposed to solve these problems. As shown in the Figure \ref{fig:relwork}, existing work can be categorized along three dimensions: 
\begin{enumerate}
\item Capacity of vehicles
\item Sequential or batch consideration of requests 
\item Nature of ``vehicle to request combination" assignment (myopic or non-myopic).
\end{enumerate}
Most existing works have considered myopic assignments (third dimension), i.e., they do not consider the future value of assignments while assigning the vehicles to current set of requests. The myopic approaches for ridesharing can further be categorized along the second dimension as:
\squishlist
\item[{\bf Finding a sequential solution:}] Assigns one request at a time to the best available vehicle. Solution approaches are appropriate for high capacity vehicles (shuttles, buses etc.)~\cite{tong2018unified,huang2014large,ma2013t} \textbf{or} 
\item[{\bf Finding a batch solution:}] Assigns all active requests together in a batch to the available vehicles. Most of the proposed approaches are only applicable for low capacity vehicles (e.g., taxis)~\cite{dutta2018hashing,zheng2018order}. 
\squishend
The sequential solution is faster to compute but the quality of solution obtained is typically poor. On the other hand, batch solution takes significantly more time to compute, but the solution quality is significantly better than the incremental solution. 

As opposed to unit capacity taxi matching (first dimension), where a myopic batch solution can be computed by performing a bipartite matching between vehicles and requests, finding a myopic batch solution in multi-capacity ridesharing is challenging. This is because the underlying matching graph in multi-capacity ridesharing changes from bipartite (vehicles and requests) to tripartite (vehicles, requests, request combinations). Similar to Alonso {\em et al.}~\cite{alonso2017demand,alonso2017predictive} and Shah {\em et al.}~\cite{neuradp}, in this paper, we focus on this most challenging category of obtaining a batch solution for high capacity vehicles in real-time. 

The myopic approach by Alonso {\em et al.}~\cite{alonso2017demand} is a generalization of the greedy approach typically employed by taxi companies~\cite{grabshare,grabshare1,lyftline} and is divided into two parts: 
\begin{itemize}
\item The first part constructs an RTV (Request Trip Vehicle) graph. The nodes in the graph are requests, vehicles and trips. A trip in an RTV graph corresponds to a combination of requests that is feasible (with respect to available delay for customers). There is an edge between request and trip if the request is a part of the trip. There is an edge between trip and vehicle if the vehicle can serve all the requests in that trip. 
\item From all allowable allocations of vehicles to trips, the second part computes an optimal allocation of vehicles to trips that minimizes delay or maximizes the number of requests served. 
\end{itemize}
\begin{figure*}[htbp]
 \centering
\subfloat[]{\label{fig:figtbf}\includegraphics[width=0.5\textwidth,height=2.0in]{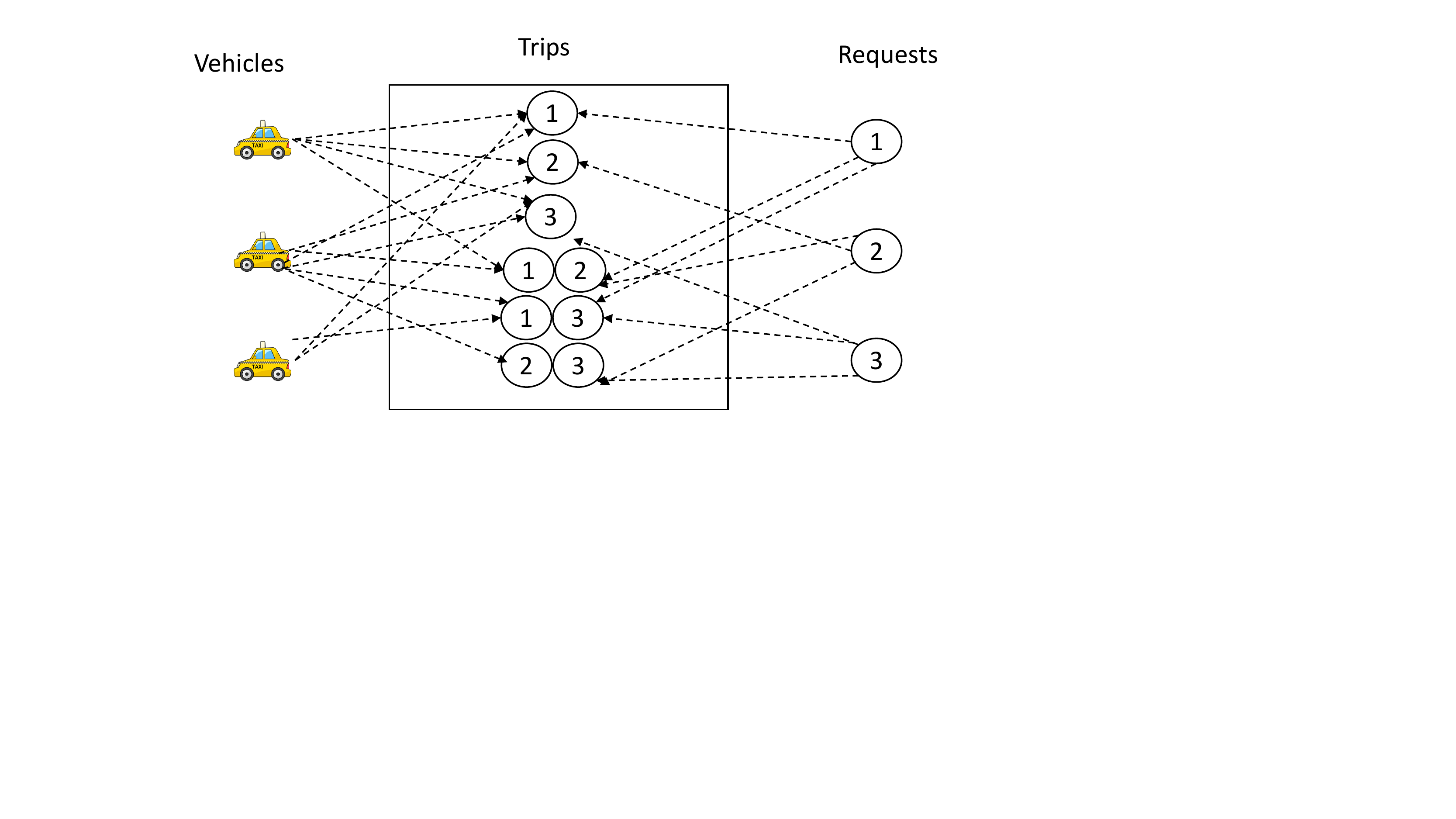}}
\subfloat[]{\label{fig:figzac}\includegraphics[width=0.5\textwidth,height=2.0in]{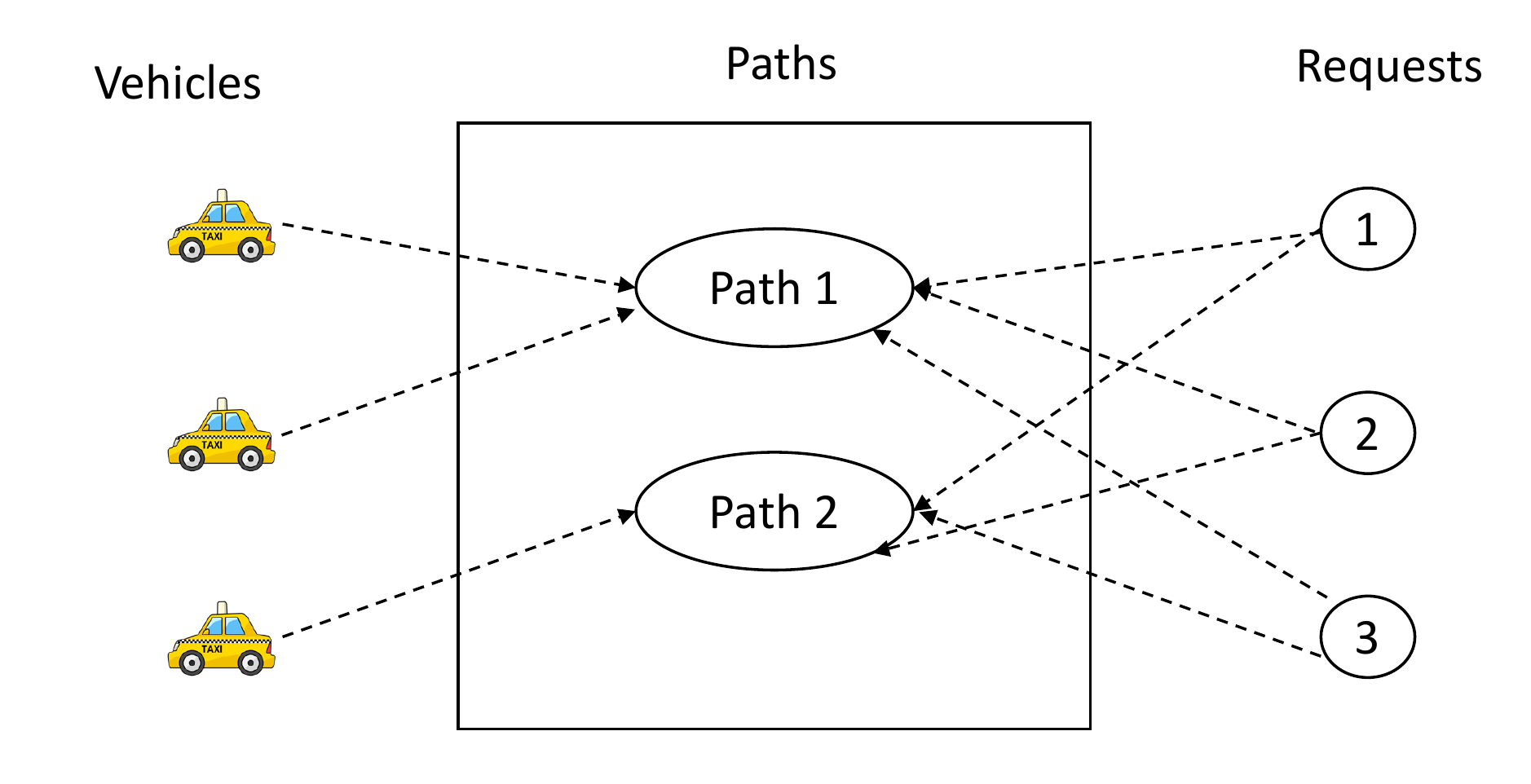}}
 \caption{(a) Representation of RTV graph generated by the model in ~\protect\cite{alonso2017demand} for capacity 2. (b) Representation of RPV graph generated by ZAC approach.}
 \label{fig:tbf}
\end{figure*}
This approach is limited in scalability as the set of possible trips increases exponentially with the increase in the number of requests and capacity of vehicles. To ensure scalability and address the key challenge of identifying as many relevant trips as possible in real-time, this approach employs ad hoc heuristics (e.g., limiting time available and edges in RTV graph). 

The non-myopic approach by Alonso {\em et al.}~\cite{alonso2017predictive}) is a minor extension of their myopic approach and does not provide any improvement in the number of requests served (more details in Section \ref{sect:related}). The non-myopic approach by Shah {\em et al.}~\cite{neuradp} uses similar approach as Alonso {\em et al.}~\cite{alonso2017demand} to generate the feasible trips and then uses a neural network based value function approximation to estimate the future effect of current assignment of vehicle to trips. While the approach provides better results than myopic approaches, it has two main issues: (1) This approach still has to employ ad hoc heuristics to identify the relevant request combinations considered for learning; and (2) Due to the need of training a separate network model for each dataset and each change of input parameter, it is not easily adaptable to different settings. 
%We highlight this issue using experiments on real-world datasets. 

To that end, we make three major contributions in this paper. 
\textbf{First}, we propose a framework called ZAC~({\textbf Z}one p{\textbf A}th {\textbf C}onstruction), that is myopic and employs two crucial ideas to identify significantly more relevant trips in real-time:
\squishlist
\item \textbf{\em Focus on zone paths instead of trips:} A zone path is a path that connects zones (a zone is an abstraction for multiple individual locations) and therefore it can group multiple trips that have ``nearby" or ``on the way" pick-ups and drop-offs. This focus on zone paths helps automatically capture multiple {\em relevant} trips with one zone path. 
\item \textbf{\em Offline-online computation of zone paths:} Since, we focus on zone paths, we can generate partial zone paths offline. This helps capture more relevant trips online in real-time, where the partial paths are completed. 
\squishend

Instead of an RTV (Request Trip Vehicle) graph in Alonso {\em et al.}'s~\cite{alonso2017demand} approach, we construct an RPV (Request Path Vehicle) graph, where we associate requests and vehicles to zone paths. This is shown in Figure~\ref{fig:tbf}. Once the RPV is constructed, we then employ a scalable integer linear program to find the optimal assignment (e.g., maximize revenue, maximize the number of requests served or minimize the delay) of vehicles and requests to paths.

\textbf{Second}, we provide a non-myopic extension of ZAC, called ZACBenders, which approximates the expected future value of assignments by considering multiple samples of future demand. By grouping the requests in demand samples based on zones and approximating how future requests are served (details in the Section \ref{sect:zacbenders}), the integer optimization in ZAC is modified to optimize the sum of values of current assignment and expected future value over multiple demand samples. However, this results in an increase in the complexity of optimization formulation. Therefore, we propose using Benders decomposition~\cite{Benders} to break the large optimization formulation into multiple smaller problems that are solved in parallel.

\textbf{Finally}, we provide an exhaustive evaluation of our contributions in comparison to two leading approaches for real-time ridesharing, Trip based Formulation (TBF)~\cite{alonso2017demand} and NeurADP~\cite{neuradp}, on both real-world and synthetic datasets. We get up to 14.7\% improvement over TBF and up to 12.48\% over NeurADP. Even when NeurADP is allowed to optimize learning over test settings, results largely remain comparable except in a couple of cases, where NeurADP performs better. We also perform experiments on a synthetic dataset introduced by Bertsimas {\em et al.}~\cite{bertsimas2018online}. We simulate the first and last mile transportation requests in this dataset and show that in these settings we can obtain a staggering 20\% gain over TBF.

%
%We first compare our myopic approach, ZAC, against the Alonso {\em et al.}'s~\cite{alonso2017demand} approach, referred to as TBF, on synthetic and two real world datasets and show that our approach not only is very efficient (with respect to runtime) but also serves more requests. On real datasets, as shown in the experimental results, ZAC obtains upto 4\% gain over TBF. 
%
%There is also a recent increase in the popularity of on demand shuttle services~\cite{shotl,beeline,grabshuttle}. These shuttles have fixed pick-up/drop-off points which is a very small subset of complete road network of a city. Therefore, we perform experiments on a synthetic dataset introduced by Bertsimas {\em et al.}~\cite{bertsimas2018online}, where the first mile and last mile transportation requests are simulated. In these cases, ZAC obtains a staggering 20\% gain over TBF, further supporting our claim that ZAC is suitable for real-time ridesharing with higher capacities.
%%\section{Related Work}
%
%We then compare our non-myopic approach, ZACBenders, with the myopic approaches across all datasets and different parameters and show that ZACBenders obtains 14.7\% improvement over TBF. Finally, we also compare ZACBenders with NeurADP~\cite{neuradp} on the New York Yellow Taxi Dataset and show that ZACBenders serves upto 4.5\% more requests than NeurADP. Our non-myopic approach not only outperforms all the existing approaches but is also easily adaptable to different settings. 
\begin{figure}[t]
 \centering
{\includegraphics[width=0.95\textwidth,height=3.5in]{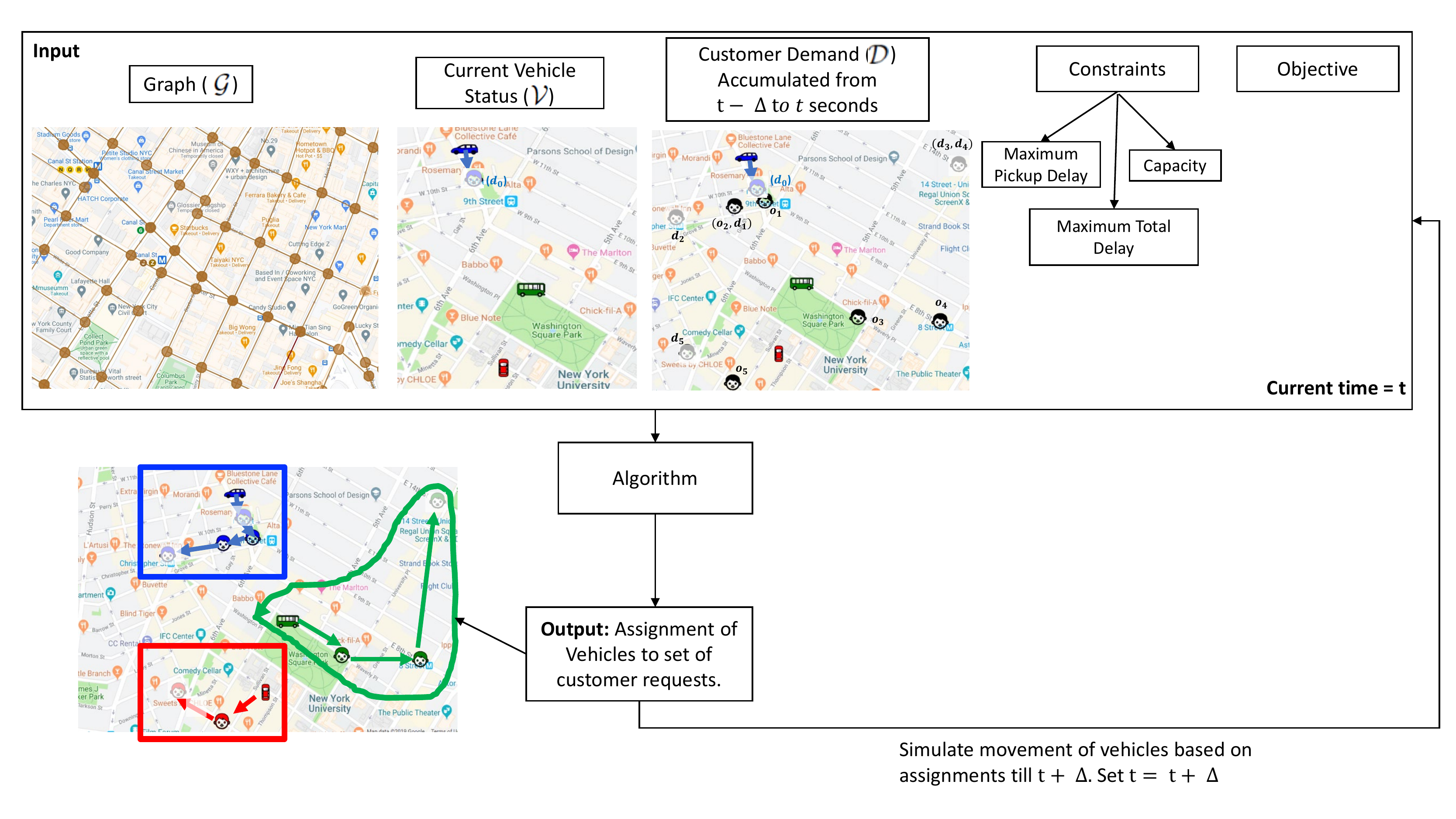}}
 \caption{RMP Description: Graph (${\cal G}$) in this example is the road network of Manhattan (New York) with nodes as the road intersections. ${\cal V}$ provides the current status of vehicles, i.e., their location and currently assigned passengers. In the example, blue vehicle is carrying a customer which needs to be dropped at the location represented using $d_{0}$. Customer requests are accumulated over $\Delta$ duration. The example shows 5 customer requests. The origin location of customer $i$ is represented by $o_{i}$ and destination is represented by $d_{i}$. We use transparent person image to highlight the destination location. The algorithm should output the assignment of vehicles to customer requests while satisfying the constraints and optimizing the objective.}
 \label{fig:rmp}
\end{figure}
\section{Rideshare Matching Problem (RMP)}
\label{sect:existride}
Real-time ridesharing is a service provided by platforms such as Uber, Lyft etc. for arranging shared rides for multiple customers at a very short notice. Customers request for a shared ride from a source to destination. The platform then groups all those requests that can share a ride -- based on whether the delay\footnote{Time taken to reach the destination using the shared ride minus the time taken using an individual ride} of reaching the destination is less than a given threshold for all requests sharing the ride. These services are popular because customers are ready to accept a delay in exchange for a reduced fare. However, after an acceptable threshold, delay can not be compensated with monetary benefits. Therefore, when making groups of customer requests, platforms need to ensure that matching algorithms find groupings which do not increase the delay of individual customers beyond an acceptable threshold~\footnote{Threshold values can be potentially learnt by surveying different customers.}.

%Platforms run a matching algorithm at regular time intervals that groups customer requests and assigns them to vehicles. Platforms and drivers gain (due to higher revenue and lower per customer cost) by serving multiple customers in a single vehicle, customers benefit by paying lesser amount as compared to hiring an individual vehicle. These services are popular because customers are ready to accept a delay in exchange for a reduced fare. However, after an acceptable threshold, delay cannot be compensated with monetary benefits. Therefore, when making groups of customer requests, platforms need to ensure that matching algorithms find groupings which do not increase the delay of individual customers beyond an acceptable threshold\footnote{Threshold values can be potentially learnt by surveying different customers.}. 

%

Figure \ref{fig:rmp} describes the RMP problem. Formally, we define RMP using the following tuple:
$$<{\cal G},\Delta,{\cal D},{\cal V},{\cal C},\tau,\lambda,\kappa,\rho,\xi^{D}>$$

\squishlist
\item ${\cal G}=({\cal L},{\cal E})$ is a graph with the vertices as the set of locations. For e.g., as considered in previous works~\cite{alonso2017demand} the graph ${\cal G}$ is the road network with the set ${\cal L}$ including all street intersections in the road network of a city and ${\cal E}$ denoting the set of road segments. Figure \ref{fig:rmp} shows the visual representation of a part of the graph ${\cal G}$ for the Manhattan city. We assume that vehicles only pick-up and drop people off at intersections. Travel Time (${\cal T}$) and shortest paths (${\cal S}_{p}$) between all location pairs in set ${\cal L}$ is pre-computed and stored.
\item $\Delta$ denotes the decision epoch duration in seconds, i.e., the algorithm is executed every $\Delta$ seconds. In Figure \ref{fig:rmp}, we show that the customer demand is collected over $\Delta$ seconds.
\item ${\cal D}$ denotes the set of current customer requests. Each element $j \in {\cal D}$ is represented using the tuple: $\left<o_{j},d_{j},a_{j}\right>$, where $o_{j}, d_{j} \in {\cal L}$ denote the origin and destination location and $a_{j}$ denotes the arrival time of the request $j$. If the current decision epoch is $e$, then $(e -1 ) \cdot \Delta < a_{j} \leq e \cdot \Delta, \forall j$. Figure \ref{fig:rmp} shows the visual representation of customer demand by taking 5 customer requests as an example. 
\item ${\cal V}$ denotes the set of vehicles. Each element $i \in {\cal V}$ is represented using the tuple: $\left<\mu_{i},\omega_i, q_i\right>$. $\mu_{i} \in {\cal L}$ denotes the initial location of vehicle $i$, $\omega_i$ denotes the time at which vehicle first becomes available at $\mu_i$ and $q_{i}$ denotes the set of customer requests assigned to vehicle $i$. Each element $j$ of $q_i$ is represented using the tuple: $\left<o_{j},d_{j},a_{j}\right>$, where $o_{j}, d_{j},a_{j}$ are as described in the demand tuple above. Figure \ref{fig:rmp} shows the visual representation of set ${\cal V}$ by using three vehicles at their initial location. The blue vehicle has a previously assigned customer request. 

%$o_{j}, d_{j} \in {\cal L}$ denote the origin and destination locations of the customer request, $a_{j}$ denotes the arrival time of the request $j \in q_{i}$ and $tr_{j}$ denotes the time taken to travel directly from $o_{j}$ to $d_{j}$.
\item ${\cal C}$ represents the objective (e.g. revenue, number of requests served, etc.), with ${\cal C}^{t}_{ij}$ denoting the value obtained on assigning request $j$ to vehicle $i$ at decision epoch $t$.
\item $\tau$ denotes maximum allowed wait time for a request (in seconds). The wait time is defined as the difference between the arrival time of a requests and the time at which vehicle picks the customer from its origin.
\item $\lambda$ denotes maximum allowed travel delay for requests (in seconds). The travel delay is the total delay experienced by the customer to reach its destination location. If $t^{d}_{j}$ denotes the time at which a request $j$ is dropped at its destination then the travel delay is given by $t^{d}_{j}-(a_{j}+{\cal T}(o_{j},d_{j}))$.
\item $\kappa$ denotes the maximum capacity of each vehicle.
\squishend

The goal in RMP is to assign the incoming customer requests to the vehicles such that the objective is maximized while satisfying the capacity constraints, maximum wait time and delay constraints. Figure \ref{fig:rmp} shows the assignment of customer requests to different vehicles. 

The last two elements in the tuple ($\rho,\xi^{D}$) are used to represent the expected future information. The myopic approaches ignore these elements, but non-myopic approaches can use these to improve the quality of matching.
\squishlist 
\item $\rho$ denotes the look ahead duration, i.e., the duration over which the expected future value will be computed. 
\item ${\xi}^{D}$ denotes the set of samples of future customer demand with ${\xi}^{D,k}$ denoting the set of future requests in sample $k$. Each element $j' \in {\xi}^{D,k}$ is represented using the tuple: $\left<o_{j'}^{k},d_{j'}^{k},a_{j'}^{k}\right>$, where $o_{j'}^{k}, d_{j'}^{k} \in {\cal L}$ denote the origin and destination location and $a_{j'}^{k}$ denotes the arrival time of the request $j'$. If the current decision epoch is $e$, then $e \cdot \Delta < a_{j'} \leq e \cdot \Delta + \rho, \forall j'$. 
\squishend

We first present our myopic approach ZAC in Section \ref{sect:zac} and then in Section \ref{sect:zacbenders} we provide our non-myopic approach ZACBenders, which can assign incoming customer requests to the vehicles while considering future information.
% The objective in RMP is generally to maximize the number of requests satisfied or to maximize revenue or minimize the delay. 

\section{ZAC: A Zone pAth Construction Approach for solving RMP}
\label{sect:zac}
Given the importance of zone path to ZAC, we first define and explain about zone and zone path. We then describe the intuitive advantages of using zone paths and then we explain the ZAC algorithm. 

\begin{defn}
\textbf{Zone}: refers to an abstracted location obtained by clustering locations in set ${\cal L}$.
\end{defn}
In this work, we investigated Grid Based Clustering (GBC), Hierarchical Agglomerative Clustering with Complete Linkage (HAC\_MAX) and Hierarchical Agglomerative Clustering with Mean Linkage (HAC\_AVG) to cluster locations into zones. We use these methods as they do not require prior knowledge about the number of clusters and have been used in earlier works on similar problems~\cite{ma2013t,hasan2018community}. Please refer to appendix~\ref{appendix:zonecreation} for more details on zone creation.

\begin{defn}
\textbf{Zone path}: refers to an ordered sequence of nodes, where each node corresponds to either a location from set ${\cal L}$ or a zone. 
\end{defn}

\noindent There are two key advantages to a zone path: 
\squishlist
\item Zone path represents multiple trips that have ``nearby" or ``on the way" pick-ups and drop-offs; and 
\item Zone path can be generated at different levels of granularity (e.g., individual locations, communities) depending on the time available. 
\squishend 
Due to these two advantages, zone paths assist in identifying more relevant trips (combinations of requests) within a given amount of runtime. We further enhance the ability to identify more relevant trips within limited runtime, by generating zone paths partially offline and completing them in real-time depending on the set of active requests. 

We generate the zone path of time span $\tau$ offline and complete the rest of the zone path online. This is because requests can be picked up only in initial $\tau$ seconds~\footnote{$\tau$ is typically 300 and we experiment with values between 120-420 seconds.}. Therefore, partial zone paths generated offline automatically provide a pick-up order for the active requests. As a result, online, we only need to compute drop-off order while ensuring that the delay constraints are not violated. This contrasts with Alonso {\em et al.}'s~\cite{alonso2017demand} approach, where both pick-up and drop-off order along with the delay feasibility have to be computed online.\\

%Partial zone paths generated offline automatically provide a pick-up order for active requests and therefore online, we only need to compute drop-off order while ensuring delay constraints are not violated. This is in contrast to Alonso {\em et al.}'s approach, where both pick-up and drop-off order along with delay feasibility have to be computed online.  \\

\noindent {\em 
Intuitively, the inherent nature of zone paths to capture multiple relevant trips coupled with the extra time made available online due to offline computation of partial zone paths enables ZAC to consider significantly more relevant trips in real-time. }\\

Due to abstraction of locations into zones, the travel time is approximately represented when considering zone paths. This can result in longer wait times or longer estimate of wait times than a path over locations in set ${\cal L}$. Customers prefer to have a shorter wait time pre-process~\cite{dube1989consumers,maister1984psychology}, i.e., before pick-up in this case. Therefore, it is essential to reduce this approximation in travel time computation during pick-up. We reduce this approximation by generating offline partial paths at the level of locations. This is another benefit of having an offline partial path. 

ZAC is an offline-online approach for solving the RMP every few seconds on active requests and available vehicles by using offline generated partial paths. The key components of the ZAC algorithm are as follows:
\squishlist
\item Offline: generation of all partial location paths of time span, $\tau$ from every location. 
\item Online: generation of RPV graph by loading and processing offline partial paths, completing the partial paths and identifying edges in RPV graph.
\item Online: finding optimal assignment of requests to paths to vehicles by using an efficient integer (0/1) linear optimization 
\squishend
 \begin{figure}[htbp]
 \centering
{\label{fig:figsp}\includegraphics[width=0.75\textwidth,height=2.25in]{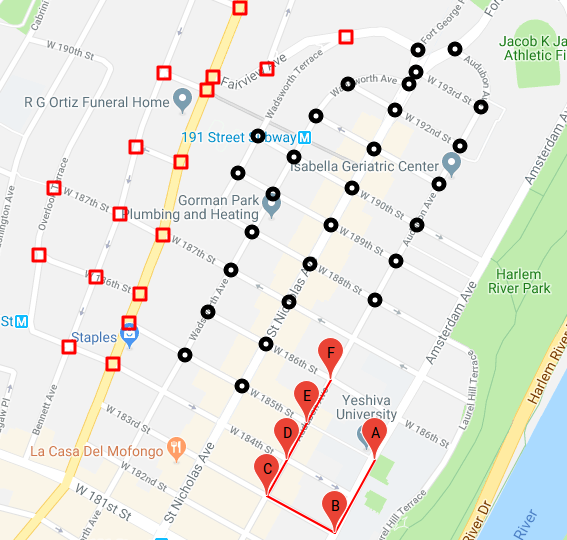}}
%{\label{fig:figsp1}\includegraphics[width=0.75\textwidth,height=1.0in]{images/onlytext.png}}
 \caption{Example Zone Based Path. Path $A \rightarrow B \rightarrow C \rightarrow D \rightarrow E \rightarrow F \rightarrow $Black Zone $\rightarrow$ Red Zone. The part $A \rightarrow B \rightarrow C \rightarrow D \rightarrow E \rightarrow F$ is generated offline. The decision to move from F to Black Zone then to Red Zone is taken online based on available requests. \\
 Black Zone - Consists of Black Circled nodes.\\
 Red Zone - Consists of Red Rectangle nodes.}
 \label{fig:samplepath}
\end{figure}

\begin{Example} 
Figure~\ref{fig:samplepath} provides an example of a zone path generated using ZAC. There is a partial zone path (generated offline) over individual locations (i.e., $A\rightarrow \ldots \rightarrow F$) and the completion of that zone path (online) using larger zones (black and red).
\end{Example}

\begin{figure}
 \centering
\includegraphics[width=0.95\textwidth,height=3.0in]{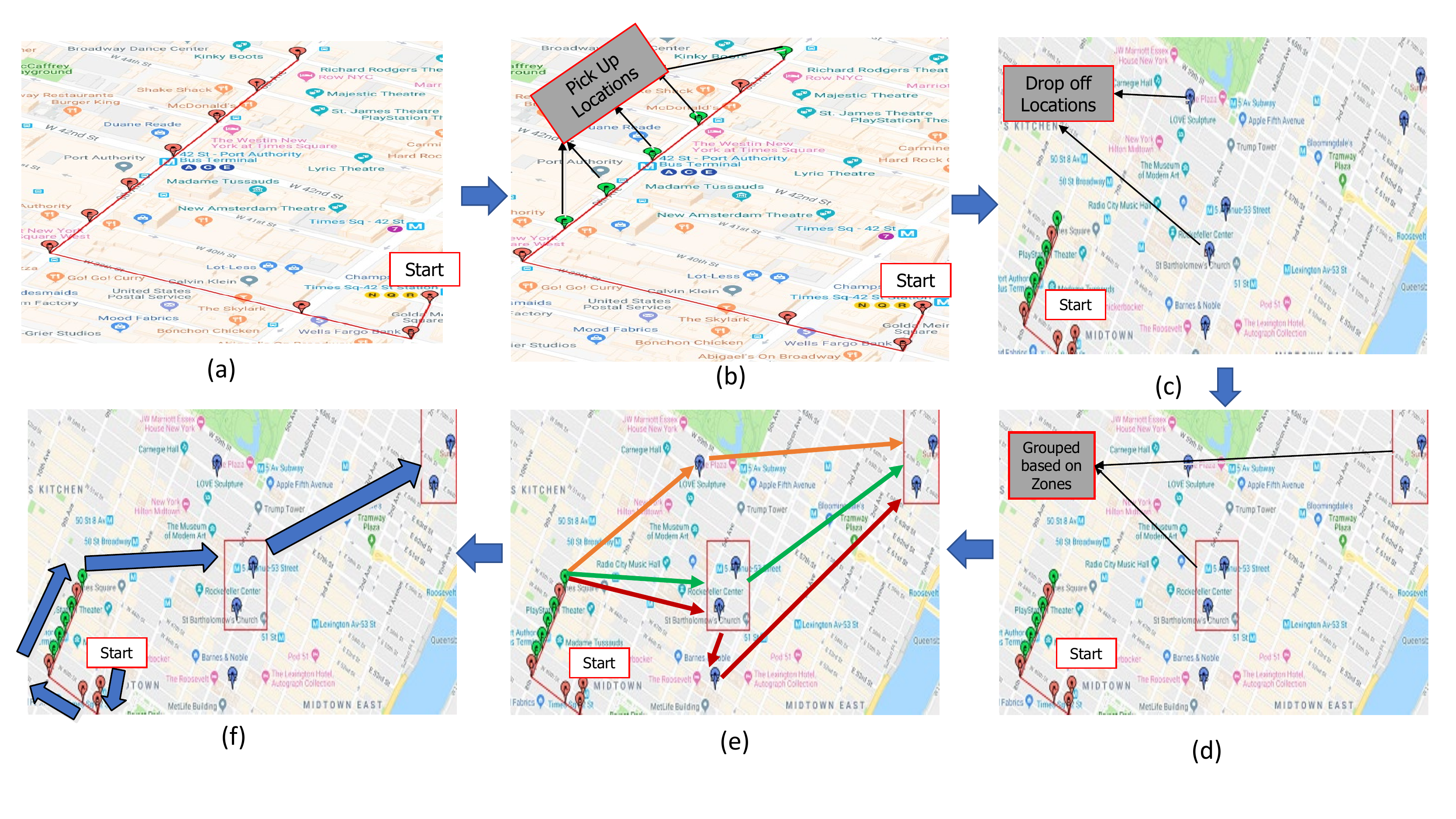}
 \vspace{-0.1in}
 \caption{Example showing different steps of ZAC. (a) One of the offline partial paths. ``Start" represents the start location of a vehicle. (b) Green markers represent the pick-up location of the incoming customer requests grouped along the path. (c) The blue markers represent the destination locations of the requests which are grouped along the path based on their pick-up location. (d) The destination locations are grouped together based on different zones (e) The path is completed by starting exhaustive search at the last green marker. The exhaustive search generates three different paths shown in orange, green and red colour. (f) The path represented by blue big arrows shows one of the complete zone paths starting at the initial location of the vehicle.}
 \label{fig:example}
 \vspace{-0.2in}
\end{figure} 

\begin{Example} 
Figure \ref{fig:example} shows all the steps of offline-online path generation process used as part of ZAC. We will refer to these steps in Section~\ref{sec:offline} and \ref{sect:online}.
\end{Example}

\subsection{Offline: Partial Paths Generation} 
\label{sec:offline}
The main challenge with generating a partial path online at the level of individual locations -- even for a time span of only maximum wait time, $\tau$ -- is the time taken to generate all the paths. Therefore, we compute these partial paths (of span $\tau$ seconds) offline by generating all simple paths of duration $\tau$ in the network ${\cal G}$. The number of all possible paths grows exponentially with the increase in the value of $\tau$ and increase in the number of locations. In case all possible paths can not be generated due to memory constraints, we can employ a data driven approach (based on historical data) to generate paths that have high likelihood of grouping large number of requests. More details about the data driven approach is present in appendix~\ref{appendix:offline}. 

Figure~\ref{fig:example}(a) shows one of the offline generated partial paths as an example starting at the location denoted by ``Start" and visiting all the locations represented by red markers. 

The start point of each offline partial path is one of the locations from the set ${\cal L}$ and the end point can be any of the locations which is reachable within $\tau$ duration from the start location. These offline partial paths (${\cal P}_{off}$) are stored by indexing on the start location and start time (${\cal P}_{off}[l,t]$)~\footnote{We discretize the time at the level of 10 seconds.}. The start time associated with the path indicates the time at which the first node (location) in the path is visited. For a clear explanation, two paths starting at the same location but having different start times are considered different. This is because vehicles can become available at the same location but at different time. These offline partial paths are further indexed by the location and time of each node present in the path for quick online processing.

\begin{algorithm}
\caption{ZAC-Online()}
{ %\small
\begin{algorithmic}[1]
\STATE{$t=starttime$ (in seconds)}
\STATE{${\cal P}_{off} = \bigcup\limits_{\substack{l \in {\cal L},\\t' < \tau}} P_{off}[l,t']=$ LoadOfflinePartialPaths()}
\STATE{${\cal T} =$LoadTravelTimes(), ${\cal S}_{p}=$LoadShortestPaths()}
\WHILE{$t~ <~ endtime$}
\STATE{$t_{1} = t-starttime$}
\IF{$(t_1)\%\Delta == 0$}
\STATE{${\cal D},{\cal V} \leftarrow $GetCurrentDemand-VehicleStatus($t$)}
%\STATE{${\cal V} \leftarrow $GetCurrentVehicleStatus($t_1$)}
\STATE{${\cal P},Pv,Pr,b,N$ = GenerateRPVGraph($t,{\cal P}_{off},{\cal D},{\cal V},{\cal T},{\cal S}_{p}$)}
\STATE{SolveOptimization(${\cal P},Pv,Pr,b,N$)}
\ENDIF
\STATE{UpdateVehicleStatus()}
\STATE{$t=t+1$}
\ENDWHILE
\end{algorithmic}
\label{alg:zac}}
\end{algorithm}

\subsection{Online}
\label{sect:online}
We now describe the crucial online component of ZAC that generates the RPV graph and finds the optimal match on the generated RPV graph. The pseudocode for the online component ZAC-Online is provided in Algorithm \ref{alg:zac}. After loading the offline computed partial paths, travel times and shortest paths, at every decision epoch, ZAC-Online considers the currently available batch of requests and current vehicle status to find the optimal assignment in two steps: (1) Generation of the Request, Path and Vehicle (RPV) graph and (2) Finding optimal match in RPV graph using a linear integer optimization model.

\noindent We now describe the two steps of ZAC in detail.
\subsubsection{Generation of the RPV graph}
As shown in Algorithm \ref{alg:rpvgraph}, there are three key steps to RPV graph generation: (1) Online processing of Offline Partial Paths; (2) Online Partial Zone Path Completion; (3) Identifying edges in the RPV graph.

\begin{algorithm}[htbp]
\caption{GenerateRPVGraph($t$, ${\cal P}_{off}$, ${\cal D}$, ${\cal V}$, ${\cal T}$, ${\cal S}_{p}$)}
{%\small
\begin{algorithmic}[1]
\STATE{${\cal P}'_{off},{\cal R}' = $ ProcessOfflinePartialPaths($t$, ${\cal P}_{off}$, ${\cal D}$, ${\cal V}$,${\cal T}$, ${\cal S}_{p}$)}
\STATE{${\cal P},{\cal R}'' =$ OnlineCompletion($t,{\cal P}'_{off},{\cal R}',{\cal T},{\cal S}_{p}$)}
\STATE{${\cal P},Pv,Pr,b,N =$ IdentifyEdgesRPVGraph($t$, ${\cal P}$, ${\cal D}$, ${\cal V}$, ${\cal R}''$)}
\RETURN{${\cal P},Pv,Pr,b,N$}
\end{algorithmic}
\label{alg:rpvgraph}}
\end{algorithm}

\noindent \underline{\textbf{\textit{Online Processing of Offline Partial Paths:}}} The offline generated partial paths are processed online based on the pick-up locations of the currently available requests and current status of all vehicles (location of vehicles, currently assigned requests to vehicles). 

Figure~\ref{fig:example} (b) and Figure~\ref{fig:example} (c) shows the online processing of the offline partial path based on the currently available requests. In Figure~\ref{fig:example}(b), the green markers represent the pick-up location of the requests which can be grouped using the offline partial path and in the Figure~\ref{fig:example}(c), we use the blue markers to represent the destination locations of the requests which had pick-up at one of the locations represented using green marker.
We also store a lower and upper bound on the time by which each of the blue marker should be visited for the delay constraints to be satisfied. Therefore, by online processing of offline partial paths, for each path, we get the set of requests which can be grouped based on their pick-up location and we also get the order in which the locations should be visited. The detailed algorithm is shown in Algorithm \ref{alg:processoffline}.

%The online processing of offline partial paths help in forming small overlapping clusters of requests based on their pickup location and it also provides a pickup order for these requests. Once we get the set of requests grouped based on the pickup location for each partial path, for each path, we can store the destination location of the requests grouped along the path and also a lower and upper bound on the time between which the destination location needs to be visited for the delay constraints to be satisfied. 

The algorithm takes as an input the set of all offline generated partial paths (${\cal P}_{off}$), incoming customer requests (${\cal D}$), and current status of all vehicles (${\cal V}$). The output of the algorithm is the set of offline partial paths (${\cal P}'_{off} \subset {\cal P}_{off}$) which can serve at least one request from the set ${\cal D}$. For each of these paths, the algorithm also provides the information about the requests grouped along the path (${\cal R}'$). The information includes the destination location and the lower and upper bound on the time by which the location needs to be visited. The individual steps of the algorithm are described below. 

Steps \ref{step:1}-\ref{step:2} of the algorithm ensures that we consider only those paths that start at a location and time where at least one vehicle is present, and these paths are processed in parallel using multiple threads. The GetPathsFromIndex function returns the set of offline partial paths that visit the given location within a given time interval and uses the pre-computed offline indexes for quick online retrieval. Step \ref{step:3} stores the set of destination locations of the currently available requests grouped along the path (based on the pick-up). In addition to the destination location, we also store the lower and upper bound on the time by which the location should be visited. Similarly, in step \ref{step:6}, we store the destination locations of the requests previously assigned to vehicles. In step \ref{step:6}, we consider only those paths that can potentially satisfy all the previously assigned requests for a vehicle. This is because a vehicle will be assigned to a path if and only if it can serve all the previously assigned requests. In addition, a vehicle should deviate from its current path only if it can be assigned to a new request, therefore, we consider only those paths which can pick at least one of the newly available requests.

\begin{algorithm}
\caption{{ProcessOfflinePartialPaths($t$,${\cal P}_{off}$,${\cal D}$,${\cal V}$, \\${\cal T}$,${\cal S}_{p}$)}}
{%\small
\begin{algorithmic}[1]
\STATE{${\cal P}'_{off}=[]$,$Lt_{v} = \bigcup\limits_{i \in {\cal V}} (\mu_{i},\omega_{i})$} \label{step:1}
\STATE{Create $H$ threads. Each thread $h$ processes ${\cal P}_{off}^{h} = \bigcup\limits_{k'\in Lt_{v}^{h}} {\cal P}_{off}[k'],$ ,s.t., $Lt_{v}^{a} \cap Lt_{v}^{b} = \phi, \forall a \neq b$ and $\cup_{h} Lt_{v}^{h} = Lt_{v}$ }\label{step:2}
\FOR{ each thread $h$}
\STATE{${\cal V}' \subset {\cal V}, s.t., \forall i' \in {\cal V'}, (\mu_{i'},\omega_{i'}) \in Lt_{v}^{h}$}
\FOR{ $j \in {\cal D}$}
\STATE{ ${\cal P}_{off}^{h,j}$ = GetPathsFromIndex(${\cal P}_{off}^{h},o_{j},a_{j}-t,a_{j}-t+\tau)$}
\FOR{each path $k \in {\cal P}_{off}^{h,j}$}
\STATE{$lb_{j} = a_{j}-t+{\cal T}(o_{j},d_{j}), ub_{j} = lb_{j} + \lambda$}
\IF{${\cal R}[k]$ contains $d_{j}$}
\STATE{${\cal R}[k][d_{j}][1] = max(R[k][d_{j}][1],ub_{j})$} \label{step:hupper}
\ELSE 
\STATE{${\cal R}[k].add(d_{j},(lb_{j},ub_{j}))$}\label{step:3}
\ENDIF
\STATE{${\cal R}_{p}[k].add(o_{j})$}
\IF{$lb_{j} < \tau$ and $k$ visits $d_{j}$}\label{step:d1}
%\IF{$k$ visits $d_{j}$}
\STATE{${\cal R}_{p}[k].add(d_{j})$}
\ELSIF{$ub_{j} < \tau$ and $k$ does not visit $d_{j}$}
\STATE{${\cal R}[k].remove(d_{j},(lb_{j},ub_{j}))$}
\ENDIF
%\ENDIF
\ENDFOR
\ENDFOR
\FOR{$ i \in {\cal V}'$} \label{step:5}
\STATE{${\cal R}[k],{\cal R}_{p}[k]$ = GetPathsForVehicle($i,q_i,{\cal R}[k],{\cal R}_{p}[k],{\cal P}_{off})$}\label{step:6}
\ENDFOR
\FOR{each path $k$ }\label{step:8}
\IF{$|{\cal R}_{p}[k]| > 0$}
\STATE{Remove nodes not in $R_{p}[k]$, update ${\cal R}_{k}$ using ${\cal T}$, ${\cal S}_p$}\label{step:9}
\STATE{${\cal P}_{off}^{'h}.add(k)$}
\ENDIF
\ENDFOR
\ENDFOR
\FOR{ each thread $h$}
\STATE{${\cal P}'_{off}.addAll({\cal P}^{'h}_{off})$}
\ENDFOR
\RETURN{${\cal P}'_{off},{\cal R}$}
\end{algorithmic}
\label{alg:processoffline}}
\end{algorithm}

Steps \ref{step:d1} and \ref{step:6} ensure that if the drop-off location of request can be visited in the partial path, then it is considered in the processing. 

In the end, in steps \ref{step:8}-\ref{step:9}, as an optimization, we only keep those locations in the partial paths that correspond to a pick-up or drop-off location and update the travel time and path between the locations using ${\cal T}$ and ${\cal S}_p$.

{\em The offline generated partial paths significantly improve the scalability of completing the path online using exhaustive search. This provides more time online for considering more zone paths and hence more relevant trips. }\\

\noindent \underline{\textbf{\textit{Online Partial Zone Path Completion:}}} The subset of offline partial paths obtained from the algorithm used in previous step (Algorithm~\ref{alg:processoffline}) are completed online in this step using exhaustive search starting at the last location in the partial path. For the example in Figure~\ref{fig:example}, we complete the offline partial path by starting the exhaustive search at the last green marker. The search space is the set of destination locations represented by blue markers. To reduce the computational complexity, we group nearby destination locations using zones. Figure~\ref{fig:example} (d) shows that the destination locations of the requests which are nearby are grouped using zones. Figure~\ref{fig:example}(e) shows that we get multiple zone paths by completing a single offline partial path using exhaustive search. We use orange, green and red colour arrows to highlight three different paths. Finally, we highlight one of the completed zone paths (out of the three possible paths) starting at the location ``Start" in Figure~\ref{fig:example}(f) using big blue arrows. 

The complete process is formally shown in the Algorithm~\ref{alg:processonline}. The output of Algorithm~\ref{alg:processoffline} serves as an input to the Algorithm~\ref{alg:processonline}, i.e., the algorithm takes as an input, the set of offline partial paths (${\cal P}'_{off}$) and the information about requests associated with each of the offline partial path (${\cal R}'$). The output of the algorithm is the set of completed zone paths (${\cal P}$). For each of these zone paths in set ${\cal P}$, the algorithm also outputs the set of requests which can potentially be served using the zone path (${\cal R}$). The detailed steps of the algorithm are explained below. 

As the offline partial paths are independent of each other, to speed up the path generation process, we perform the online path completion of the offline partial paths ${\cal P}'_{off}$ in parallel by creating multiple threads as shown in the pseudocode provided in Algorithm \ref{alg:processonline}. To complete each offline partial path, we need to consider the destination locations of all the requests associated with the path. As we also have a lower and upper bound on the time by which the destination locations of the requests should be visited, we can prune the search space by exploring only those branches in the search tree where these time limits are satisfied. 

The computational complexity of online partial path completion is dependent on the number of destination locations (size of ${\cal R}[k]$ in Algorithm \ref{alg:processonline}) and can be significant, therefore, we use zones (and not individual locations) in this step. As mentioned before, by using zones, travel time is approximately represented, which can result in additional delay for requests. The additional delay introduced is dependent on the size of the zones chosen. The size of the zone is defined as the time taken to travel within a zone. Zone size 0 indicates that locations in set ${\cal L}$ are used. 

Therefore, to consider a trade-off between computational complexity and the quality of solution, we propose picking the zone sizes dynamically for each offline partial path. In order to fix the amount of dynamism in zone size, we use a parameter $M$ that defines the number of different zone sizes that can be used in completion of offline partial paths. $M=1$, implies static zone sizes, i.e., using zones of a fixed size for online completion of all offline partial paths. The zones of $M$ different sizes are generated offline and in the step \ref{step:zsize}, depending on the number of destination locations and $M$ available zone sizes, we decide the appropriate zone size for the partial path $k$~\footnote{In the experiments, we use $M=4$ with zone sizes 0,60,120,300 and use the zone size that reduces the number of locations to 12 (this provides the best trade-off between runtime and solution quality and is determined based on experiments.}. Please note that the exhaustive search in step \ref{step:ex}, will return multiple completed zone paths corresponding to a single partial path $k$ as shown in the example (Figure \ref{fig:example}) before.

\begin{algorithm}[h]
\caption{{ OnlineCompletion($t,{\cal P}'_{off},{\cal R}',{\cal T},{\cal S}_{p}$)}}
{%\small
\begin{algorithmic}[1]
\STATE{${\cal P} = [],{\cal R} =[]$} 
\STATE{Create $H$ threads. \\Each thread $h$ processes ${\cal P}_{off}^{'h} \subset {\cal P}'_{off}$ ,s.t., ${\cal P}_{off}^{'h} \cap {\cal P}_{off}^{'h'} = \phi, \forall h \neq h'$ and $\cup_{h} {\cal P}_{off}^{'h} = {\cal P}'_{off}$ }
\FOR{ each thread $h$}
\STATE{${\cal P}_{on}^{h} = [],{\cal R}_{on}^{h} = []$}
\FOR {each path $k$}
%\STATE{${\cal R}' = {\cal R}[k]$}
%\IF{$|{\cal R}'| > 12$}
%\FOR{each available zone size z}
\STATE{$z =$ getAppropriateZoneSize$({\cal R}[k],M)$}\label{step:zsize}
\STATE{${\cal R}' =$ convert$({\cal R}[k],z)$}
%\IF{$|{\cal R}'| <= 12$}
%\STATE{break}
%\ENDIF
%\ENDFOR
%\ENDIF
\STATE ${\cal P}^{h}_{on},{\cal R}^{h}_{on}$ = ExhaustiveSearch($end\_node(k),{\cal R}',{\cal P}_{on}^{h},{\cal R}_{on}^{h}$) \label{step:ex}
\ENDFOR
\ENDFOR
\FOR{ each thread $h$}
\STATE{${\cal P}.addAll({\cal P}^{h}_{on})$}
\STATE{${\cal R}.addAll({\cal R}^{h}_{on})$}
\ENDFOR
\RETURN{${\cal P},{\cal R}$}
\end{algorithmic}
\label{alg:processonline}}
\end{algorithm}

For the objective of maximizing the number of requests served, the paths that start at the same location at the same time and serve a subset of requests served by another path are redundant. This is because we check for capacity constraints in the optimization formulation presented in the next section. So, a single path serving $r$ requests can be used to represent all request combinations, $\sum_{i=1}^{r} \binom{r}{i}$. Therefore, the search tree in step \ref{step:ex} of Algorithm \ref{alg:processonline} can be pruned appropriately to search only for non-redundant paths. This reduces the size of set ${\cal P}$, which, in turn reduces the complexity of the optimization formulation presented in the Section \ref{sect:opt}.  

\begin{figure}[htbp]
 \centering
 {\label{fig:figna}\includegraphics[width=0.95\textwidth,height=2.0in]{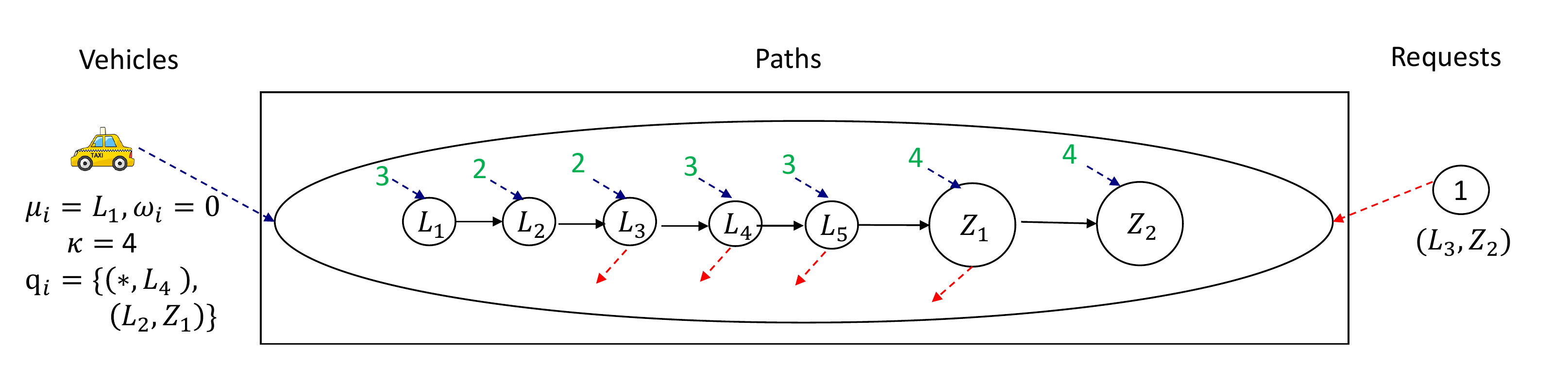}}
%\subfloat[]{\label{fig:figsp}\includegraphics[width=0.9\textwidth,height=1.6in]{maxflow.png}}
 \caption{Representation of assignment of vehicle and request to a zone path}
 \label{fig:assignpath}
\end{figure}

\noindent \underline{\textbf{\textit{Identifying Edges in the RPV Graph:}}} Once the zone paths (${\cal P}$) are created using the offline-online method described above, we construct the RPV (Request Path Vehicle) graph by finding the set of requests and vehicles that can be assigned to each of the generated zone path. We use the information available from the previous 2 steps about the requests and vehicles that can be assigned to these zone paths and process the paths in parallel using multiple threads to speed up the computation. This step is essential as in Algorithm \ref{alg:processoffline}, when the same destination location has a different value for the upper limit on time in step \ref{step:hupper}, we take the maximum value. Therefore, in the path generated using Algorithm \ref{alg:processonline}, the delay constraint may be violated for some requests in such cases. 
 
In this step, we ensure that a request is assigned to a zone path, if and only if, the path visits the pick-up and drop-off location of a request within the delay constraints. The binary constants $b$ (defined in Table \ref{table:notations} and used in optimization formulation presented next) are also populated in this step. A vehicle $i$ represented by the tuple $(\mu_{i},\omega_{i},q_{i},\kappa)$ can be assigned to a zone path if the initial location of the vehicle, $\mu_i$, is same as the starting location of the path, the start time of the path is same as the availability time of vehicle $\omega_{i}$ and the currently assigned set of requests, $q_i$, can be served using the path. The vehicle capacity $\kappa$ along with $q_{i}$ is used to compute the number of free seats ($N$) in the vehicle at each zone/location. 

\begin{Example}
Figure \ref{fig:assignpath} shows a graphical view of the same for a single vehicle, request and path. The path is represented using a sequence of locations/zones in the order in which they will be visited. In Figure~\ref{fig:assignpath}, we use blue arrows to denote the incoming flow by vehicle $i$ assignment and green numbers indicate the number of free seats at the location/zone for vehicle $i$. The number of free seats is computed by taking $\kappa$ and $q_i$ of the vehicle into consideration. In figure, in the representation of $q_i$, we use * to indicate that customer is already present in the vehicle and provide its drop-off location. The red arrows indicate outgoing flow by request assignment. \end{Example}

At each location, the optimization formulation presented next, will ensure that the outgoing flow (the number of requests assigned) is less than or equal to the incoming flow (total number of free seats in the vehicles assigned to the path), i.e., at each location/zone capacity constraint are satisfied. 

%\begin{algorithm}
%\caption{{IdentifyAndUpdateEdges(${\cal P},{\cal R}'',t,{\cal D},{\cal V}$)}}
%{\small
%\begin{algorithmic}[1]
%\STATE{}
%\end{algorithmic}
%\label{alg:identifyedges}}
%\end{algorithm}
\subsubsection{Finding Optimal match in RPV graph} 
\label{sect:opt}
We now describe the integer linear programming optimization formulation to optimize the assignment of requests and vehicles to zone paths. ${\cal P}$ denotes the set of zone paths generated in previous step. ${\cal P}_{m}^{n}$ is used to denote the $n^{th}$ location/zone in zone path $m$. Let $Pr_{j} \subset {\cal P}$ denotes the set of paths that can serve request $j$ while satisfying delay constraints. Similarly $Pv_{i}$ denotes the set of paths that can be assigned to vehicle $i$ based on its current location $\mu_{i}$, availability time $\omega_{i}$ and already assigned/picked-up requests $q_{i}$. Binary constants $b_{jm}^{n}$ are set to 1 if the pick-up location of request $j$ is visited but drop-off location/zone is not visited along path $m$ by $n^{th}$ location/zone. These are computed as part of generation of RPV graph as shown in previous section. 
Table \ref{table:notations} describes the notation used in the optimization formulation. 
\begin{table}[htbp]
{%\small
\begin{center}
\begin{tabular}{|l|l|}
\hline
\textbf{Variable} & \textbf{Description}\\
\hline
$x_{jm}$ & Binary variable denoting if the request $j \in {\cal D}$ is assigned to path $m$.\\
\hline
$y_{im}$ & Binary variable denoting if the vehicle $i$ is assigned to the path $m$.\\
\hline
$Pv_{i}$ & $Pv_{i} \subset {\cal P}$ denotes the set of paths that can be assigned to vehicle $i$ based on  \\
& its current status $\mu_{i}$, $\omega_{i}$ and $q_{i}$.\\
\hline
$Pr_{j}$ & $Pr_{j} \subset {\cal P}$ denotes the set of paths that can be assigned to request $j \in {\cal D}$.\\
\hline
$b_{jm}^{n}$ & Binary constant: 1 if $\exists n' : n > n'~ P_{m}^{n'} = o_{j}$ $\&\&~ \nexists n'': n'' < n, n'' > n' ~ P_{m}^{n''} == d_{j}$ \\
\hline
$N(i,m,n)$ & Number of free seats in the vehicle $i$ for path $m$ at $n^{th}$ location/zone. \\
\hline
\end{tabular}
\end{center}
}
\caption{ Notations}
\label{table:notations}
    \vskip -10pt
\end{table} 
    
The objective of the optimization formulation described in Table \ref{opt:ride} is to maximize the number of served requests. Constraints \eqref{cons:2} and \eqref{cons:3} ensure that each vehicle and each request is assigned to at most one path. Constraint \eqref{cons:4} ensure that for every path at every location/zone capacity constraints are satisfied. The capacity constraints can be violated only while picking up a new request, therefore, the constraint \eqref{cons:4} is redundant for the locations/zones visited after $\tau$ duration. 

The formulation is run at every decision epoch, i.e., after every $\Delta$ seconds. The solution of the optimization formulation provides assignment of vehicles and requests to paths. Using these assignments, we can perform the assignment of requests to vehicles. The paths assigned to vehicle are also updated to keep only those locations that correspond to pick-up or drop-off location of assigned requests and update the travel time and path between the locations using ${\cal T}$ and ${\cal S}_p$. 

Once a vehicle is assigned to a set of requests at any decision epoch, the assignment is not changed but the path of vehicle can change at next decision epoch to accommodate additional requests. The current set of requests assigned to a vehicle, $q_i$, limits the number of paths to which it can be assigned in subsequent decision epochs. The number of free seats in vehicle $i$ for path $m$ at location/zone $n$, $N(i,m,n)$ is computed based on $\kappa$ and $q_{i}$ (as shown in Figure \ref{fig:assignpath}) and is 0 if $m \notin Pv_{i}$.

Similar to Alonso {\em et al.}~\cite{alonso2017demand}, we perform a re-balancing of unassigned vehicles to high demand areas at the end of optimization formulation. 

\begin{table}[htbp]
\center
    \begin{tabular}{|r|}
    \hline

    \begin{minipage}{0.95\textwidth}
~\\
\textbf{SolveOptimization(${\cal P},Pv,Pr,b,N$):}
{%\small
\begingroup
\addtolength{\jot}{-2pt}
\begin{align}
\max \quad & \sum_{j \in {\cal D}} \sum_{m \in Pr_{j}} x_{jm} \\
s.t. \quad & \sum_{m \in Pr_{j}} x_{jm} \leq 1 ::: \forall j \in {\cal D} \label{cons:2}\\
& \sum_{m \in Pv_{i}} y_{im} \leq 1 ::: \forall i \in {\cal V} \label{cons:3}\\
& \sum_{j \in {\cal D}} x_{jm} \cdot b_{jm}^{n} \leq \sum_{i} y_{im} \cdot N(i,m,n) ::: \forall m \forall n \label{cons:4}
\end{align}
\endgroup }
\end{minipage} \\
    \hline
    \end{tabular}
    \caption{Optimization Formulation for ZAC}
    \label{opt:ride}
    \vskip -10pt
    \end{table}
%\section{Multi Period Two Stage Stochastic Optimization}

%\section{Approximations}
%\subsection{Benders Decomposition}
\section{ZACBenders: A non-myopic approach for solving RMP}
\label{sect:zacbenders}
In this section, we first present the challenges in solving RMP with future information (represented using samples, $\xi^D$ in the RMP model) and then present our non-myopic approach ZACBenders. The potential samples, $\xi^D$ can be obtained by considering the demand observed in the past data. After considering future information, the goal is to find the assignment of vehicles to requests that maximizes the sum of objective value at the current decision epoch and the expected objective value for the future decision epochs.

\subsection{Challenges in solving RMP with future information}
\label{sect:challenges}

\begin{figure}
 \centering
 {\label{fig:figexact}\includegraphics[width=0.95\textwidth,height=3.2in]{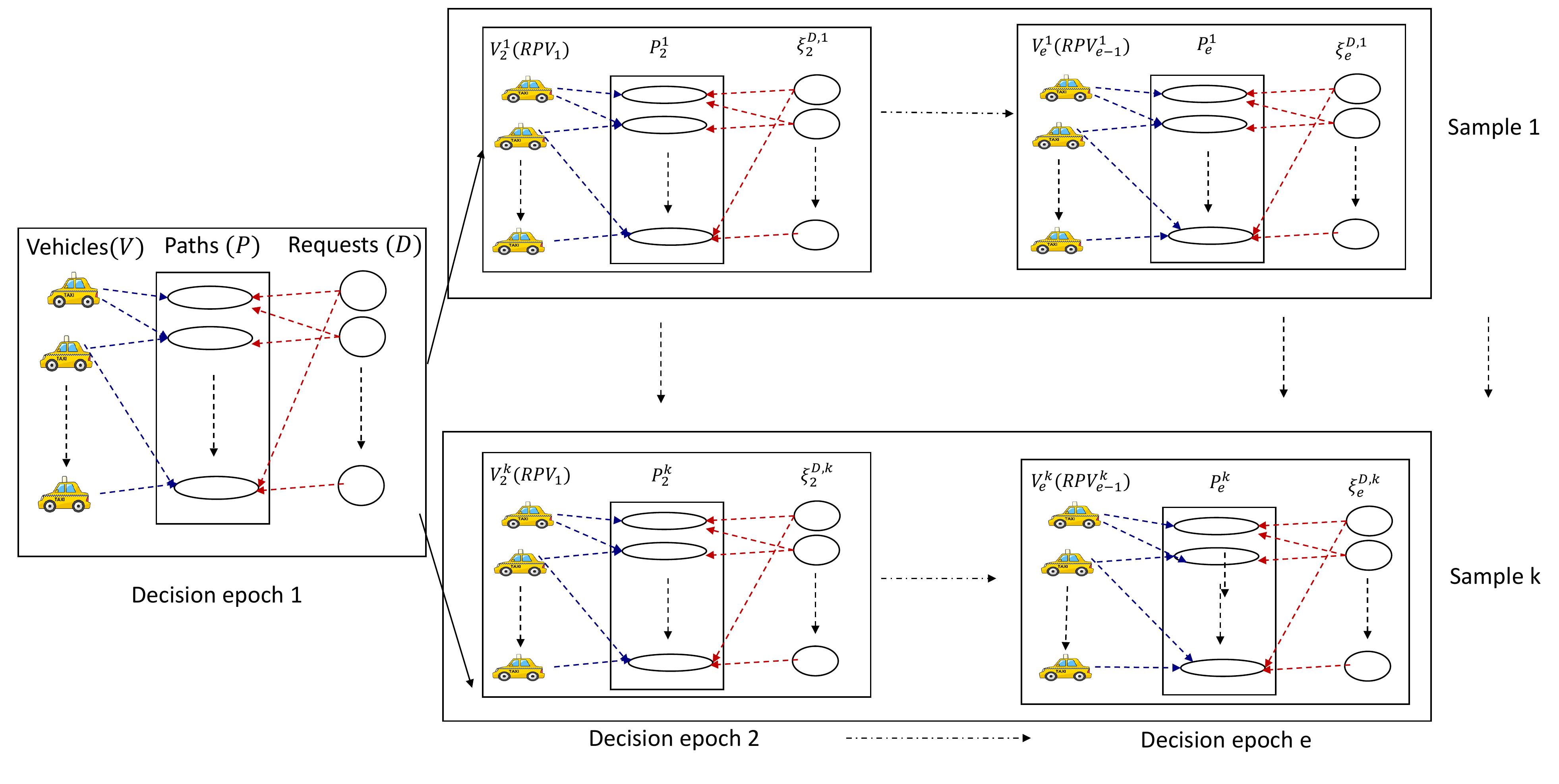}}
%\subfloat[]{\label{fig:figsp}\includegraphics[width=0.9\textwidth,height=1.6in]{maxflow.png}}
 \caption{Assignment of vehicles and requests to zone paths over multiple samples of future demand. We use $RPV^{k}_{e}$ to denote the RPV graph for decision epoch $e$ in sample $k$. For the special case of first decision epoch, we denote the graph by $RPV_{1}$. ${\cal P}^{k}_{e}$ denote the set of paths generated for decision epoch $e$ in sample $k$. $\xi^{D,k}_{e}$ denote the set of requests available at decision epoch $e$ in sample $k$. $V^{k}_{e}(RPV_{e-1}^{k})$ denote the state of vehicles at decision epoch $e$ in sample $k$ as a result of assignments obtained by solving the RPV graph at previous decision epoch in the same sample. Therefore, at each decision epoch for each sample, it is a tripartite matching between vehicles, paths and requests. }
 \label{fig:exactassignsamples}
\end{figure}

As shown in the Figure \ref{fig:exactassignsamples}, to solve RMP with future information ($\xi^D$) we need to assign vehicles and request to the zone paths at each decision epoch and for each sample ($\xi^{D,k}$). The state (i.e., location and requests being served) of vehicle at each decision epoch should be updated based on the assignments (obtained by solving the RPV graph) at previous decision epochs. The paths at future decision epochs for each sample need to be generated by considering the requests present in the sample and the optimization problem should be updated to consider the assignments at future decision epochs for all the samples. 

There are two major bottlenecks in the above process.
\begin{enumerate}
\item As described in the Section \ref{sect:zac}, the path generation process for a single decision epoch is challenging, so generating paths in real-time after considering requests in each sample for all future decision epochs for all possible updates to the vehicles states is computationally intractable. 
\item The optimization formulation of assigning vehicles and requests to zone paths is an integer optimization problem. Including requests for all the samples at future decision epochs in this integer optimization formulation increases the number of variables and constraints, which makes it difficult to solve online in real-time.
\end{enumerate}

\subsection{ZACBenders Approach}
The overall flow of ZACBenders is similar to ZAC with the only difference in the step of finding the optimal assignment of vehicles and requests to zone paths. Table \ref{table:zacbendersdifference} highlights the difference between ZAC and ZACBenders approach. ZACBenders considers future information in the step of finding the optimal assignment of vehicles and requests to zone paths. As mentioned in Section \ref{sect:challenges}, incorporating future information makes the problem challenging, therefore, we first provide a two-stage stochastic approximation, ZACFuture to handle these challenges. To efficiently solve the ZACFuture optimization formulation in real-time, we employ Benders Decomposition.

\begin{table*}[htbp]
\center
	\begin{tabular}{|l|l|}
	\hline
	\begin{minipage}[t]{0.46\textwidth}
\textbf{{~~~~~~~~~~~~~~~~~~~~~~~~~ZAC}}
\end{minipage} & 
\begin{minipage}[t]{0.46\textwidth}
\textbf{~~~~~~~~~~~~~~ZACBenders}
\end{minipage} \\
	\hline
    \multicolumn{2}{|c|}{\textbf{Offline}} \\
    	\hline
~&~\\
1. Generation of all partial location paths & 1. Same \\
of time span, $\tau$ from every location. & ~\\
	\hline
    \multicolumn{2}{|c|}{\textbf{Online}} \\
\hline
1. Generation of RPV graph & 1. Same \\
2. Finding optimal assignment of requests &2. Process the requests in $|\xi^{D}|$ samples\\
and vehicles to zone paths by using the & to generate the second stage of the \\
(0/1) integer optimization in Table \ref{opt:ride}. & proposed two-stage approximation. \\
~& 3. Follow the steps in Figure ~\ref{fig:zacbender} to find \\
~& the optimal assignment of requests and \\
~&vehicles to zone paths while considering \\
~&future information.\\
\hline 
%completing the partial paths and identifying edges in RPV graph.  & \\
%\hline
%\hline
	\end{tabular}
	\caption{Differences between ZAC and ZACBenders}
	\label{table:zacbendersdifference}
	\vskip -10pt
	\end{table*}

\subsubsection{Two-Stage Stochastic Approximation}
\label{sect:approx}
As mentioned in the Section \ref{sect:challenges}, it is difficult to solve RMP with future information by generating paths considering requests in all samples, therefore, we propose a two-stage stochastic approximation~\footnote{We have experimented with many other approximations such as considering simple extensions of the existing zone paths to include request in samples and then assigning requests in samples to these extended paths but we describe in detail the approximation that worked best in practice. We acknowledge that it is possible to improve the performance even more by designing better approximations.}. The first stage assigns vehicles and requests available at current decision epoch to the zone paths. In the second stage, for each sample, instead of solving a tripartite matching problem (between vehicles, paths and requests) at each future decision epoch, we solve a weighted bipartite matching between vehicles and requests available for assignment at all future decision epochs. The tripartite matching is NP-hard, but the weighted bipartite matching is polynomial time solvable, therefore, this approximation makes the second stage problem simpler. Specifically, we employ a decomposition of the resultant optimization problem (more details in Section \ref{sect:zacbendersdetail}) to get real-time performance.

We now describe the approximations which allow us to simplify the second stage problem for each sample by modelling it as a weighted bipartite matching problem. \\

\squishlist 
\item{ \em Approximation 1:} Instead of using exact locations from set ${\cal L}$, we use abstracted locations, i.e., zones. The origin/destination of each request in sample and the location of each vehicle is mapped to the zones. 
\item {\em Approximation 2:} A vehicle will serve requests in samples ($\xi^{D}$) only after it finishes serving all the currently assigned requests. That is to say, we ignore that any future request can be inserted in the vehicle's path and as a result a vehicle is considered available again for assignment only after it reaches the end of zone path generated in first step. 
\item {\em Approximation 3:} Requests in samples ($\xi^{D}$) can be assigned to the same vehicle if and only if they have identical origin zone, identical destination zone and the decision epoch at which they become available for assignment is also the same. 
\squishend

These approximations help in reducing the complexity of the problem, but they still allow us to get a good estimate of the future because of the following reasons:

\squishlist
\item The second approximation ensures that the vehicles that are considered for assignment at second stage are empty, i.e., they do not have any request assigned to them. So when the assignment optimization problem is solved (with limited look ahead duration of $\rho$), instead of assigning the vehicle to a longer duration path (which keeps it occupied for more than $\rho$ duration), it will assign the vehicle to those shorter duration zone paths (in the first stage) that redirect it to zones where future requests are present. At any decision epoch, it is easier to assign multiple requests to an empty vehicle as compared to a vehicle that has a passenger on board. Therefore, despite of ignoring that future requests can be picked up before dropping all currently assigned requests, this provides a good approximation.
\item The third approximation (along with first approximation) ensures that the requests are grouped when they have nearby pick-up and drop-off locations. Though we will miss grouping the requests that have on the way pick-ups/drop-offs, by making this approximation and using an appropriate zone size, we will still be able to implicitly consider a subset of possible paths.
\squishend

Formally, we map the origin and destination location of requests in the future decision epochs to zones of size~\footnote{As mentioned before, the size of the zone is defined as the time taken to travel within a zone.} $Z^{s}$ and the elements in ${\xi}^{D,k}$ are grouped together based on the origin/destination zone and decision epoch. After grouping, each element $j'$ of ${\xi}^{D,k}$ is represented using tuple $\left<o^{z,k}_{j'},d^{z,k}_{j'},e_{j'}^{k},\eta_{j'}^{k}\right>$ where $o^{z,k}_{j'}$ denotes the origin zone of the element $j'$ in sample $k$, $d^{z,k}_{j'}$ denotes the destination zone of the element $j'$ in sample $k$, $e_{j'}^{k}$ denotes the decision epoch at which element $j'$ of sample $k$ will be considered for assignment and $\eta_{j'}^{k}$ denotes the number of requests with origin $o^{z,k}_{j'}$, destination $d^{z,k}_{j'}$ at decision epoch $e_{j'}^{k}$ in sample $k$. We use ${\cal A}$ to denote the set containing all possible pairs of zones of size ${\cal Z}^{s}$ and decision epochs $e'$ such that if $e$ is the current decision epoch then $ e < e' \leq \floor{\frac{\rho}{\Delta}}$. Each vehicle is mapped to an element in set ${\cal A}$ and is assigned requests in samples by using above approximations.

\begin{figure}[htbp]
 \centering
 {\label{fig:figapprox}\includegraphics[width=0.95\textwidth,height=3.2in]{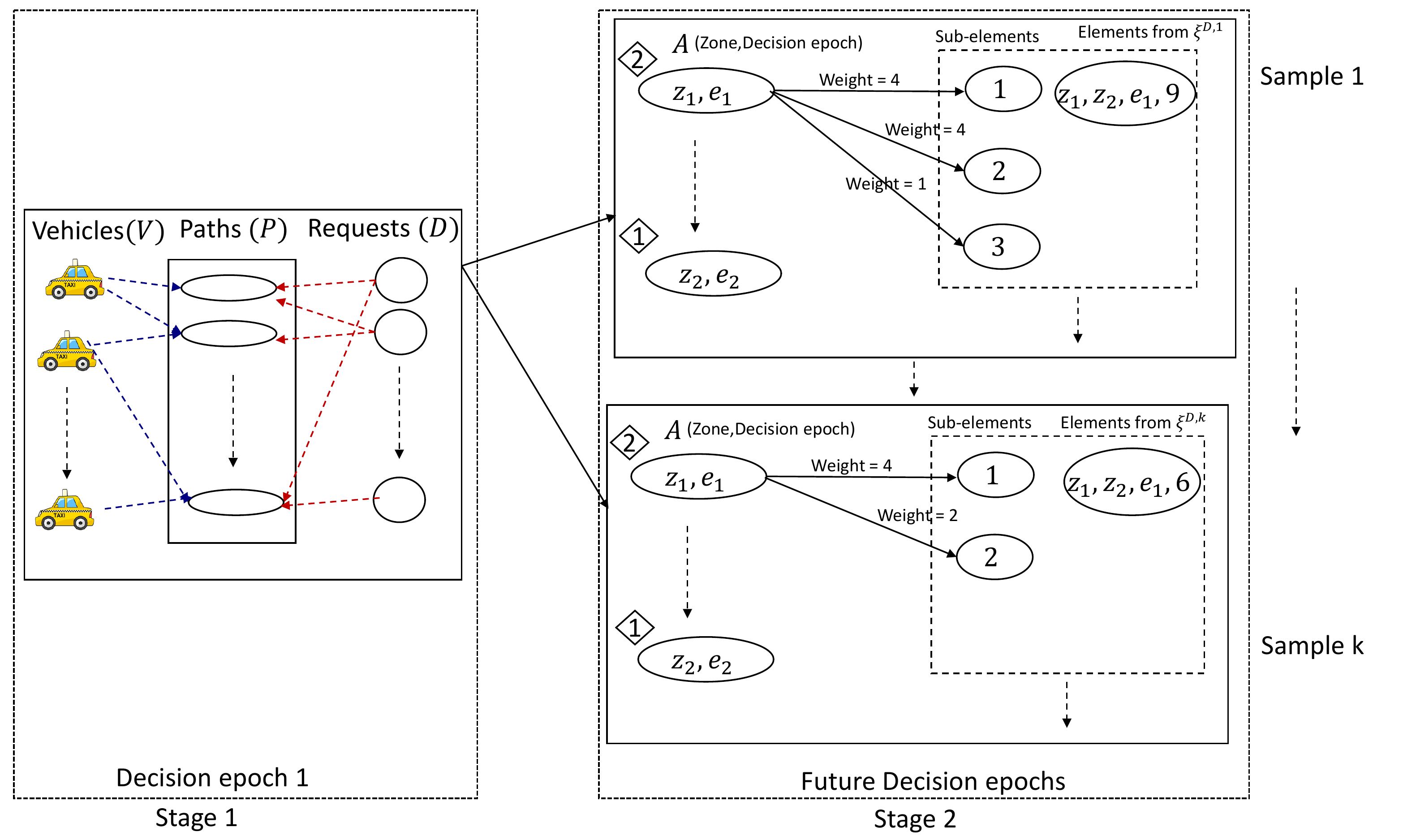}}
%\subfloat[]{\label{fig:figsp}\includegraphics[width=0.9\textwidth,height=1.6in]{maxflow.png}}
 \caption{Two-Stage stochastic approximation for assignment of vehicles and requests to zone paths over multiple samples of future demand (For $\kappa=4$). 
The zone and decision epoch mentioned in the oval are the zone and decision epoch at which the paths at the first stage in set ${\cal P}$ ends. Therefore, vehicles have dropped all the assigned requests from set ${\cal D}$ (in first stage), once they reach the zone and decision epoch present in the oval. The number inside diamond represents the number of empty vehicles present in the zone and decision epoch mentioned in the oval box. At second stage, for each sample, a bipartite matching is performed between empty vehicles and available requests for all future decision epochs as compared to the tripartite matching between vehicles paths and requests for each sample in each decision epoch as shown in Figure \ref{fig:exactassignsamples}.
 }
 \label{fig:twostageapprox}
\end{figure}

The assignment of vehicles to paths at first stage determines the zone and the decision epoch at which the vehicles will become available again for assignment (Approximation 2), and as all the vehicles have identical maximum capacity~\footnote{In the experiments, we show that even if all vehicles do not have identical maximum capacity, the approximation still works well. In that case, in the second stage, we take the maximum capacity of each vehicle as an average of all vehicle's maximum capacity.}, in the second stage, we can group the vehicles based on the zone and the decision epoch at which they become available again for assignment. A vehicle can be assigned to a request if and only if it can reach the origin location of request within the maximum allowed wait time, i.e. $\tau$. Let ${\cal E}^{k}$ denotes the set of all such assignment edges between vehicles and requests for sample $k$. To ensure that a vehicle can be assigned at most $\kappa$ requests and these requests can be grouped together as per Approximation 3, if $\eta_{j'}^{k} > \kappa$, we divide the element into $\ceil{\frac{\eta_{j'}^{k}}{\kappa}}$ subelements of element $j'$ and allow each subelement to be assigned at most once, but if a vehicle of type $i'$ is assigned to $r^{th}$ subelement of element $j'$ in sample $k$, the weight received is given by 

$w_{i'j'r}^{k} = \begin{cases} 
							0 ~~~ if (i',j',r) \notin {\cal E}^{k} \\
							\kappa ~~~ if (i',j',r) \in {\cal E}^{k}~and~r < \floor{\frac{\eta_{j'}^{k}}{\kappa}} \\
						   \eta_{j'}^{k} \mathbin{\%} \kappa  ~~~~ otherwise \end{cases} \label{eq:weighteq}$

Therefore, this creates a bipartite graph with one side containing vehicles grouped based on their type (zone and decision epoch at which they become available based on the path assigned in the first stage) and the other side containing the request groups (all subelements of the element $j' \in \xi^{D,k}, \forall j'$). Figure \ref{fig:twostageapprox} shows the graphical representation of the two-stage stochastic approximation, where first stage performs the tripartite matching between vehicles, requests and zone paths, and in the second stage within each sample, there is a bipartite matching. 

We now describe the optimization formulation that can be used to solve this two-stage stochastic approximation.  

%The first stage computes the assignment of vehicles and requests to the zone paths. In the second stage, for each sample, vehicle are assigned to requests at future decision epochs. 

\begin{table}[htbp]
\center
    \begin{tabular}{|r|}
    \hline

    \begin{minipage}{0.95\textwidth}
~\\
\textbf{ZACFuture(${\cal P},Pv,Pr,b,N, \xi^{D}$):}
{\small
\begingroup
\addtolength{\jot}{-2pt}
\begin{align}
\max \quad & \sum_{j \in {\cal D}} \sum_{m \in Pr_{j}} x_{jm} + \frac{1}{|\xi^{D}|}\sum_{k=0}^{|\xi^{D}|} \sum_{j' \in {\xi^{D,k}}}\sum_{r=0}^{\ceil{\frac{\eta_{j'}^{k}}{\kappa}}}\sum_{i' \in {\cal A}} w_{i'j'r}^{k} \cdot u_{i'j'r}^{k} \\
s.t. \quad & \sum_{m \in Pr_{j}} x_{jm} \leq 1 ::: \forall j \in {\cal D} \label{cons:s1}\\
& \sum_{i' \in {\cal A}} u_{i'j'r}^{k} \leq 1 ::: \forall j' \in \xi^{D,k}, 0 \leq  r < \ceil{\frac{\eta_{j'}^{k}}{\kappa}}, \forall 0 \leq k <  |\xi^{D}| \label{cons:s2}\\
& \sum_{m \in Pv_{i}} y_{im} \leq 1 ::: \forall i \in {\cal V} \label{cons:s3}\\
& \sum_{j \in {\cal D}} x_{jm} \cdot b_{jm}^{n} \leq \sum_{i} y_{im} \cdot N(i,m,n) ::: \forall m \forall n \label{cons:s5}\\
& \sum_{j' \in {\xi^{D,k}}}\sum_{r=0}^{\ceil{\frac{\eta_{j'}^{k}}{\kappa}}} u_{i'j'r}^{k} \leq \sum_{i \in {\cal V}}\sum_{m; f(m) = i'} y_{im}  ::: \forall i' \in {\cal A} \label{cons:s6}\\
&y_{im} \in \{0,1\} ::: \forall i \in {\cal V},m \in {\cal P} \\
&x_{jm} \in \{0,1\} ::: \forall j \in {\cal D},m \in {\cal P} \\
&u_{i'j'r}^{k} \in \{0,1\} ::: \forall i' \in {\cal A},j' \in \xi^{D,k},0 \leq  r < \ceil{\frac{\eta_{j'}^{k}}{\kappa}} ,0 \leq k < |\xi^{D}|
\end{align}
\endgroup }
\end{minipage} \\
    \hline
    \end{tabular}
    \caption{Optimization Formulation for Two-Stage stochastic approximation of RMP with future samples}
    \label{table:completeopt_samples_zac}
    \vskip -10pt
    \end{table}

\subsubsection{ZACFuture: Optimization formulation to Solve the Two-Stage Stochastic Approximation}
In this section, we describe the optimization formulation ZACFuture to solve the two-stage stochastic approximation presented in previous section. Table \ref{table:completeopt_samples_zac} presents the optimization formulation ZACFuture, which maximizes the number of requests served for the current decision epoch and the expected number of requests served over future demand samples. 

We use $u_{i'j'r}^{k}$ to denote the assignment of vehicle of type $i'$ (i.e., the vehicles present in the zone $z_{i'}$ at decision epoch $e_{i'}$, where $(z_{i'},e_{i'})$ is the tuple representation of element $i' \in {\cal A}$) to the $r^{th}$ subelement of $j'$ element of $\xi^{D,k}$. We also use $f(m)$ to denote the tuple $(z_{m},e_{m})$ where $z_{m}$ and $e_{m}$ denote the zone and decision epoch at which the vehicle will become available if it is assigned to path $m$. Constraints \eqref{cons:s2} ensure that the each subelement of $j'^{th}$ element of $\xi^{D,k}$ is assigned at most once and Constraints \eqref{cons:s6} ensure that the number of type $i'$ vehicles assigned is less than the number of vehicles available of type $i'$.

\subsubsection{Benders Decomposition to Efficiently Solve ZACFuture Optimization Formulation}
\label{sect:zacbendersdetail}
The complexity of the optimization formulation ZACFuture increases with the increase in the number of samples. To reduce this complexity, we exploit the following observation:
\begin{obsv}
\label{obs:dec}
In ZACFuture, once the assignment of vehicles to paths at the current decision epoch ($y_{im}$) is given, the optimization models for computing the assignment of vehicles to requests at future decision epochs, ($u^{k}_{ij'r}$) for each of the samples $k$, are independent of each other. 
\end{obsv}

The observation \ref{obs:dec} allows us to use Benders Decomposition~\cite{Benders} to decompose the large optimization formulation into multiple smaller problems that can be solved in parallel. Benders Decomposition is a master slave decomposition technique where the master problem finds the solutions for the integer variables; and the slave problem(s) is (are) used to find the solutions to all other variables (which can take any value in the interval and need not be integers) while keeping the values of the integer variables fixed to the value obtained by the master problem. The values obtained by slave problems help in generating benders cuts, which are added to the master problem and the master problem is solved again with these cuts to obtain an improved solution. This process is repeated till no more cuts can be added to the master problem. It is widely used to solve such two-stage stochastic problems~\cite{Murphy13,lowalekar2018online}.

Based on Observation~\ref{obs:dec}, $y_{im}$ are the difficult variables as they impact the values assigned to all the other variables. $x_{jm}$ are also difficult variables as they can take only integer values. As described in previous section, the second stage problem for each sample is a weighted bipartite matching problem. As the constraint matrix for weighted bipartite matching is totally unimodular, therefore, integrality constraints on the $u_{i'j'r}^{k}$ variables can be relaxed~\cite{hoffman2010introduction} after fixing the values of $y_{im}$ variables. Therefore, the master problem obtains the assignments for the ``difficult'' integer variables ($x_{jm}$ and $y_{im}$) and the slave problem(s) obtain the assignments to the $u^{k}_{ij'r}$ variables.

\begin{table}[htbp]
\center
    \begin{tabular}{|r|}
    \hline

    \begin{minipage}{0.95\textwidth}
~\\
\textbf{Master(${\cal P},Pv,Pr,b,N$):}
{\small
\begingroup
\addtolength{\jot}{-2pt}
\begin{align}
\max \quad & \sum_{j \in {\cal D}} \sum_{m \in Pr_{j}} x_{jm}  + \frac{1}{|\xi^{D}|} \sum_{k=0}^{|\xi^{D}|}{\cal Q}(\{y_{im}\}_{i \in {\cal V}},(j \in {\cal P}, k)\\
s.t. \quad & \sum_{m \in Pr_{j}} x_{jm} \leq 1 ::: \forall j \in {\cal D} \label{cons:m11}\\
& \sum_{m \in Pv_{i}} y_{im} \leq 1 ::: \forall i \in {\cal V} \label{cons:m12}\\
& \sum_{j \in {\cal D}} x_{jm} \cdot b_{jm}^{n} \leq \sum_{i} y_{im} \cdot N(i,m,n) ::: \forall m, \forall n \label{cons:m13}
\end{align}
\endgroup }
\end{minipage} \\
    \hline
    \end{tabular}
    \caption{Optimization Formulation for Master problem - ZACBenders}
    \label{opt:master_zac}
    \vskip -10pt
    \end{table}

For the master (Table~\ref{opt:master_zac}), in the optimization provided in ZACFuture, we replace the part of the objective dealing with future variables, $\{u^{k}_{i'j'r}\}$ by the recourse function ${\cal Q}(\{y_{im}\}_{i \in {\cal V},(m \in {\cal P}}, k)$, which becomes the objective function in the slave problems. The recourse function ${\cal Q}()$ needs to be computed for each value of $y_{im}$. In the slaves (Table~\ref{opt:slave_zac}), we consider the fixed values of $y_{im}$ and to avoid confusion, we refer to them using the capital letter notation, $Y_{im}$. 

\begin{table}[htbp]
\center
    \begin{tabular}{|r|}
    \hline

    \begin{minipage}{0.95\textwidth}
~\\
\textbf{SlavePrimal(${\cal P},Pv,Pr,b,N,Y,k$):}
{\small
\begingroup
\addtolength{\jot}{-2pt}
\begin{align}
\max \quad &\frac{1}{|\xi^{D}|} \sum_{j' \in {\xi^{D,k}}}\sum_{r=0}^{\ceil{\frac{\eta_{j'}^{k}}{\kappa}}}\sum_{i'} w_{i'j'r}^{k} \cdot u_{i'j'r}^{k} \\
s.t. \quad & \sum_{j' \in {\xi^{D,k}}}\sum_{r=0}^{\ceil{\frac{\eta_{j'}^{k}}{\kappa}}} u_{i'j'r}^{k} \leq \sum_{i}\sum_{m} Y_{im}  ::: \forall i' \label{cons:s11} \\
& \sum_{i'} u_{i'j'r}^{k} \leq 1 ::: \forall j' \in \xi^{D,k}, 0 \leq  r < \ceil{\frac{\eta_{j'}^{k}}{\kappa}}  \label{cons:s12} 
\end{align}
\endgroup }
\end{minipage} \\
    \hline
    \end{tabular}
    \caption{Optimization Formulation for Slave problem (Primal)- ZACBenders}
    \label{opt:slave_zac}
    \vskip -10pt
    \end{table}

The dual~\cite{bertsimas1997introduction} of the primal slave problems are provided in Table~\ref{opt:slave_zacdual}, where $\alpha$ variables are the dual variables corresponding to the constraints \eqref{cons:s11} and $\beta$ variables are the dual variables corresponding to the constraints \eqref{cons:s12}. 
\begin{table}[htbp]
\center
   \begin{tabular}{|r|}
    \hline

    \begin{minipage}{0.95\textwidth}
~\\
\textbf{SlaveDual(${\cal P},Pv,Pr,b,N,Y,k$):}
{\small
\begingroup
\addtolength{\jot}{-2pt}
\begin{align}
\max \quad &\sum_{i'} \sum_{i} \sum_{m \in Pv_{i}}\alpha_{i'}^{k} \cdot Y_{im} + \sum_{j' \in {\xi^{D,k}}} \sum_{r} \beta_{j'r}^{k}\\
s.t. \quad & \alpha_{i'}^{k} + \beta_{j'r}^{k} \geq w_{i'j'r}^{k} ::: \forall i',j',r\\
 & \alpha_{i'}^{k} \geq 0 ::: \forall i',k \\
 & \beta_{j'r}^{k} \geq 0 ::: \forall j',r,k
\end{align}
\endgroup }
\end{minipage} \\
    \hline
    \end{tabular}
    \caption{Optimization Formulation for Slave problem (Dual) - ZACBenders}
    \label{opt:slave_zacdual}
    \vskip -10pt
    \end{table}
  
The {\em weak duality theorem}~\cite{bertsimas1997introduction} states that the solution to a maximization primal problem is always less than or equal to the solution of the corresponding dual problem. Therefore, using the concept of weak duality, we can say that, by taking the dual of the slave problems, we can find an upper bound on the value of the recourse function (${\cal Q()}$)(objective of primal slave problem), in terms of the master problem variables $y_{im}$. These can then be added as optimality cuts to the master problem~\cite{Murphy13} for generating better first stage assignments\footnote{As the slave problems are always feasible for any value of the master variables we only need to add optimality cuts to the master problem.}. 

Let $\theta^{k}$ be the approximation of ${\cal Q}$() function then the master problem with optimality cuts is provided in the Table~\ref{opt:master_zac_cuts}. 

\begin{figure}[htbp]
 \centering
 \frame{\label{fig:figbender}\includegraphics[width=0.99\textwidth,height=3.0in]{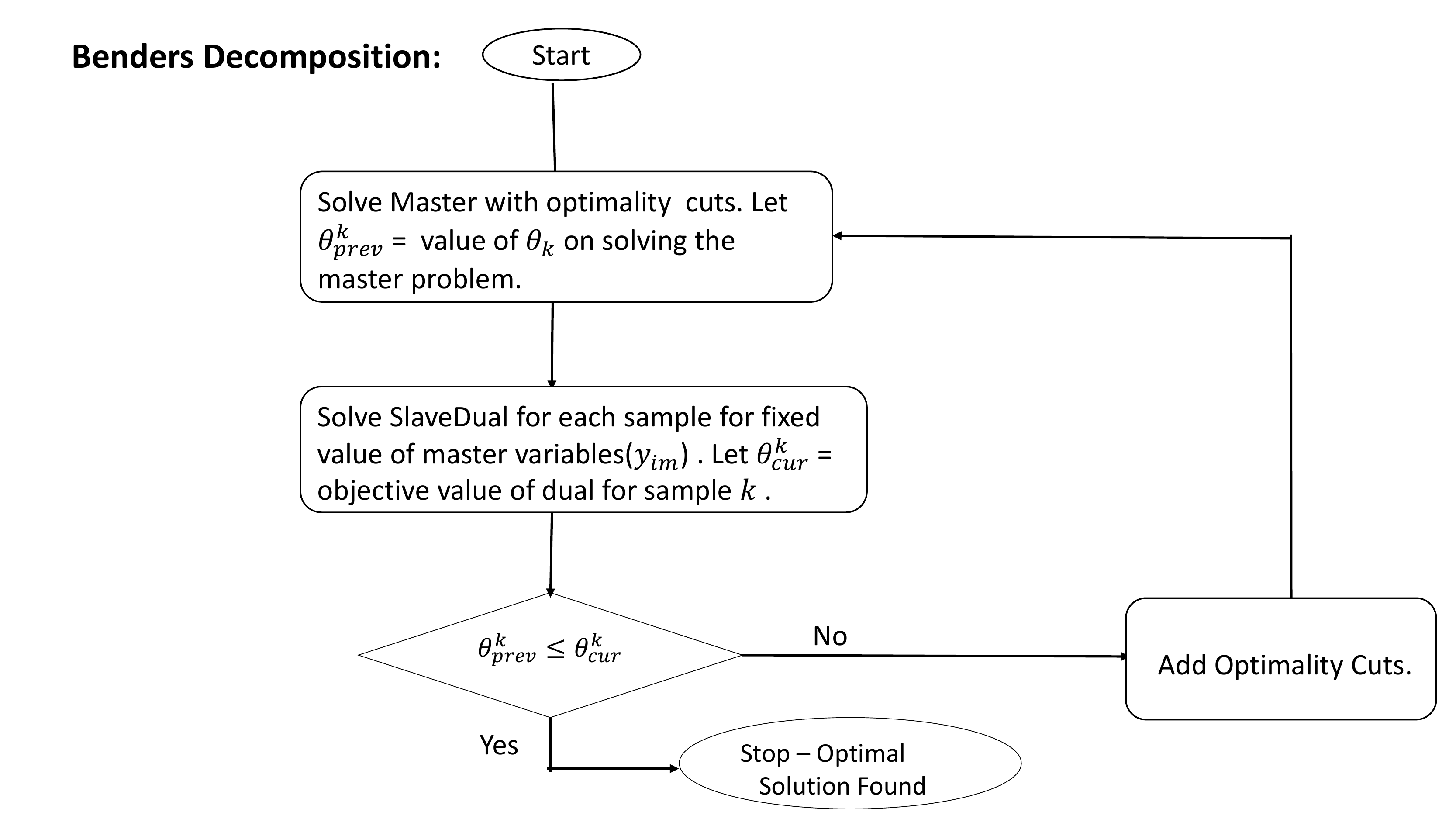}}
%\subfloat[]{\label{fig:figsp}\includegraphics[width=0.9\textwidth,height=1.6in]{maxflow.png}}
 \caption{ZACBenders Approach: Finding Optimal Assignment of requests to paths to vehicles}
 \label{fig:zacbender}
\end{figure}

It should be noted that we are using $y_{im}$ variables in the ``master with optimality cuts'' and not the fixed values, $Y_{im}$. 
%where \begin{align*}{\cal N}_{i}^{2} = {\cal N}_{i}^{1} -  \sum_{\mathclap{j\in \mathcal{D}}} x_{ij}^{1} +  \xi^S_i + \sum_{\mathclap{\substack{j\in \mathcal{D},d_{j}=s_i}}} \quad \sum_{m}  x_{i'j}^{1} 
%\end{align*}
In each iteration we solve the master problem and the computed $y_{im}$ variable values are passed to the dual slave problems. After solving the dual slave problems, optimality cuts are generated. If the current values of $\theta^{k} (\forall k)$ satisfy the optimality cut conditions, then we have obtained an optimal solution, else cuts are added to the master problem and the master problem is solved again. Figure \ref{fig:zacbender} shows the flow diagram for the same.

The slave problems are independent of each other (Table \ref{opt:slave_zacdual}) and are only connected by the choice of the master variables (``difficult'' integer variables). Therefore, once the master variables are fixed, the slave problems can be solved in a parallel fashion.

\begin{table}[htbp]
\center
    \begin{tabular}{|r|}
    \hline

    \begin{minipage}{0.95\textwidth}
~\\
\textbf{MasterWithOptimalityCuts(${\cal P},Pv,Pr,b,N$):}
{\small
\begingroup
\addtolength{\jot}{-2pt}
\begin{align}
\max \quad & \sum_{j \in {\cal D}} \sum_{m \in Pr_{j}} x_{jm}  + \frac{1}{|\xi^{D}|} \sum_{k} \theta^{k} \\
s.t. \quad & \theta^{k} \leq   \sum_{i'} \sum_{i} \sum_{m \in Pv_{i}}\alpha_{i'}^{k} \cdot y_{im} + \sum_{j' \in {\xi^{D,k}}} \sum_{r} \beta_{j'r}^{k}\\
& \sum_{m \in Pr_{j}} x_{jm} \leq 1 ::: \forall j \in {\cal D} \label{cons:mo11}\\
& \sum_{m \in Pv_{i}} y_{im} \leq 1 ::: \forall i \in {\cal V} \label{cons:mo12}\\
& \sum_{j \in {\cal D}} x_{jm} \cdot b_{jm}^{n} \leq \sum_{i} y_{im} \cdot N(i,m,n) ::: \forall m, \forall n \label{cons:mo13}
\end{align}
\endgroup }
\end{minipage} \\
    \hline
    \end{tabular}
    \caption{Optimization Formulation for Master problem (with optimality cuts) - ZACBenders}
    \label{opt:master_zac_cuts}
    \vskip -10pt
    \end{table}

\section{Experiments}
\label{sect:experiments}
The goal of the experiments is to evaluate the performance of ZAC and ZACBenders in comparison to TBF~\footnote{The complexity of TBF increases with the increase in vehicle capacity. It is not possible to run it up to optimality. Therefore, we run it with the heuristics mentioned in the paper (0.2 second for each vehicle and keeping 30 vehicles for each request (but keeping all request edges)). We use the objective of maximizing the number of requests served for all algorithms. The objective can be changed to the objective of minimizing the delay or maximizing the revenue for both TBF and ZAC.}. We also compare the performance of our algorithms against NeurADP~\cite{neuradp}, which learns the expected future value of each assignment of vehicle to trips by using a neural network based value function in the approximate dynamic programming framework. As NeurADP requires training a different model for each change in an input parameter, using limited academic resources, it was not possible to run the exhaustive set of experiments with NeurADP. Therefore, we first show the detailed experimental results comparing the performance of TBF, ZAC and ZACBenders and then in section~\ref{sect:compareneuradp} we compare against NeurADP. For ZACBenders we kept a maximum timelimit of $\Delta$ seconds for each assignment, but the Benders Decomposition can converge before the maximum timelimit is reached. 

We evaluate the algorithms on following metrics: (1) Service Rate, i.e., percentage of total available requests served. (2) Runtime to compute a single step assignment. We experimented by taking demand distribution from two real-world and one synthetic dataset.

Table \ref{table:expoutline} provides the outline for this section. We will show two main results that demonstrate the significant utility of our approaches:
\squishlist
\item Our myopic approach ZAC outperforms the current best myopic approach TBF. While the improvement varies, ZAC serves up to 4\% more requests on real-world datasets and up to 20\% more requests on synthetic dataset. 
\item Our non-myopic approach ZACBenders further improves the performance of ZAC. It provides 14.7\% improvement over TBF. 
NeurADP when hyper-optimized to the test settings can perform better than ZACBenders on certain cases. However, ZACBenders gets improvements of up to 12.48\% when NeurADP is not hyper-optimized for test settings.
\squishend 

\begin{table}[h]
{%\small
\begin{center}
\begin{tabular}{|l|l|l|}
%{|p{0.5in}|p{0.5in}|p{0.25in}|p{0.55in}|p{0.70in}|}
\hline
\textbf{Section} & \textbf{Description} & \textbf{Key Content} \\
\hline
\ref{sect:datasets} & Datasets & Details on the datasets and different data\\
&~& fields used from the datasets. \\
\hline
{\ref{sect:expsettings}} & {Experimental Settings} & {Details on the different inputs, parameters}\\
&~& and evaluation settings used. \\
\hline
{\ref{sect:expres}} & {Results on Real-World Datasets} & {Describes our key results on real-world}\\
&~& datasets and shows the comparison of\\
& ~& TBF, ZAC and ZACBenders for different\\
&~& parameters at on the three metrics of service\\
&~& rate and runtime. \\
\hline 
{\ref{sect:compareneuradp}} & {Comparison with NeurADP} & {Describes our key result by comparing TBF,}\\
&~& ZAC and ZACBenders with NeurADP.\\
\hline
{\ref{sect:synthetic}} & {Results on Synthetic Dataset} & {Describes the performance of algorithms on}\\
&~& specially created first and last mile scenarios \\
&~&~where it is advantageous to explore more\\
&~& request combinations at a decision epoch.\\
\hline
\end{tabular}
\end{center}
}
\caption{Experiment Section Outline}
\label{table:expoutline}
\vskip -10pt
\end{table} 
\subsection{Datasets}
\label{sect:datasets}
The first real-world dataset is the publicly available New York Yellow Taxi Dataset~\cite{yellowtaxi}, henceforth referred to as the NYDataset. The name of the other real-world dataset can not be revealed due to confidentiality agreements. It is referred to as Dataset1. We use the street intersections as the set of locations ${\cal L}$. To find out the street intersections in real-world dataset, we take the street network of the city from openstreetmap using osmnx with drive network type ~\cite{boeing2017osmnx}. From these we remove the network nodes that do not have any outgoing edges, i.e., we take the largest strongly connected component of the network. For NYDataset, as considered in earlier works~\cite{alonso2017demand}, we only consider the street network of Manhattan as 75\% of the requests have pick-up and drop-off locations in Manhattan. Moreover, less than 15\% of the total requests have pick-up and drop-off location in different boroughs of New York indicating that these boroughs can be solved independently. 

Both real-world datasets contain data of past customer requests for taxis at different time of the day and for different days of the week. From these datasets, we take the following fields: (1) Pick-up and drop-off locations (latitude and longitude coordinates) - These locations are mapped to the nearest street intersection. (2) Pick-up time - This time is converted to appropriate decision epoch based on the value of $\Delta$. The travel time on each road segment of the street network is taken as the daily mean travel time estimate computed using the method proposed in ~\cite{santi2014quantifying}. 
\begin{figure}[htbp]
 \centering
\subfloat[]{\label{fig:figsyn1}\includegraphics[width=0.45\textwidth,height=2.5in]{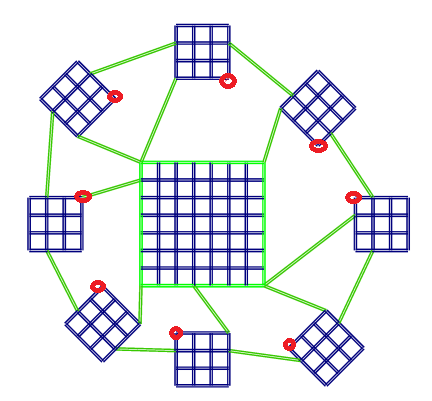}}
 \caption{Street network for synthetic dataset. Train stations are marked with red.}
 \label{fig:figsynthetic}
\end{figure}

\begin{table}[h]
{%\small
\begin{center}
\begin{tabular}{|l|l|l|l|l|}
%{|p{0.5in}|p{0.5in}|p{0.25in}|p{0.55in}|p{0.70in}|}
\hline
\textbf{Dataset} & \textbf{Locations} & \textbf{Edges} & \textbf{Avg No. of Requests} & \textbf{Avg No. of Requests}\\
~ & \textbf{($|{\cal L}|$)} & \textbf{$(|{\cal E}|)$} & \textbf{per day (on test days)}& \textbf{per hour (Peak)}\\
~ &  &  & \textbf{} & \textbf{(on test days)}\\
\hline
NYDataset & 4373 & 9540 & 313683 & 20910\\
\hline
Dataset1 & 21212 & 41424 & 403770 & 23664\\
\hline
Synthetic & 192 & 640 & 173557 & 8578\\
\hline
\end{tabular}
\end{center}
}
\caption{Details for different datasets}
\label{table:dataset}
\vskip -10pt
\end{table} 

To simulate the scenario for on demand shuttle services~\cite{shotl,beeline,grabshuttle} having a small set of pick-up/drop-off points in a city, we also perform experiments on a synthetic dataset introduced by Bertsimas {\em et al.}~\cite{bertsimas2018online}. The network (Figure \ref{fig:figsynthetic}) has one downtown area represented by the big square in center and 8 suburbs. We create a train station at one node of each suburb (marked by red circle) to simulate special cases of first and last mile transportation. At each decision epoch, requests are randomly generated by taking pick-up and drop-off location uniformly. In addition, every 180 seconds (frequency of arrival of train at the train stations), we generate first and last mile requests in each suburb (representing arrivals by train). 

The number of nodes/locations, edges in the street network of the city and the number of requests present in each dataset are shown in the Table \ref{table:dataset}.

%$M$ & 2,4,6 \\
%\hline
%Clustering Method & GBC, HAC\_MAX,HAC\_AVG \\
%\hline
\subsection{Experimental Settings}
\label{sect:expsettings}
There are three different categories of experimental settings that have an impact on the performance of algorithms

\begin{enumerate}
\item \textbf{Inputs provided to all algorithms:} These include
\squishlist
\item Number of Vehicles ($|{\cal V}|$): The number of vehicles used is dependent on the fleet size of the company. At the start of the experiment, empty vehicles are distributed uniformly at random in different locations. Based on the assignment obtained by algorithms at any decision epoch, the status of vehicles at the next decision epoch is updated. In the results section, we vary the number of vehicles to show the performance of algorithms for different number of vehicles. 
\item Maximum Capacity ($\kappa$): The maximum number of passengers that can be present in a vehicle at any time. 
%In the end, we also show the experiments by having a distribution over maximum capacity of vehicles. 
\item $\tau$ and $\lambda$: $\tau$ represents the maximum time within which the vehicle should reach the origin location of request and $\lambda$ denotes the maximum allowed travel delay for any request (in seconds). 
\item Decision epoch duration ($\Delta$): This parameter determines how often the algorithm should be executed, and assignment decisions are made. For example, if $\Delta=60$ seconds, then requests are batched for the duration of 60 seconds and the decision of serving or rejecting these requests is taken every 60 seconds by the algorithm. We vary this parameter to show the performance of algorithms for different values.
\squishend
Table \ref{table:inp_setting} shows the values of different input parameters considered in the experiments. 

\begin{table}[h]
{%\small
\begin{center}
\begin{tabular}{|l|l|}
\hline
\textbf{Input Parameter} & \textbf{Values considered in Experiments}\\
\hline
$\Delta$ (in seconds) & 10,30,60 \\
\hline
$\tau$ (in seconds) & 120,180,300,420\\
\hline
$\lambda$ (in seconds) & 240,600,840,900\\
\hline
$|{\cal V}|$ & 1000,2000,3000,5000,8000,10000 \\
\hline
$\kappa$ & 1,2,3,4,8,10 \\
\hline
\end{tabular}
\end{center}
}
\caption{Inputs to all algorithms}
\label{table:inp_setting}
\vskip -10pt
\end{table} 
\item \textbf{Parameters of the algorithm:} The parameters required by our algorithms are:
\squishlist
\item Clustering Method: To construct zones from the set of locations ${\cal L}$, we compare the performance by using different clustering methods and different static zone sizes. Zone size is taken as the intra zone travel time (in seconds).    
\item Number of Different Zone Sizes (for drop-off locations) ($M$): In the online completion phase of the offline path, instead of using a fixed zone size, ZAC dynamically decides the zone size to be used from a predefined fixed set of zone sizes. We vary the number of different zone sizes from which the ZAC algorithm picks the best zone size for a path. 
 %The value of Q is taken such that  ∗ (Q +1) = 30 minutes, i.e., if  is 1, Q is taken as 29 and
%if  is 15, Q is taken as 1. We choose a fix value of 30 minutes because more 
\item Zone Size for Samples (${\cal Z}^{s}$): For samples we use a static zone size of 600 seconds. While it is possible to improve the performance of ZACBenders by using different zone size for different capacities and different value of $\tau$ and $\delta$, we observe in the experiments that, by using a fixed zone size, it is possible to get improvement across different parameters and different datasets.  
\item Number of Samples ($|\xi^{D}|$): While computing an assignment at decision epoch $e$, our non-myopic approach ZACBenders require samples of customer requests at decision epochs $e + 1,e + 2,..,e + Q$, where Q = $\floor{\frac{\rho}{\Delta}}$, from past data (at the same decision epoch on the past days). We identify the right value for the number of samples through experiments as described in results section. Each sample correspond to past one day. For example, if 10 samples are used, it means that requests from past 10 days are used to compute the expected future value in ZACBenders. 
\item LookAhead Duration ($\rho$): This determines how far ahead ZACBenders look into the future. If look ahead duration is 600 seconds and current time is 09:00AM, ZACBenders considers samples of customer requests up to 09:10AM. 
\squishend

Table \ref{table:param_settings} shows the different values for the parameters used in the experiments. To obtain the right set of parameter values, we compare the performance of approaches by running them on 5 different weekdays from 21-03-2016 to 25-03-2016 and taking the average value over these five days. 
\begin{table}[h]
{%\small
\begin{center}
\begin{tabular}{|l|l|l|}
\hline
\textbf{Algorithm Parameter} & \textbf{Values considered in Experiments}\\
\hline
$M$ & 2,4,6 \\
\hline
Clustering Method & GBC, HAC\_MAX,HAC\_AVG \\
\hline
Number of Samples ($|\xi^{D}|$) & 1,3,5,8,10\\
\hline
Look Ahead Duration (in seconds) ($\rho$) & 600, 900 ,1200 ,1500\\
\hline
Zone size for Samples ($Z^{s}$) (in seconds) & 600 \\
\hline
\end{tabular}
\end{center}
}
\caption{Algorithm Parameter Settings}
\label{table:param_settings}
\vskip -10pt
\end{table} 
\item \textbf{Evaluation Settings:}
\squishlist
\item Evaluation duration: We evaluate the performance of algorithm over 1 hour by varying different input and algorithmic parameters. For a subset of parameter combinations, we also compared the performance over 24 hours~\footnote{As running all the algorithms for 24 hours over different set of parameters takes a long time and the difference in the performance of algorithms over 1 hour was following a similar trend as over 24 hours, we ran it for 24 hours only for a subset of parameters.}. 
\item Number of days and time of evaluation: We performed experiments with requests at various times of the day, 8:00 AM, 3:00 PM, 6:00 PM, 12:00AM and on different days. We evaluated the approaches by running them on 15 different weekdays between 04-04-2016 and 22-04-2016 and taking the average values over 15 days. These 15 days are different from the 5 days used to obtain the right set of algorithm parameters. 
%\begin{table}[h]
%{%\small
%\begin{center}
%\begin{tabular}{|l|l|l|}
%\hline
%\textbf{Evaluation Parameter} & \textbf{Values considered in Experiments}\\
%\hline
%Time of the day & 08:00AM,03:00PM,06:00PM,12:00AM \\
%\hline 
%Duration & 1 hour, 24 hours\\
%\hline
%\end{tabular}
%\end{center}
%}
%\caption{Evaluation Settings}
%\label{table:eval_settings}
%\vskip -10pt
%\end{table}

\squishend
\end{enumerate}

We conducted experiments with all the combinations of settings and inputs mentioned in this section. To avoid repeating similar results over and over again, we provide the representative results. All experiments are run on 24 core - 2.4GHz Intel Xeon E5-2650 processor and 256GB RAM. The algorithms are implemented in Java and optimization models are solved using CPLEX 12.6. 
For ZAC and ZACBenders, we also use offline generated paths. Table \ref{table:offlinepaths} provides the number of paths and memory requirement of the offline paths for different values of $\tau$ for NYDataset. We generated all possible paths when $\tau \leq 180$ seconds. For higher values of $\tau$, we used the data-driven approach provided in Appendix~\ref{appendix:offline}

\begin{table}[h]
{%\small
\begin{center}
\begin{tabular}{|l|l|l|l|}
\hline
\textbf{$\tau$} & \textbf{Number of Paths} & \textbf{Memory used}\\
\hline
120 & 208604  & 76 MB\\
\hline
180 & 2008895 &1.68 GB\\
\hline
300 & 3196478 & 3.98 GB\\
\hline
420 & 4256412 & 7.86 GB\\
\hline
\end{tabular}
\end{center}
}
\caption{Offline Paths}
\label{table:offlinepaths}
\vskip -10pt
\end{table}
%For the experimental analysis, we considered all vehicles have identical capacities.

\subsection{Results on Real-World Datasets}
\label{sect:expres}
In this section, we compare the number of requests served by TBF, ZAC and ZACBenders. We also compare the average time taken to compute an assignment by all the approaches. 

We choose the best configuration for parameters of algorithms (justification provided in the Section \ref{sect:justifyparam}). For ZAC and ZACBenders, we cluster locations into zones using HAC\_MAX and use $M$=4 (with zone sizes 0,60,120,300). ZACBenders uses 5 samples with a look ahead duration of 15 minutes and the value of $Z^{s}$ (zone size used in second stage) is taken as 600 seconds.

We first compare the service rate and runtime of TBF, ZAC and ZACBenders by varying different parameters on two real-world datasets.
\begin{figure*}[htbp]
 \centering
  \subfloat{\label{fig:fignyservrate1000}\includegraphics[width=0.33\textwidth,height=1.4in]{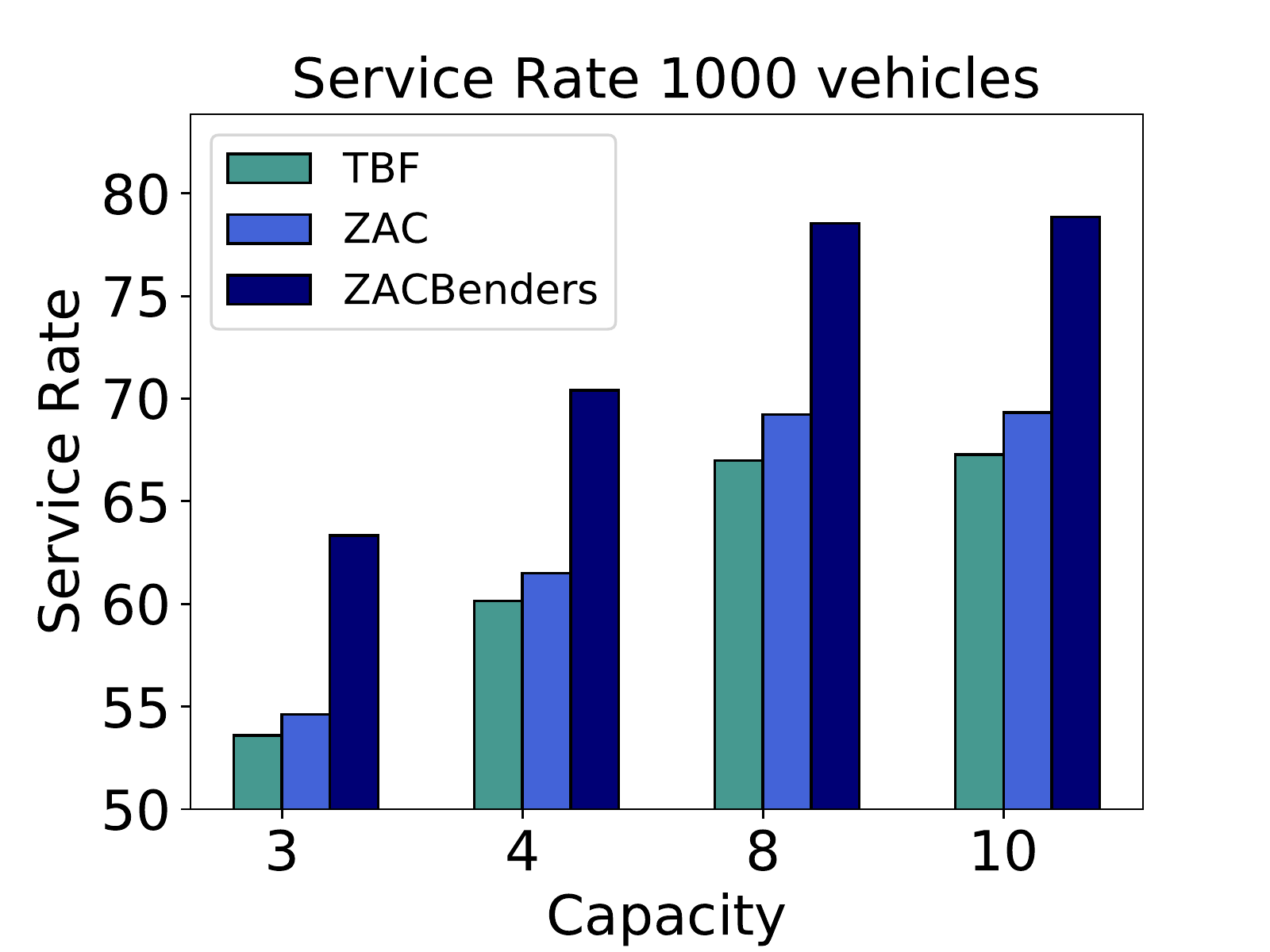}}
  \subfloat{\label{fig:fignyservrate3000}\includegraphics[width=0.33\textwidth,height=1.4in]{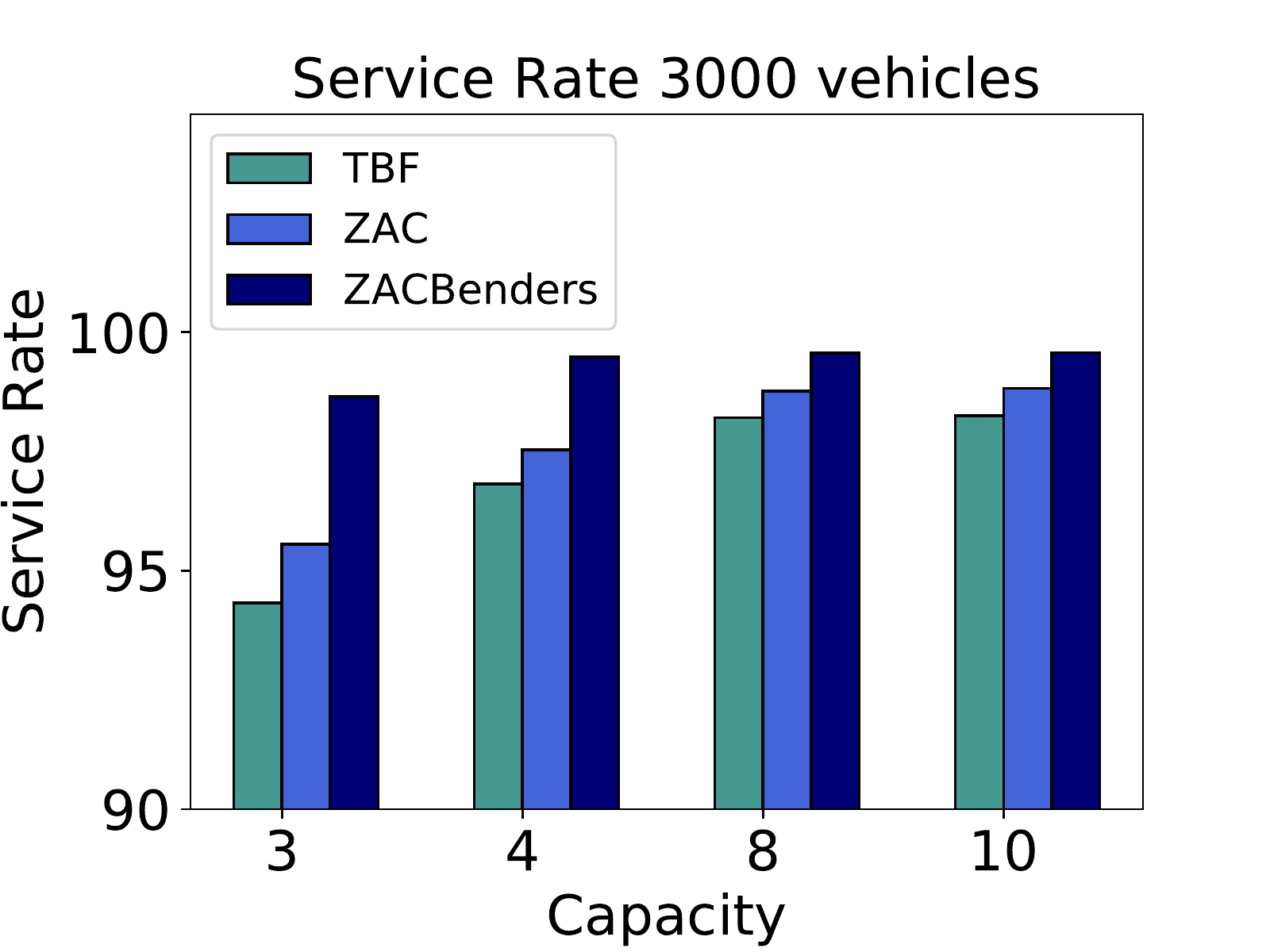}}
  \subfloat[]{\label{fig:fignyservratevcapdiff}\includegraphics[width=0.33\textwidth,height=1.4in]{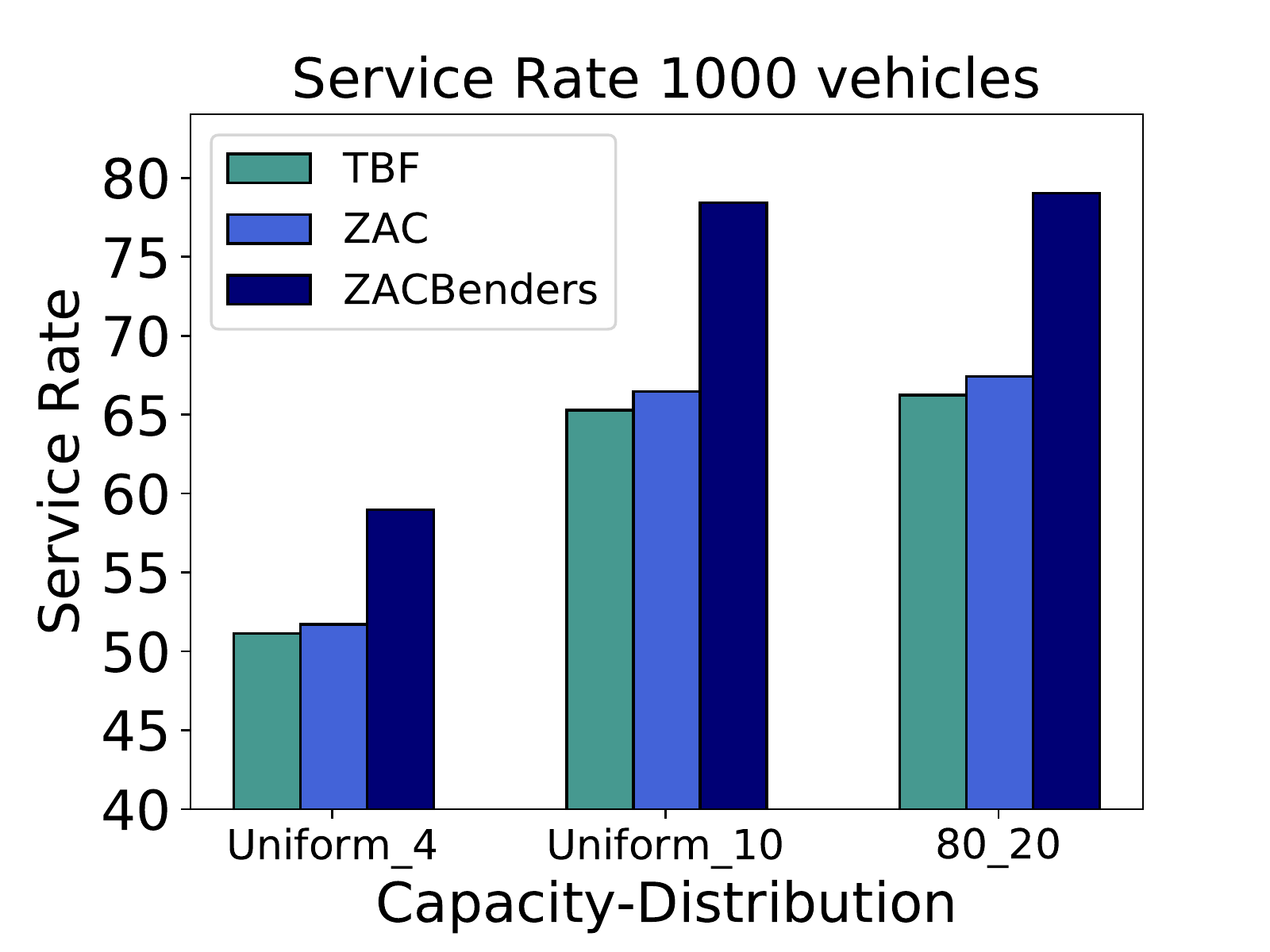}}\\
  \subfloat{\label{fig:fignyrtime1000}\includegraphics[width=0.33\textwidth,height=1.4in]{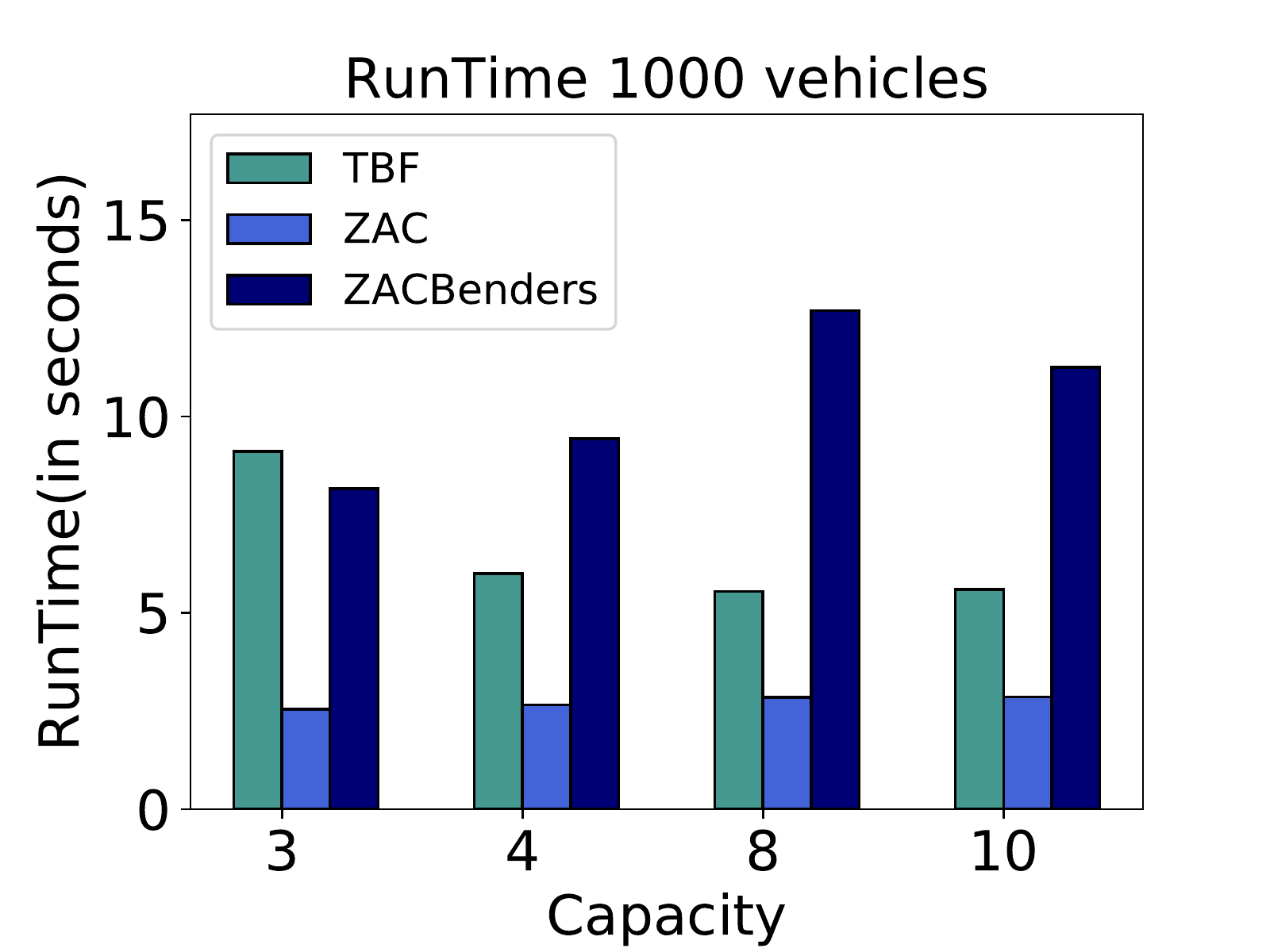}}
 % \subfloat[]{\label{fig:fignyrtime2000}\includegraphics[width=0.16\textwidth,height=1.0in]{images/2000_rtime_3min_10min_ny.pdf}}
  \subfloat{\label{fig:fignyrtime3000}\includegraphics[width=0.33\textwidth,height=1.4in]{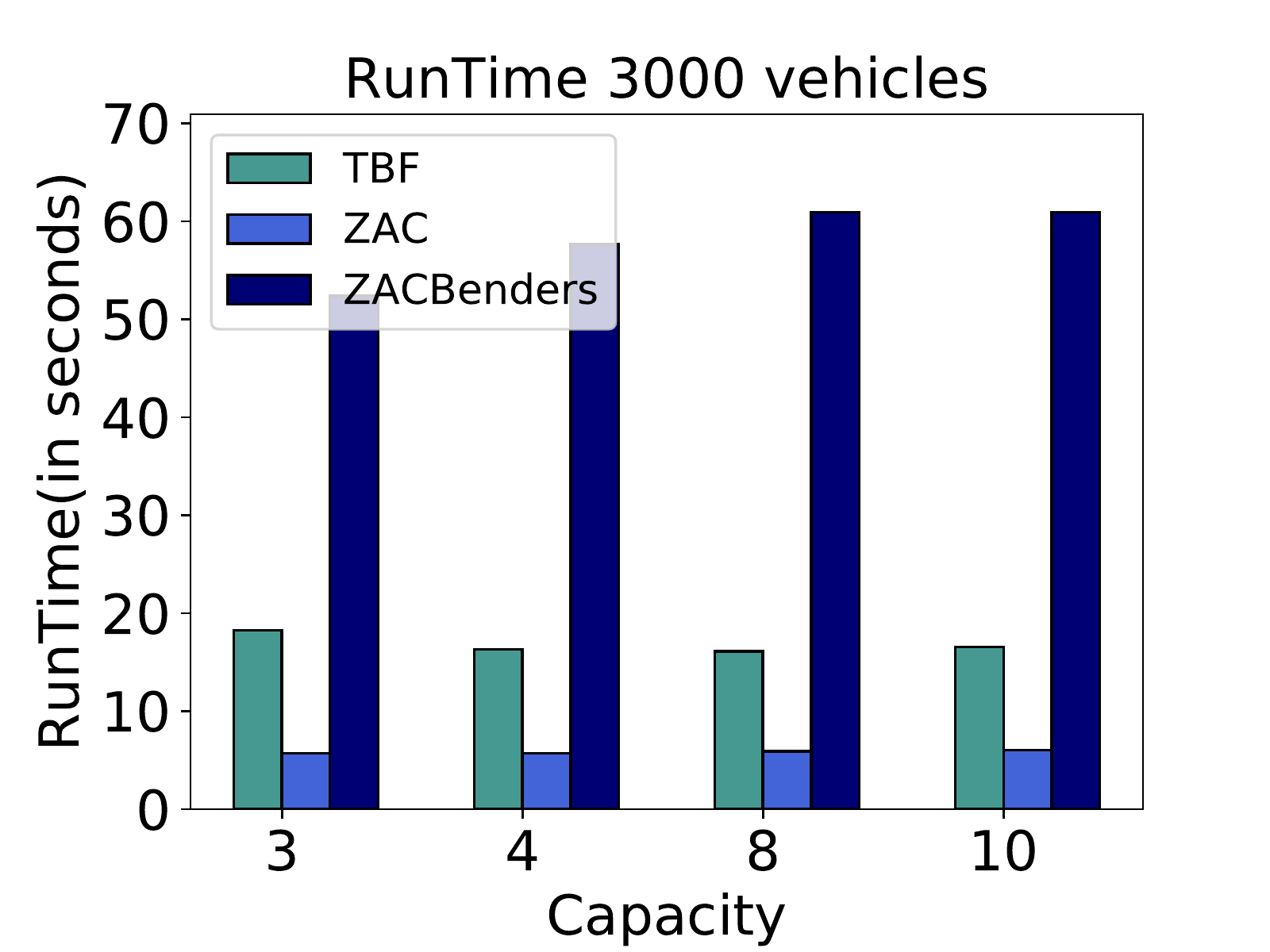}}
 \caption{{Comparison of ZACBenders, ZAC and TBF on NYDataset for $\tau$=180 seconds, $\lambda$=600 seconds and $\Delta=60$ seconds }}
 \label{fig:figny_3min}
\end{figure*}

\begin{figure*}[htbp]
 \centering
%  \subfloat[]{\label{fig:figsgservrate2}\includegraphics[width=0.16\textwidth,height=1.0in]{images/cap_2_servrate_3min_10min_sg.pdf}}
\subfloat{\label{fig:figsgservrate1000}
  \includegraphics[width=0.33\textwidth,height=1.4in]{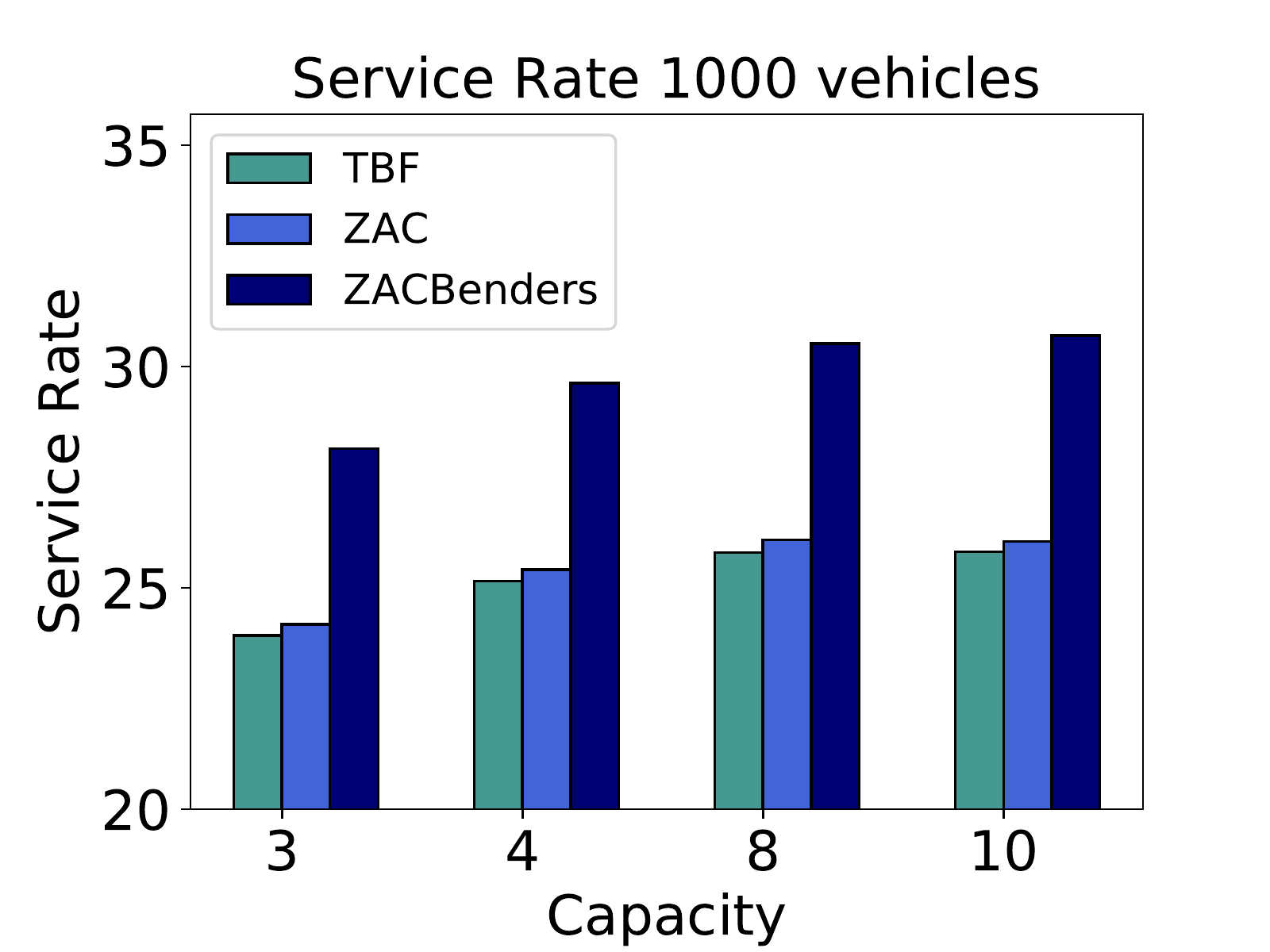}}
  \subfloat{\label{fig:figsgservrate3000}
  \includegraphics[width=0.33\textwidth,height=1.4in]{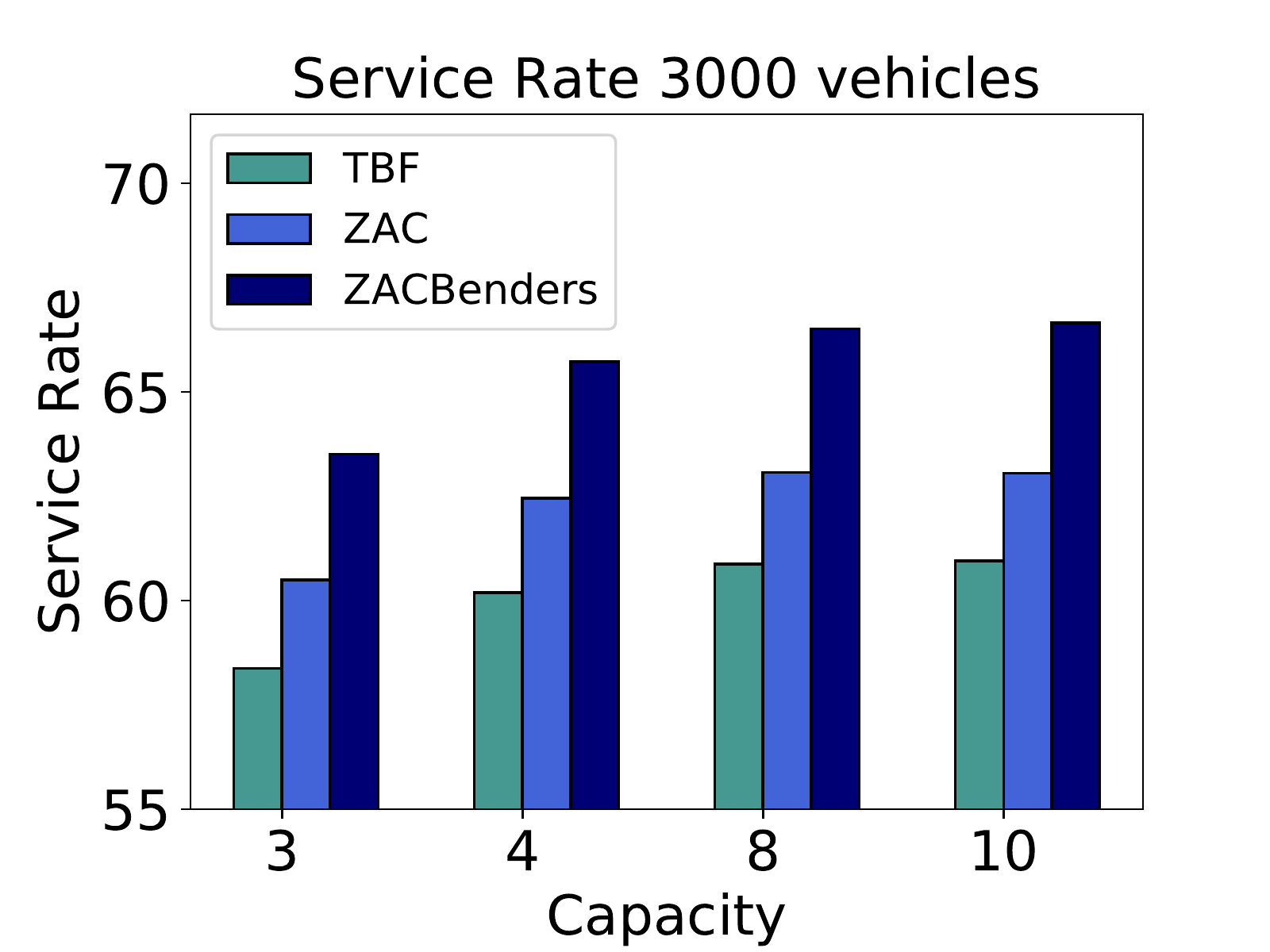}}
  \subfloat{\label{fig:figsgservrate5000}\includegraphics[width=0.33\textwidth,height=1.4in]{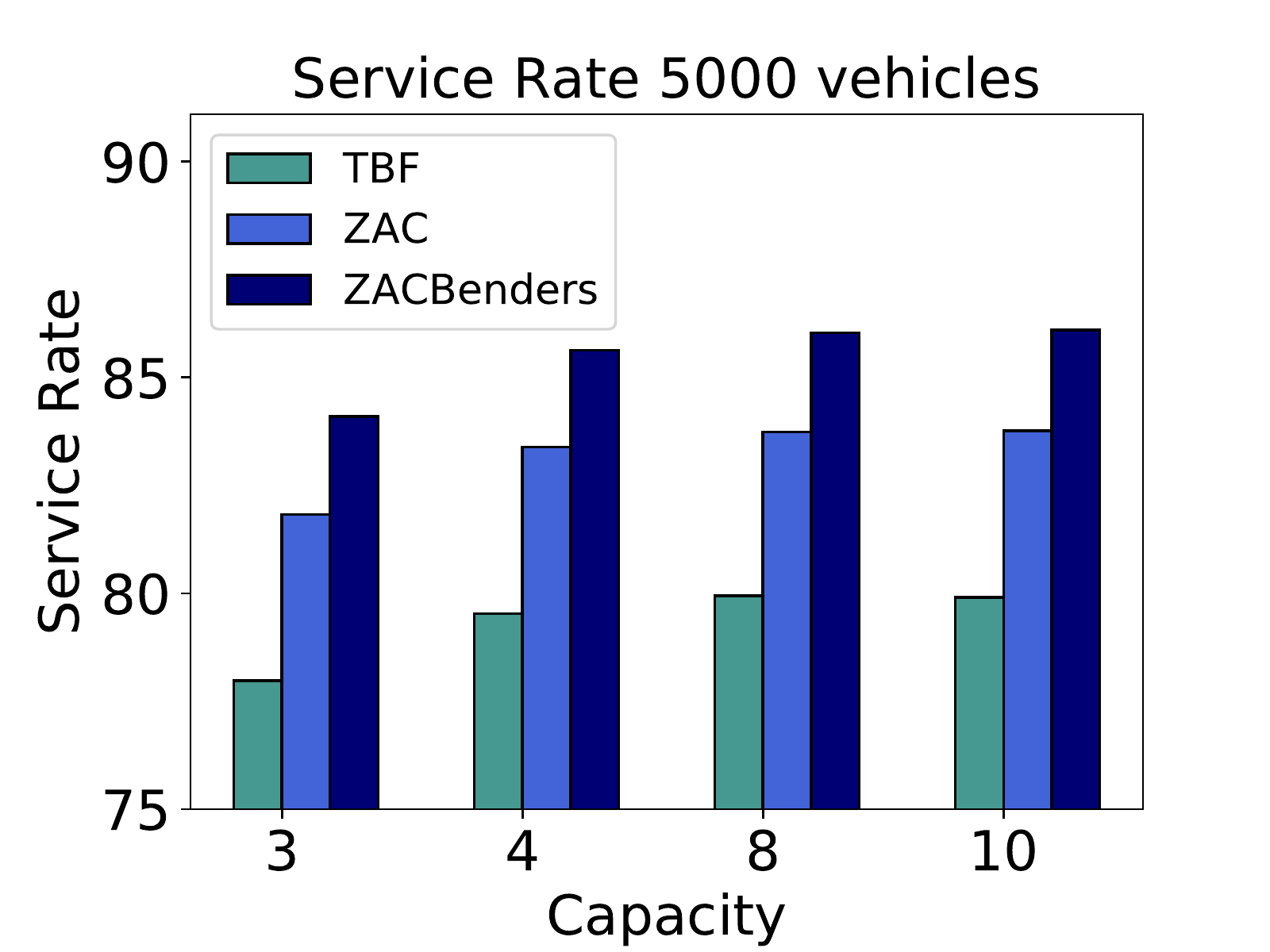}}\\
 % \subfloat[]{\label{fig:figsgrtime2}\includegraphics[width=0.16\textwidth,height=1.0in]{images/cap_2_rtime_3min_10min_sg.pdf}}
  \subfloat{\label{fig:figsgservrate10000}\includegraphics[width=0.33\textwidth,height=1.4in]{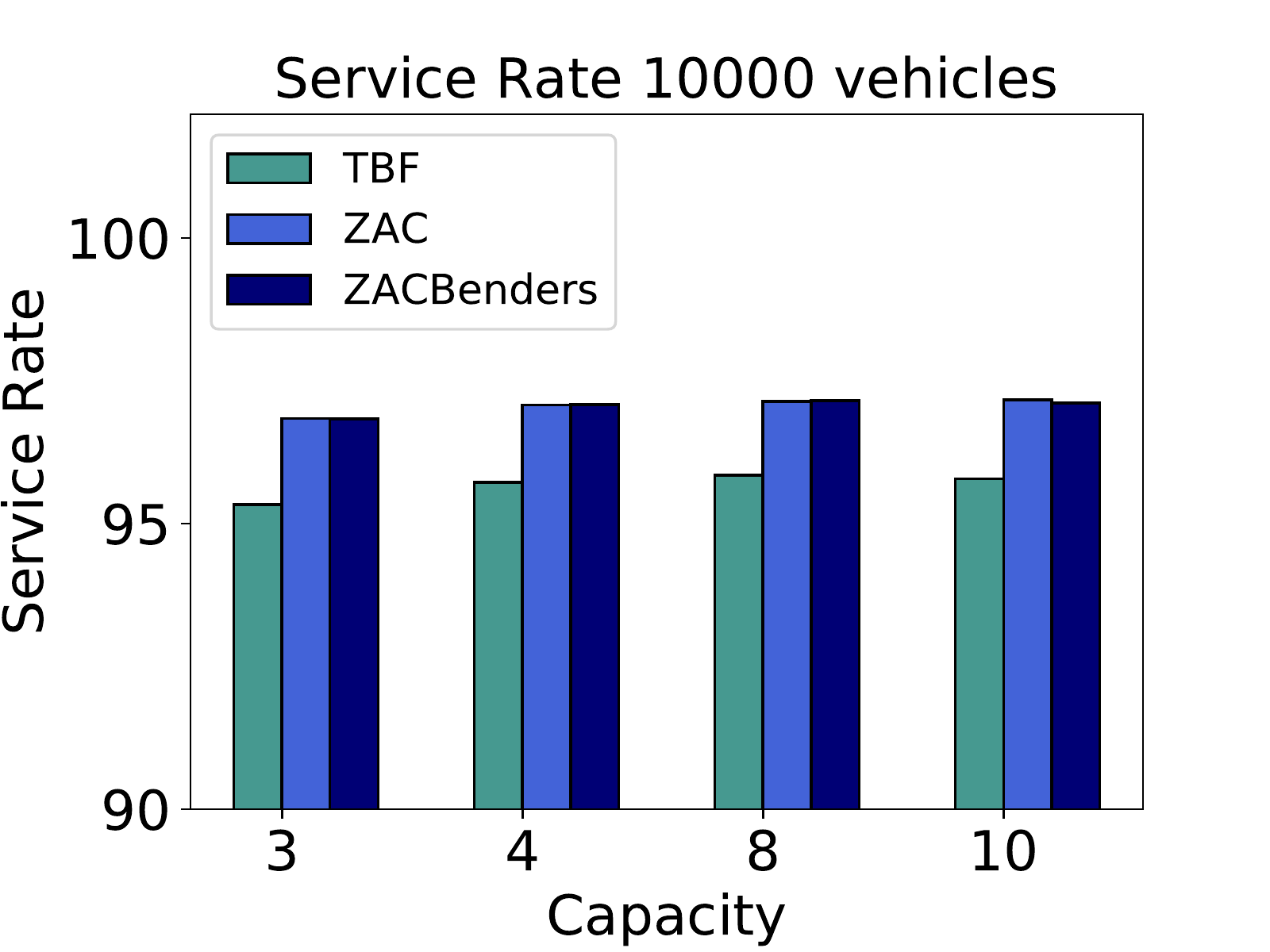}}
  \subfloat{\label{fig:figsgrtime5000}\includegraphics[width=0.33\textwidth,height=1.4in]{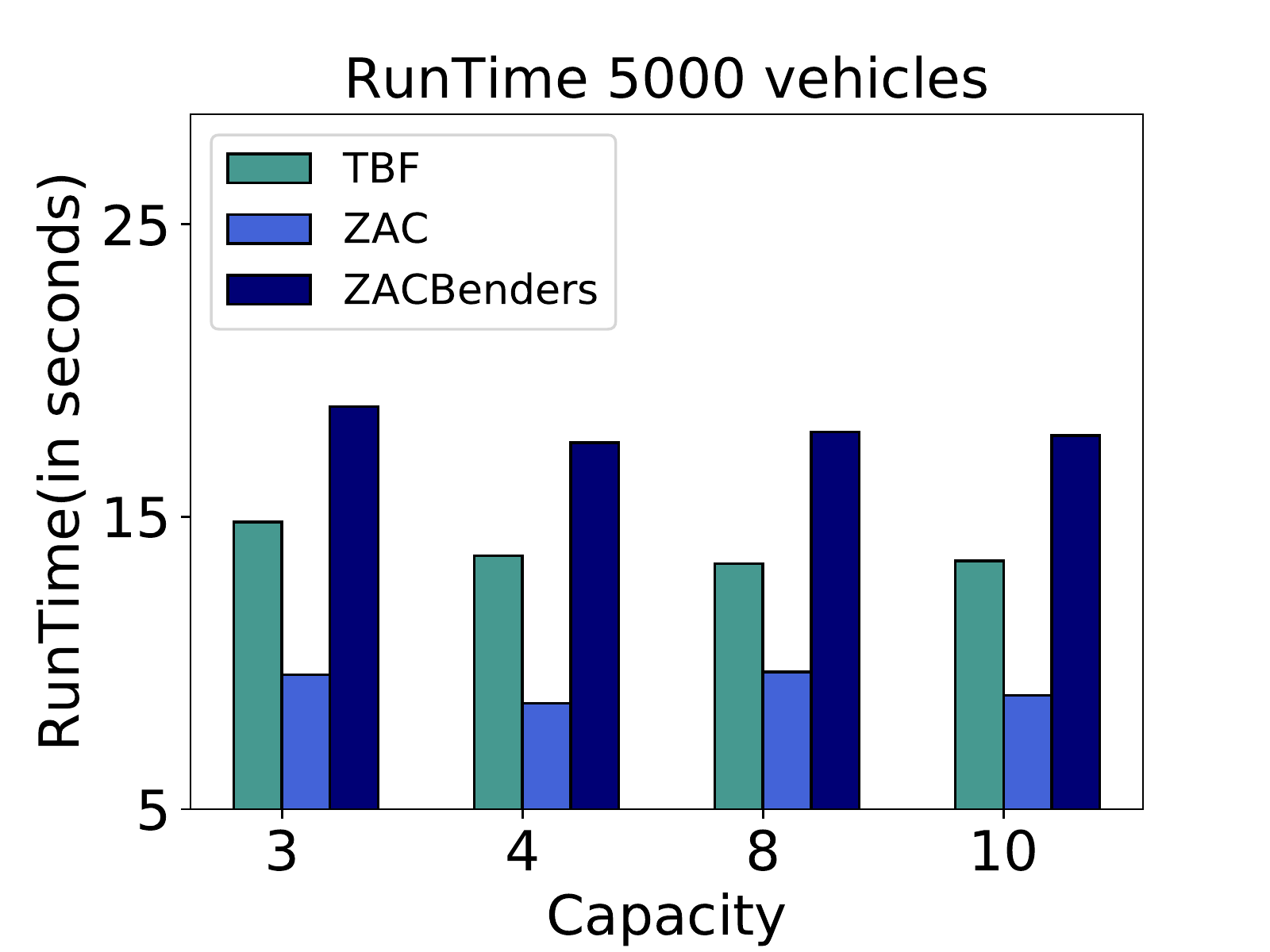}}
    \subfloat{\label{fig:figsgrtime10000}\includegraphics[width=0.33\textwidth,height=1.4in]{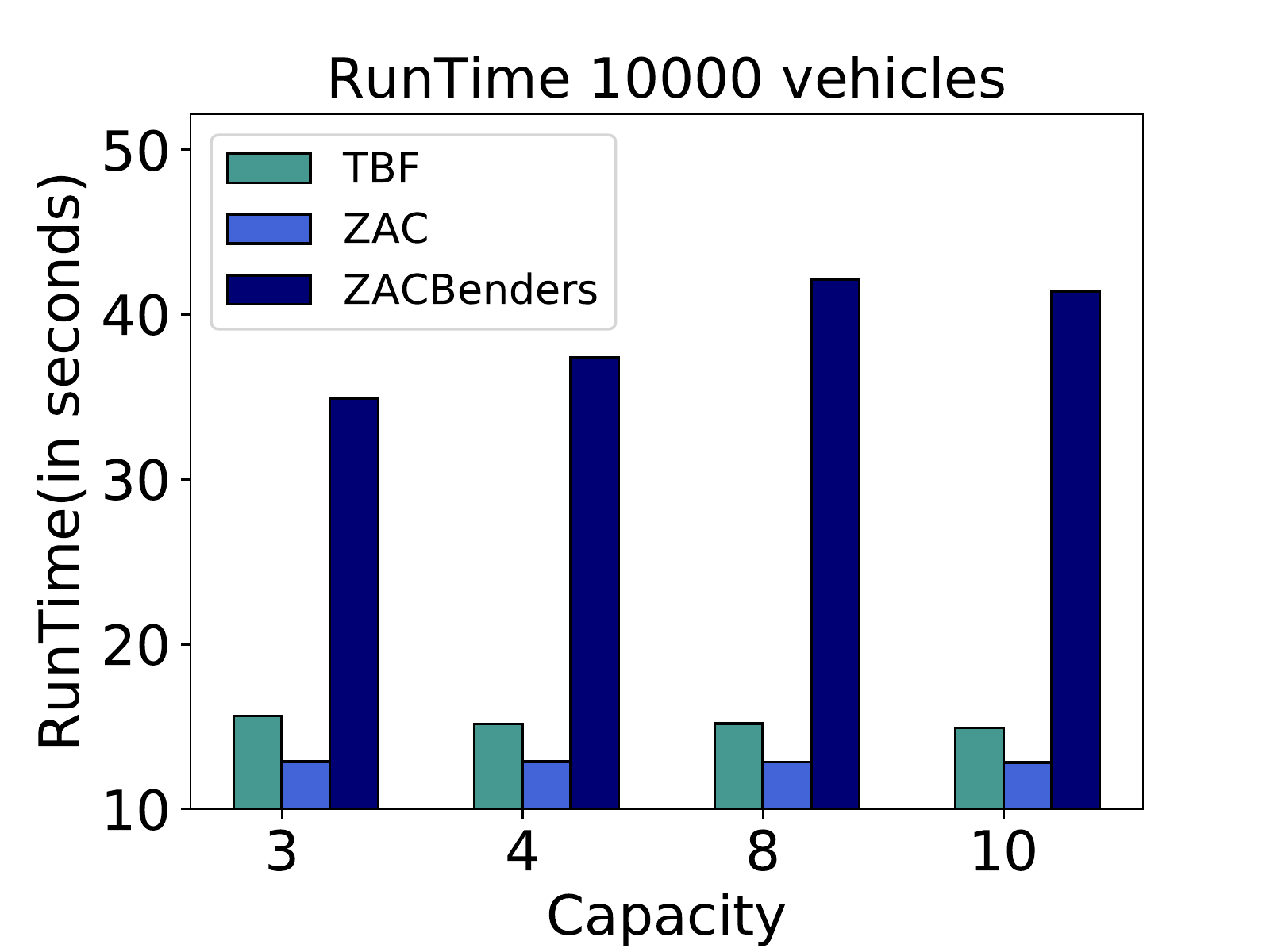}}
%    \subfloat{\label{fig:figsgrtime10000}\includegraphics[width=0.24\textwidth,height=1.0in]{images/10000_rtime_3min_10min_sg.pdf}}
 \caption{{ Comparison of ZACBenders, ZAC and TBF on Dataset1 for $\tau =180$ seconds, $\lambda = 600$ seconds and $\Delta$= 60 seconds }}
 \label{fig:figsg_3min_cap}
\end{figure*}

\noindent \textbf{Effect of change in vehicle capacity ($\kappa$) and number of vehicles ($|V|$):} Figure \ref{fig:figny_3min} and Figure \ref{fig:figsg_3min_cap} show the service rate and runtime comparison of TBF, ZAC and ZACBenders for NYDataset and Dataset1 respectively at 8am (Peak time) . 

%The time taken by TBF is less on Dataset1 as compared to NYDataset as the number of possible request combinations, at each decision epoch, are more for NYDataset. 

\noindent For the change in the number of vehicles, we make the following observations:
\squishlist
\item On Dataset1, the difference in the service rate obtained by ZAC and TBF increases as the number of vehicles increases from 3000 to 5000. One of the reasons is that TBF limits the number of vehicles considered for each request to 30, so the number of requests missed due to this limit will be more for higher number of vehicles. But on further increasing the number of vehicles to 10000, the gap between ZAC and TBF reduces. This is because, when more vehicles are available, it reduces the need of generating all combinations. On NYDataset, the difference between service rate obtained by ZAC and TBF is maximum for 1000 vehicles.
\item The difference in service rate of ZACBenders and ZAC decreases as the number of vehicles are increased. This is because when more vehicles are available, they will be free even after executing current assignments at the current decision epoch, so future demands can be met irrespective of the current assignment. On Dataset1 for capacity 4, ZACBenders obtains 4.2\% improvement over ZAC for 1000 vehicles, 3.27\% improvement for 3000 vehicles and 2.24\% improvement for 5000 vehicles. For 10000 vehicles, the service rate obtained by ZAC and ZACBenders is almost the same. On NYDataset for capacity 4, the maximum improvement obtained by ZACBenders over ZAC is 8.89\% which is for 1000 vehicles. 
\squishend

\noindent Here are the key observations when vehicle capacity is changed for a fixed number of vehicles:
\squishlist 
\item Service rate obtained by ZAC is more than TBF for both datasets. For capacity 4 with 1000 vehicles for NYDataset, the service rate obtained by ZAC is 1.36\% more than the service rate obtained by TBF and for capacity 10 we obtain a gain of 2.03\%. On the other hand, for Dataset1 for capacity 4 with 5000 vehicles, the service rate obtained by ZAC is up to 4\% more than the service rate obtained by TBF. On Dataset1, we do not observe much increase in service rate beyond capacity 4 due to large size of the network and, longer travel times which allow fewer requests to be paired. 
\item ZACBenders improves the performance of ZAC by using future information. For capacity 4 with 1000 vehicles on NYDataset, it obtains 8.89\% improvement over ZAC, which increases to 9.5\% for capacity 10. On Dataset1 for 1000 vehicles with capacity 4, ZACBenders obtains 4.2\% improvement over ZAC, which increases to 4.6\% for capacity 10. 
\item While both ZAC and TBF can compute a solution in less than 20 seconds, the time taken by ZAC is much less than TBF. The time taken by ZACBenders is much more than the myopic algorithms TBF and ZAC, but the service rate improvement compensates for the additional runtime. 

\squishend

We also experimented with all the vehicles having different maximum capacity. In this case, to compute the weight of edges in bipartite graph in ZACBenders, $\kappa$ is taken as average of all vehicle's maximum capacity. Figure \ref{fig:fignyservratevcapdiff} shows the results where vehicle capacities are generated by taking different distributions. We experimented with following three distributions:
\begin{enumerate}
\item Uniform\_4: The maximum capacity of each vehicle is sampled uniformly between 1 to 4.
\item Uniform\_10: The maximum capacity of each vehicle is sampled uniformly between 1 to 10.
\item 80\_20: 80\% of the vehicles have maximum capacity as 4 and 20\% of the vehicles have maximum capacity as 6. This is based on the observation that ridesharing companies like Uber, Lyft etc have majority of vehicles with maximum capacity 4 and some vehicles with maximum capacity 6.
\end{enumerate}
In this case also, we observe that ZACBenders obtains improvement over myopic approaches. For Uniform\_4, the improvement over ZAC is 7.26\%, for Uniform\_10 the improvement is 11.94\% and for 80\_20, the improvement obtained is 11.61\%. The two-stage stochastic approximation in this case works better than all vehicles having identical maximum capacity as when vehicles have identical maximum capacity, the $\kappa$ value used in second stage will be higher and it will allow more requests to group at future decision epoch causing the future value to be overestimated in some cases. 

 %(2) The runtime of both TBF and ZAC increases as the number of vehicles increase but as explained before, on NYDataset, the rate of increase in runtime of TBF is higher than the rate of increase in the runtime of ZAC. 

\begin{figure*}[htbp]
  \centering
%    \subfloat[]{\label{fig:figvarydeltaservny2}\includegraphics[width=0.16\textwidth,height=1.4in]{images/varydelta_2_servrate_ny.pdf}}
    \subfloat{\label{fig:figvarydeltaservny4}\includegraphics[width=0.25\textwidth,height=1.4in]{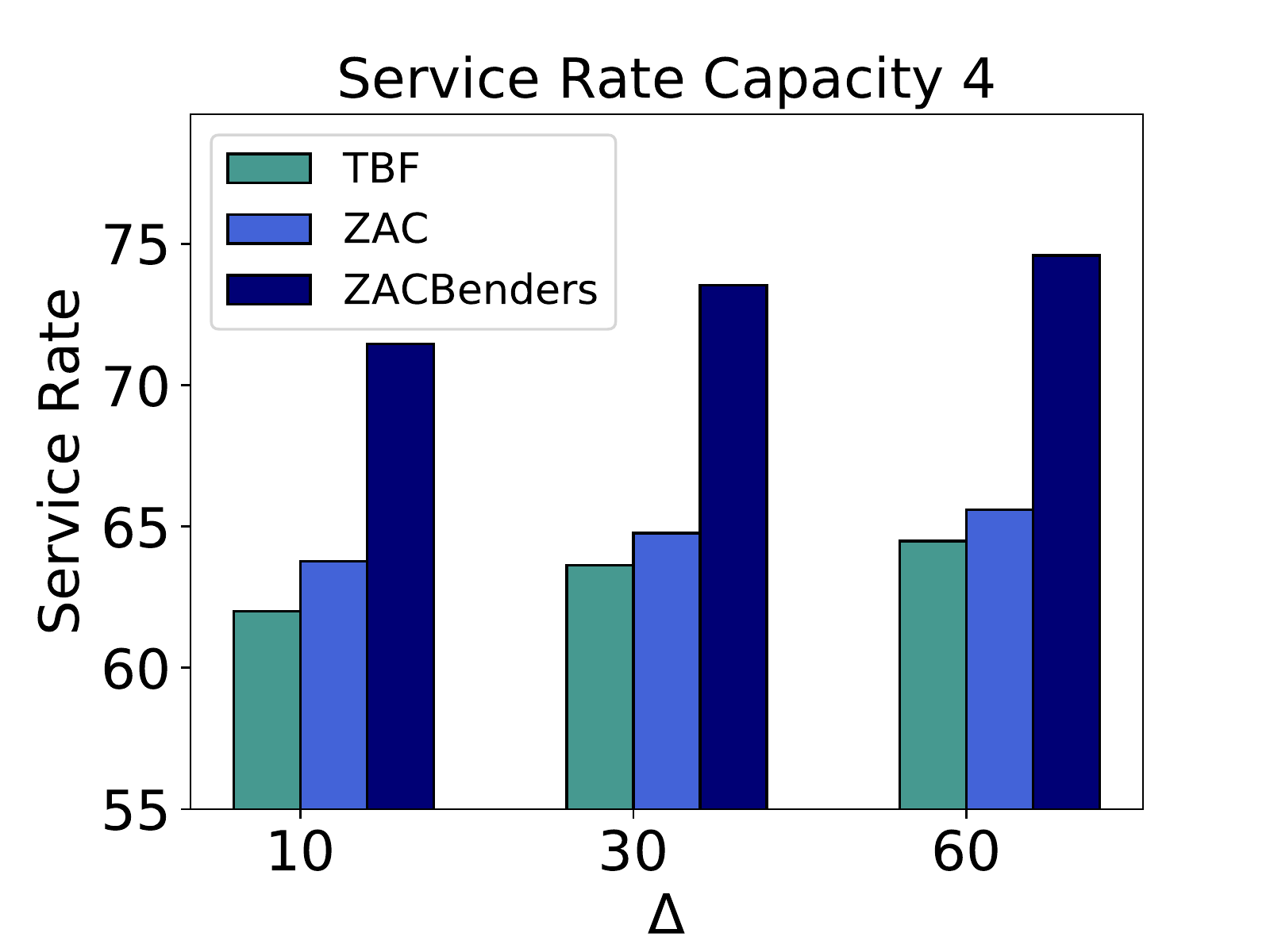}}
    \subfloat{\label{fig:figvarydeltaservny10}\includegraphics[width=0.25\textwidth,height=1.4in]{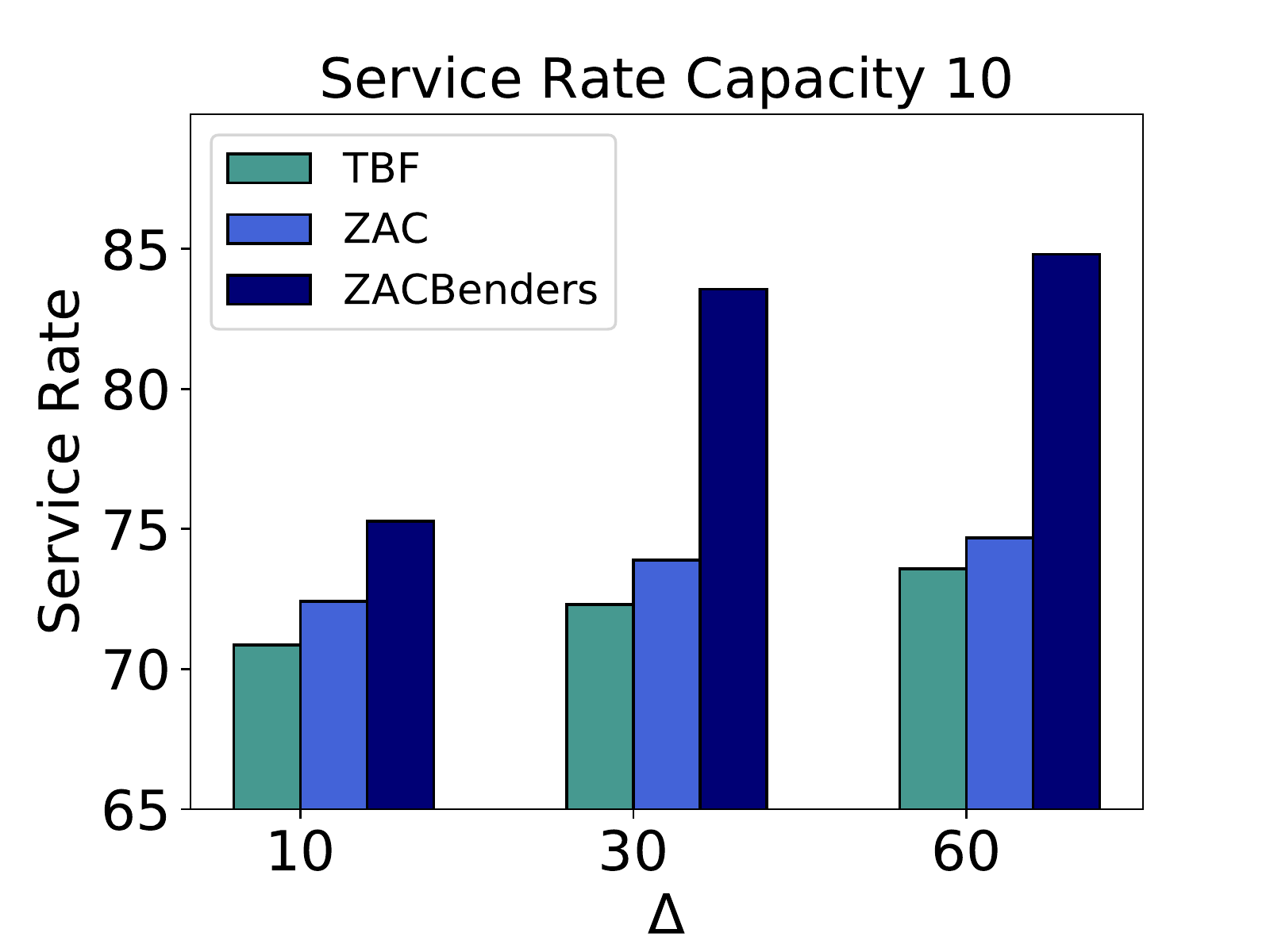}}
 %   \subfloat[]{\label{fig:figvarydeltartimeny2}\includegraphics[width=0.16\textwidth,height=1.4in]{images/varydelta_2_rtime_ny.pdf}}
    \subfloat{\label{fig:figvarydeltartimeny4}\includegraphics[width=0.25\textwidth,height=1.4in]{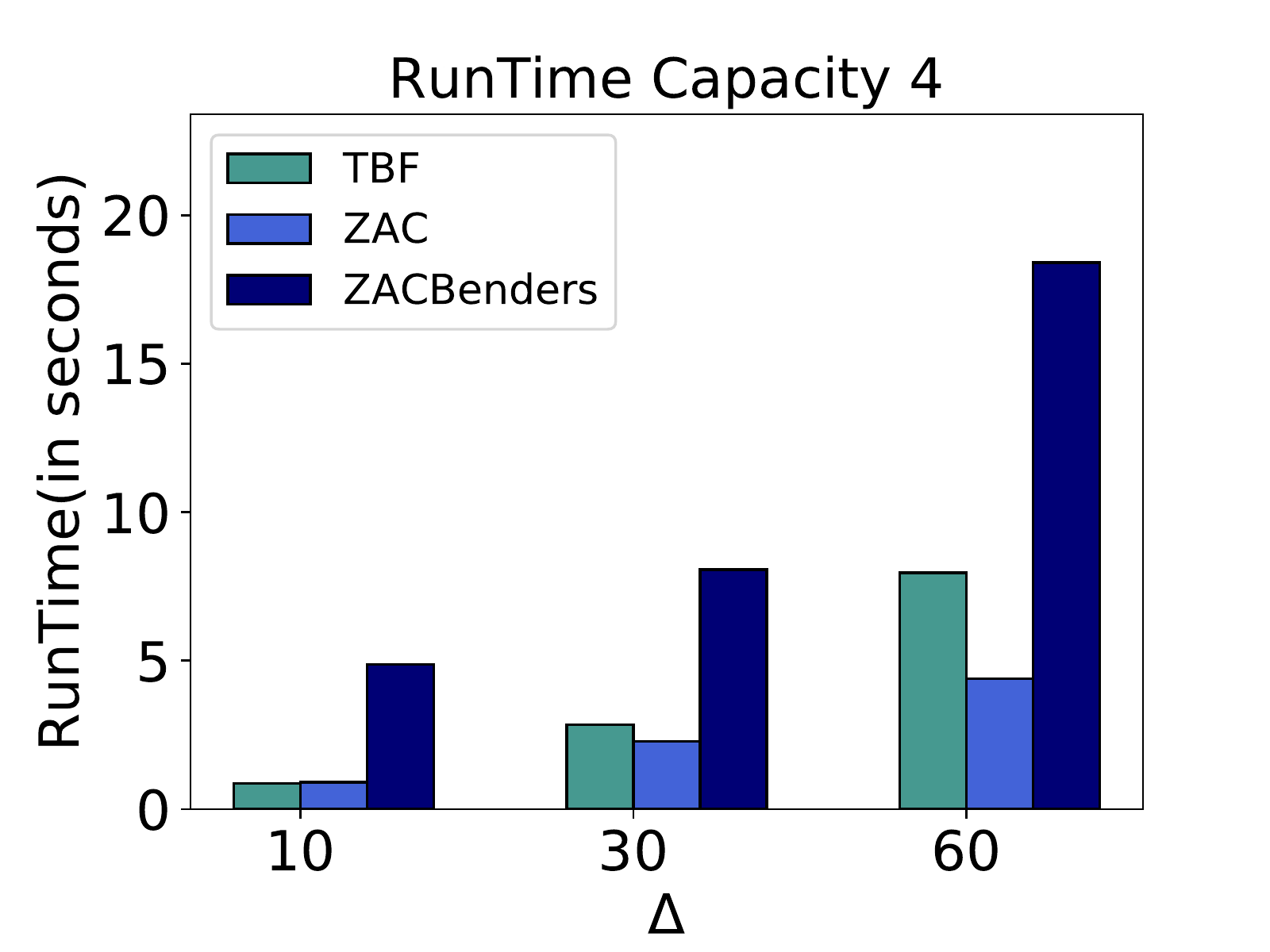}}
    \subfloat{\label{fig:figvarydeltartimeny10}\includegraphics[width=0.25\textwidth,height=1.4in]{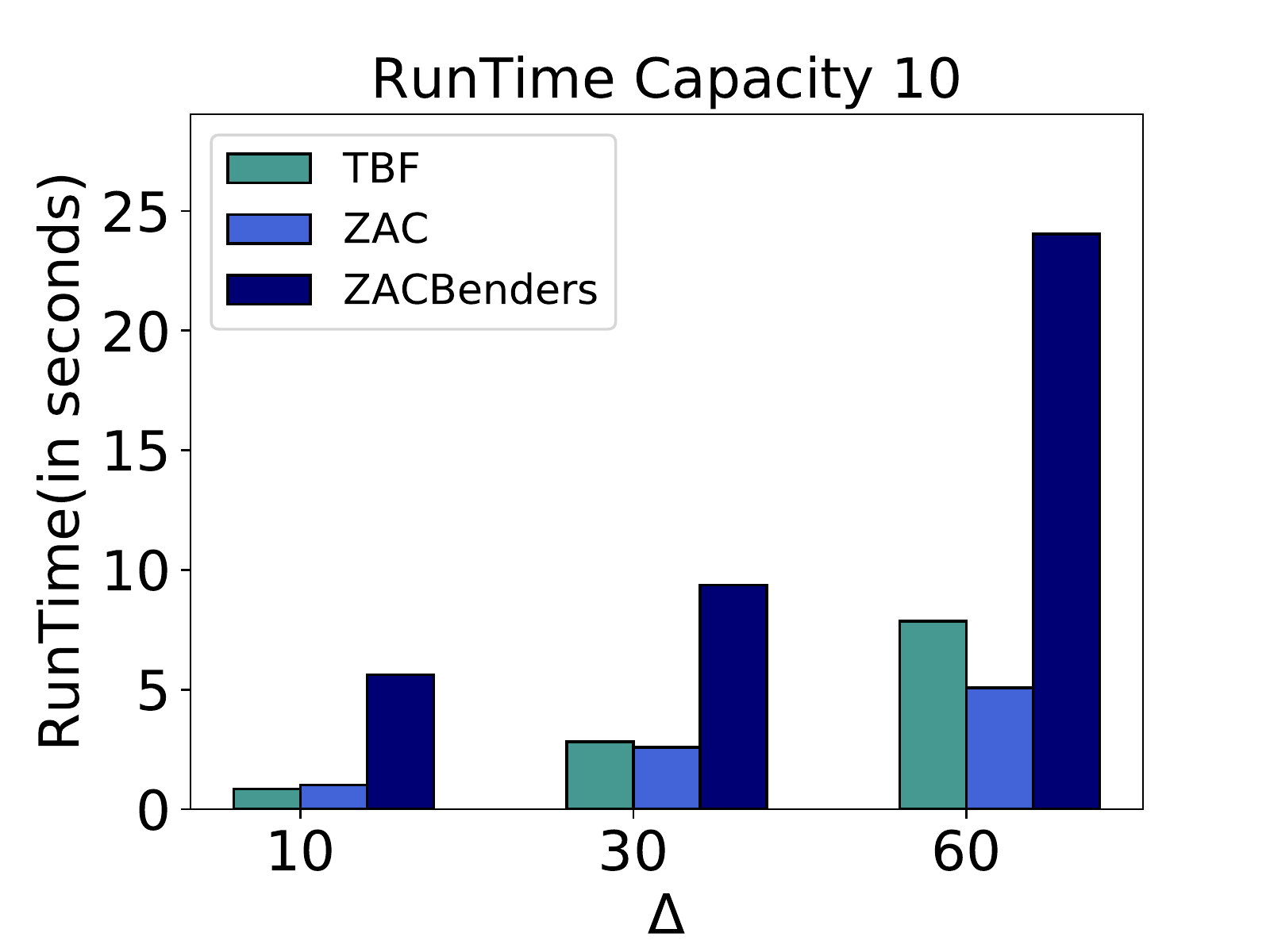}}
  \caption{Comparison of ZACBenders, ZAC and TBF for NYDataset for 1000 vehicles and varying values of $\Delta$, $\tau$ = 300, $\lambda$ = 600 seconds}
  \label{fig:figny_varydelta}
\end{figure*}

\begin{figure*}[htbp]
  \centering
%    \subfloat[]{\label{fig:figpeakservrate2ny}\includegraphics[width=0.16\textwidth,height=1.4in]{images/PeakNonPeak_2_servrate_5min_10min_ny.pdf}}
    \subfloat{\label{fig:figpeakservrate4ny}\includegraphics[width=0.25\textwidth,height=1.4in]{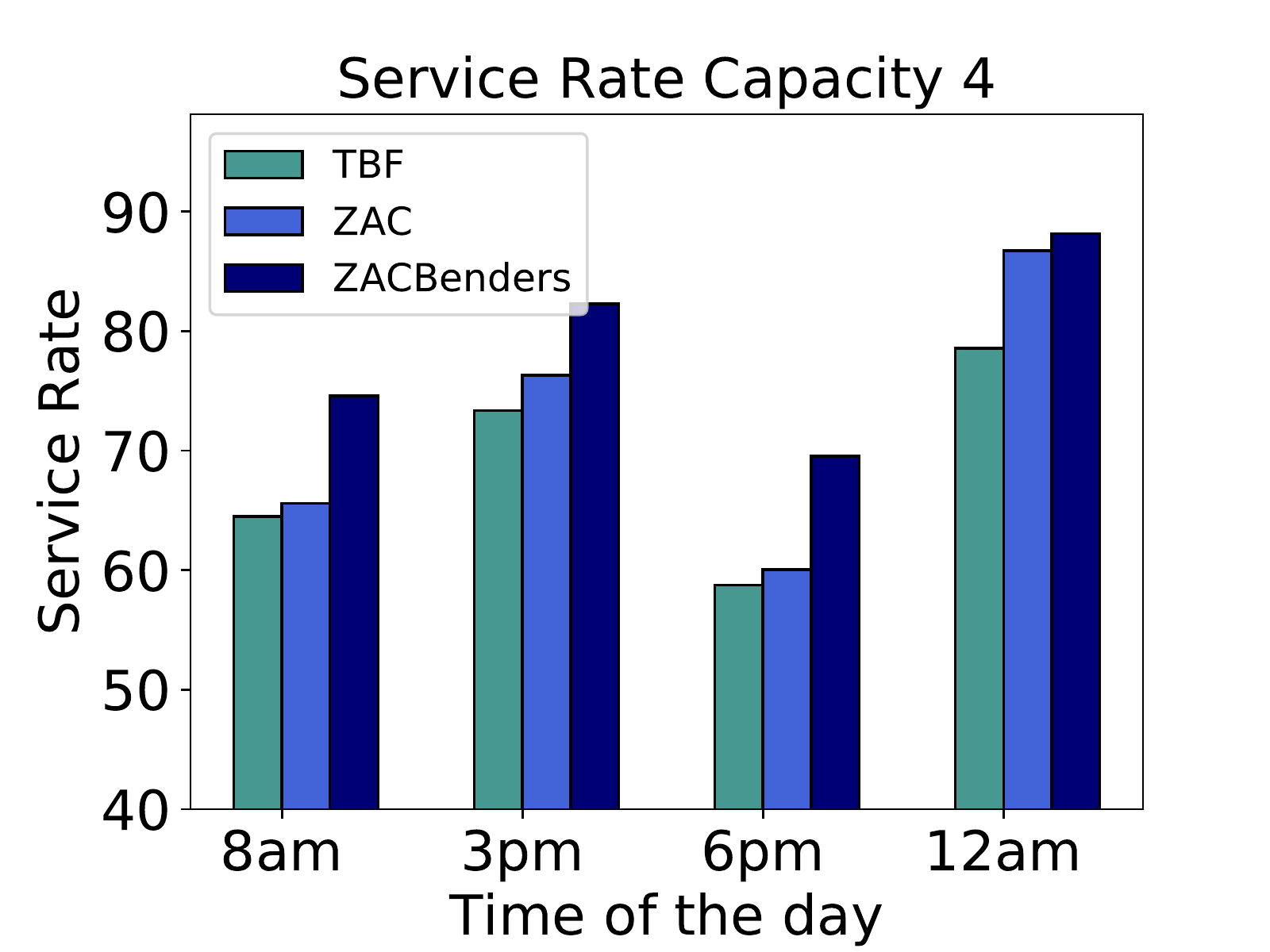}}
    \subfloat{\label{fig:figpeakservrate10ny}\includegraphics[width=0.25\textwidth,height=1.4in]{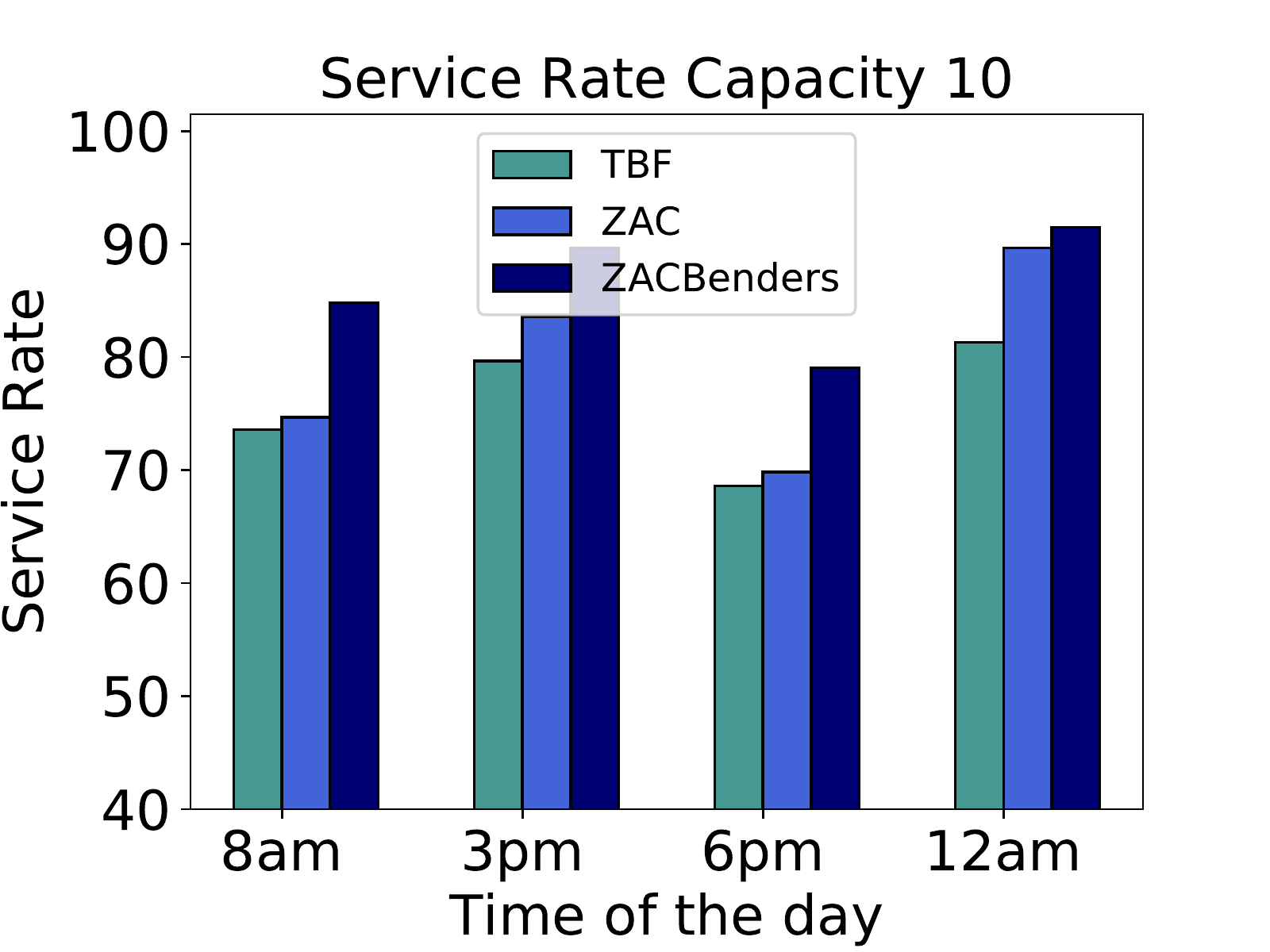}}
 %   \subfloat[]{\label{fig:figpeakrtime2ny}\includegraphics[width=0.16\textwidth,height=1.4in]{images/PeakNonPeak_2_rtime_5min_10min_ny.pdf}}
    \subfloat{\label{fig:figpeakrtime4ny}\includegraphics[width=0.25\textwidth,height=1.4in]{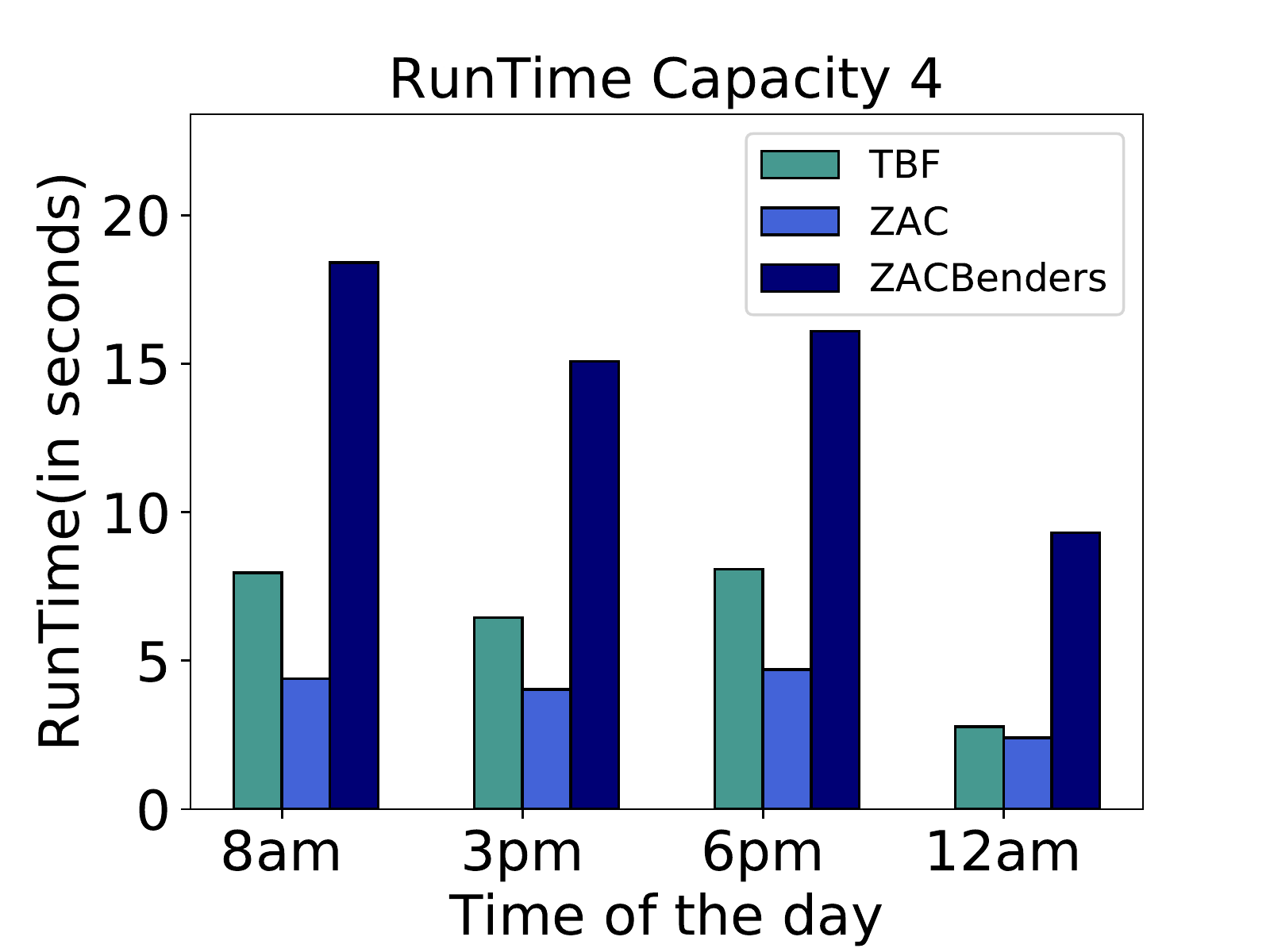}}
    \subfloat{\label{fig:figpeakrtime10ny}\includegraphics[width=0.25\textwidth,height=1.4in]{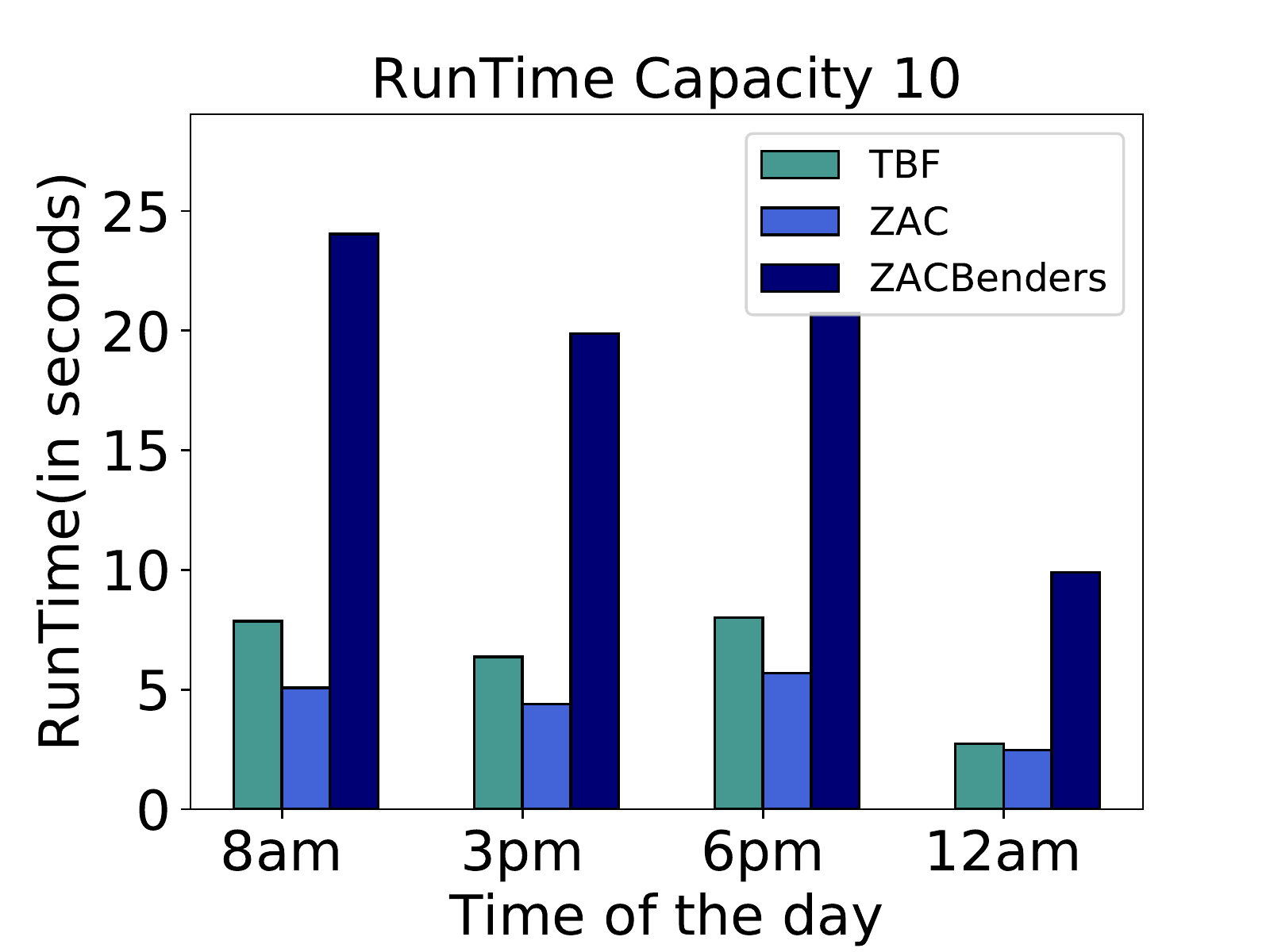}}
  \caption{Comparison of ZACBenders, ZAC and TBF for NYDataset for 1000 vehicles and different time of the day. $\tau$ = 300, $\lambda$ = 600 seconds}
  \label{fig:figny_varytime}
\end{figure*}

\noindent \textbf{Effect of change in value of $\Delta$:} We compare the service rate and runtime of algorithms for different values of $\Delta$ (Figure \ref{fig:figny_varydelta}). Here are the key observations: 
\squishlist
\item Service rate increases as the value of $\Delta$ increases. This is because more requests are available at each decision epoch which allows grouping more requests together.\item The difference between the service rate of ZACBenders and ZAC increases as the $\Delta$ increases. One of the reasons is that the value of $\Delta$ limits the time available for computation of assignments, when $\Delta$ value is low, less number of benders decomposition iterations can be executed within time limit, which affects the performance of ZACBenders. 
\item The time taken by TBF is much more than ZAC for larger $\Delta$ values due to the presence of more number of requests at each decision epoch.
\squishend

\noindent \textbf{Effect of time of the day:} We compare the effect of time of day on the performance of algorithms (Figure \ref{fig:figny_varytime}). Here are the key observations: 
\squishlist 
\item The service rate of ZAC is more than TBF in each time interval and ZACBenders further improves this service rate. 
\item The difference between service rate of ZAC and TBF is more during non-peak hours (3pm and 12am). This is likely as there are less requests available at each decision epoch, so as opposed to peak time where there is more possibility of grouping requests across decision epochs, at non-peak times it is advantageous to explore more combinations at a single decision epoch. The other reason is that ZAC is able to rebalance vehicles better by assigning them to zone paths.  
\squishend

\begin{figure}[htbp]
 \centering
%  \subfloat[]{\label{fig:figvarywtimeservny2}\includegraphics[width=0.16\textwidth,height=1.2in]{images/varyWtime_1000_2_servrate_ny.pdf}}
  \subfloat{\label{fig:figvarywtimeservny4}\includegraphics[width=0.33\textwidth,height=1.4in]{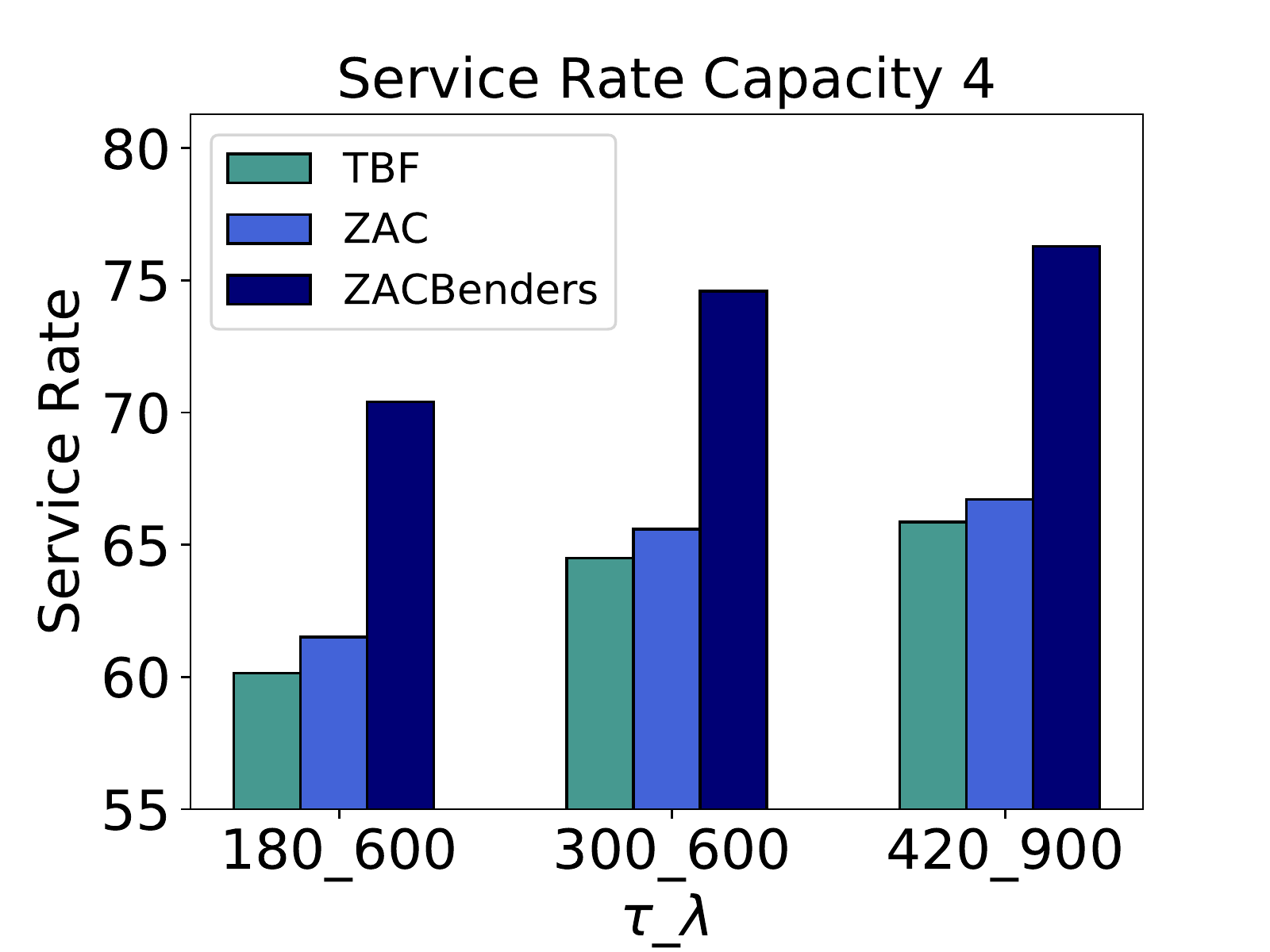}}
  \subfloat{\label{fig:figvarywtimeservny10}\includegraphics[width=0.33\textwidth,height=1.4in]{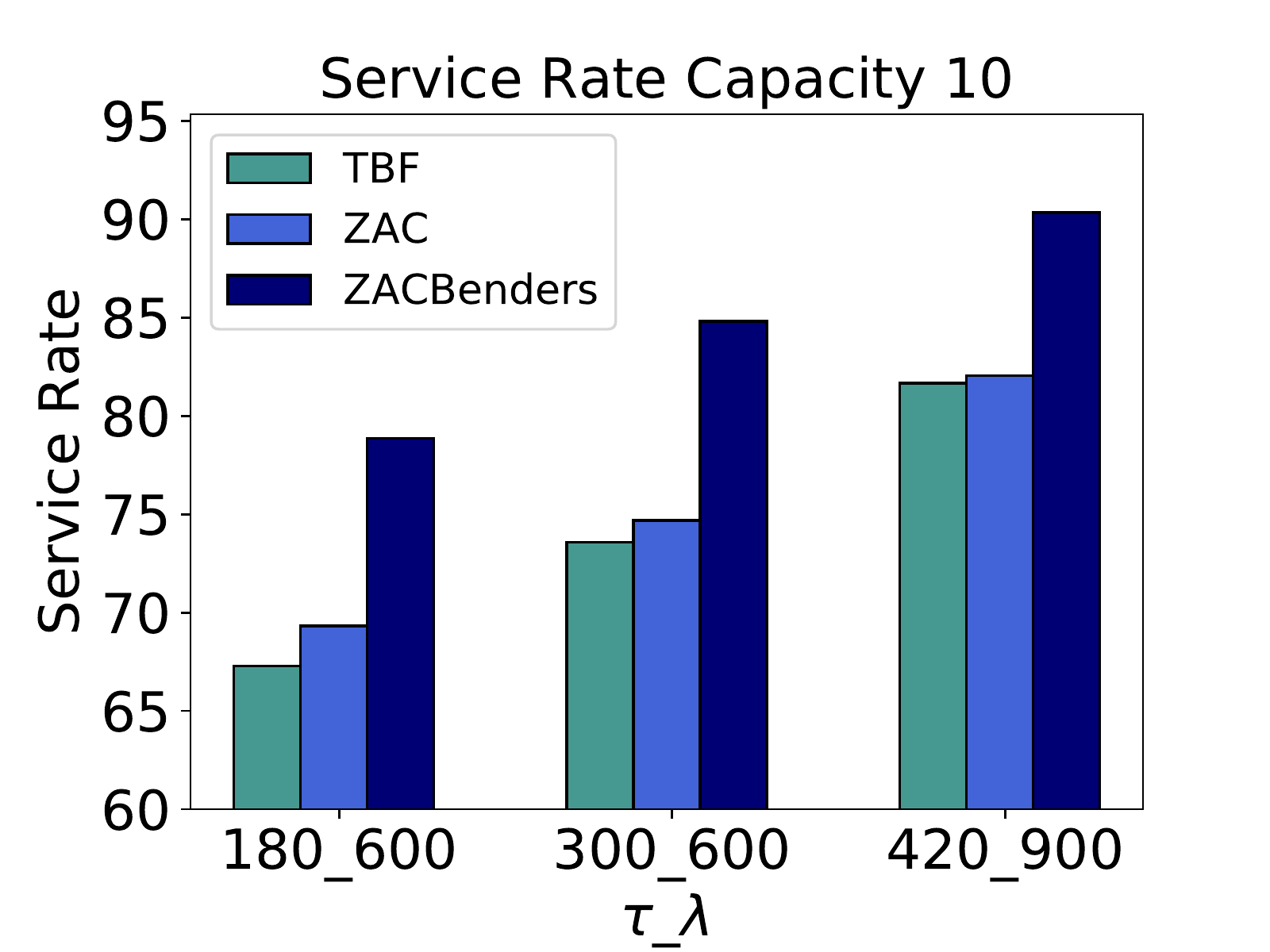}}
 % \subfloat[]{\label{fig:figvarywtimertimeny2}\includegraphics[width=0.16\textwidth,height=1.2in]{images/varyWtime_1000_2_rtime_ny.pdf}}
 % \subfloat{\label{fig:figvarywtimertimeny4}\includegraphics[width=0.16\textwidth,height=1.0in]{images/varyWtime_1000_4_rtime_ny.pdf}}
  \subfloat{\label{fig:figvarywtimertimeny10}\includegraphics[width=0.33\textwidth,height=1.4in]{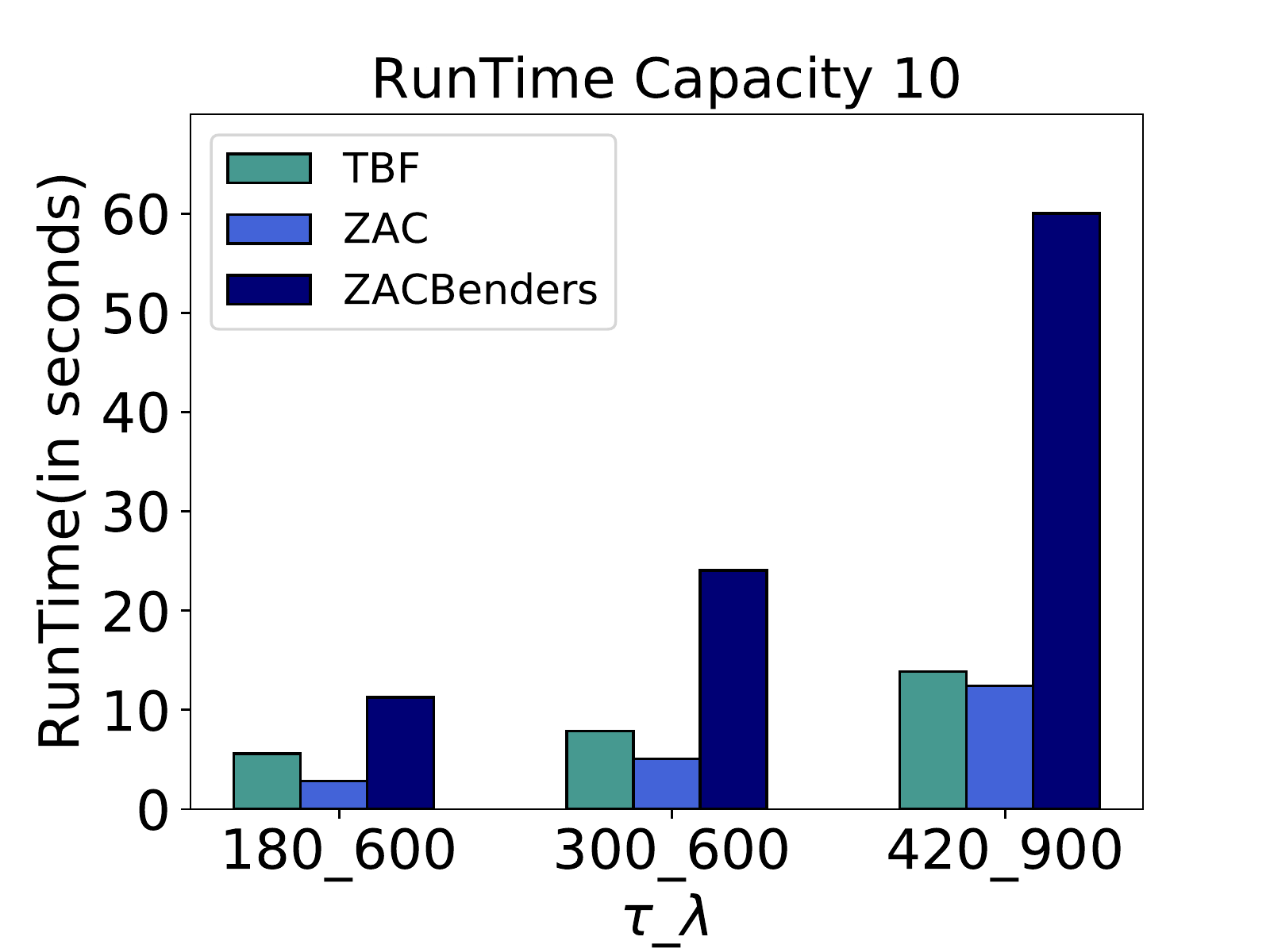}}
 \caption{{NYDataset - 1000 vehicles, $\Delta$=60 seconds.}}
 \label{fig:figny_varywtime}
\end{figure}

\noindent \textbf{Effect of change in values of $\tau$ and $\lambda$:} We show the service rate and runtime results for different values of $\tau$ and $\lambda$ in Figure \ref{fig:figny_varywtime}. Irrespective of the delay constraints, service rate obtained by ZAC is either more or same as TBF and the runtime of ZAC remains less than TBF in all cases. The improvement in the service rate obtained by ZACBenders over ZAC is also consistent across different values of $\tau$ and $\lambda$. The time taken by ZACBenders increases as the value of $\tau$ and $\lambda$ increases due to increase in the complexity of the optimization formulation. 

On real datasets, ZAC obtains up to 4\% gain in service rate over TBF across different parameter values. ZACBenders obtains nearly 10\% improvement in service rate over ZAC on NYDataset and 5\% improvement on Dataset1~\footnote{This gain further increases when evaluated over longer duration (24 hours) as opposed to 1 hour as seen in the results in the Section \ref{sect:compareneuradp}.}. Typically, even a 0.5\% gain is considered significant on real taxi datasets (as shown by a real car aggregation company~\cite{xu2018large}), so the gain obtained by our algorithms is a significant gain. 

\subsubsection{Comparison with NeurADP}
\label{sect:compareneuradp}

In this section, we compare TBF, ZAC, ZACBenders and NeurADP~\cite{neuradp} on NYDataset. Both ZACBenders and NeurADP can compute an assignment within maximum $\Delta$ seconds. We compare the performance of all algorithms over 24 hours as both ZACBenders and NeurADP consider future information and potentially ignore requests at initial decision epochs to serve more requests in the future and, as a result, achieve higher service rates when evaluated over longer durations. 
%and on academic computers due to limited resources, it is not possible to get the models trained for all the datasets and parameter settings. Therefore, we perform a comparison with NeurADP on the limited set of parameters for which we have a trained model available. Later, we show that the models do not work well if parameter values are changed emphasizing the need of training a new model for every change.  
As indicated earlier, NeurADP was specifically trained on these parameter setting for NYDataset. Following are the key observations:
\begin{figure*}[h]
  \centering
    \subfloat[]{\label{fig:figwchange}\includegraphics[width=0.33\textwidth,height=1.4in]{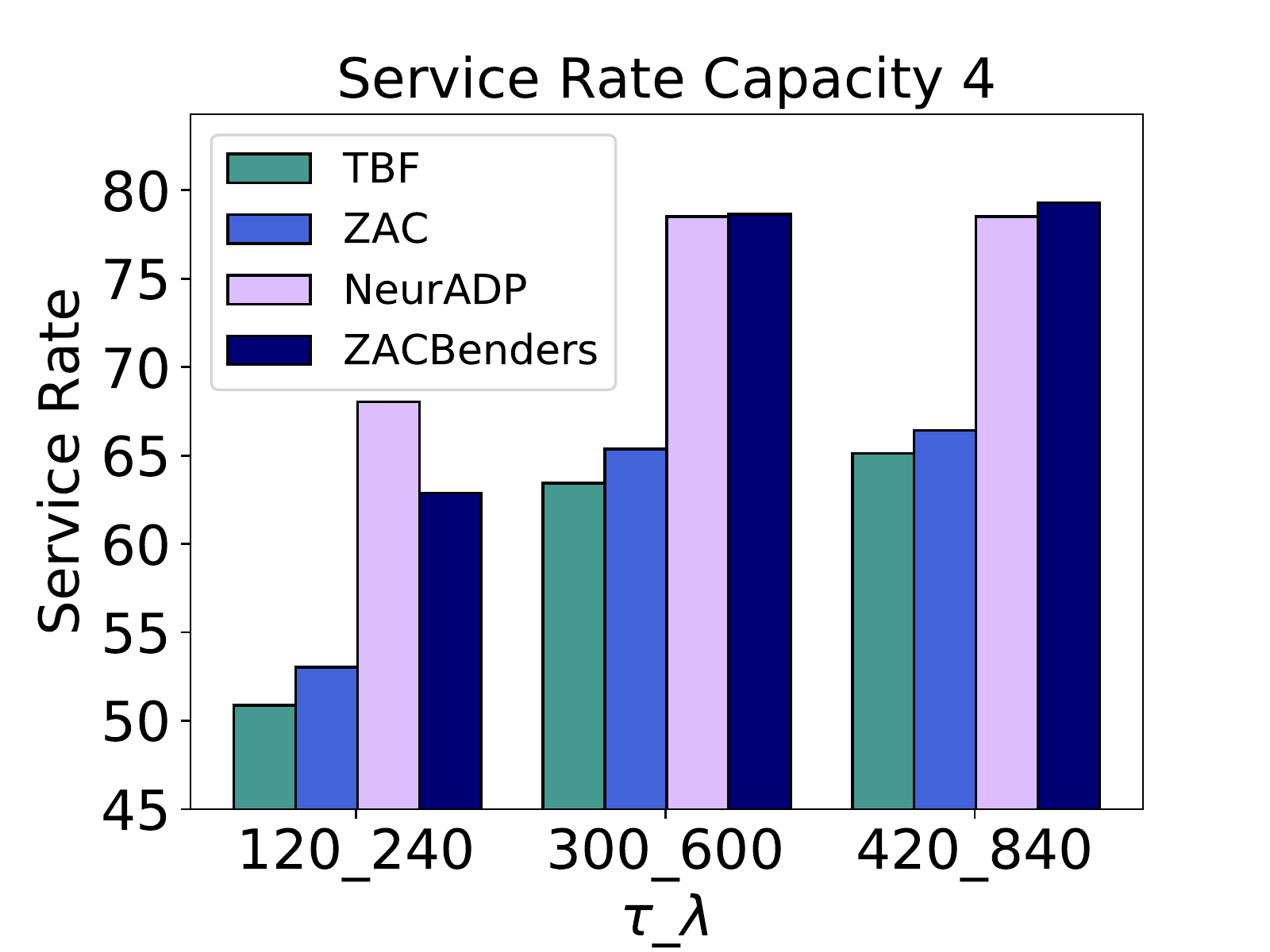}}
    \subfloat[]{\label{fig:figcapchange}\includegraphics[width=0.33\textwidth,height=1.4in]{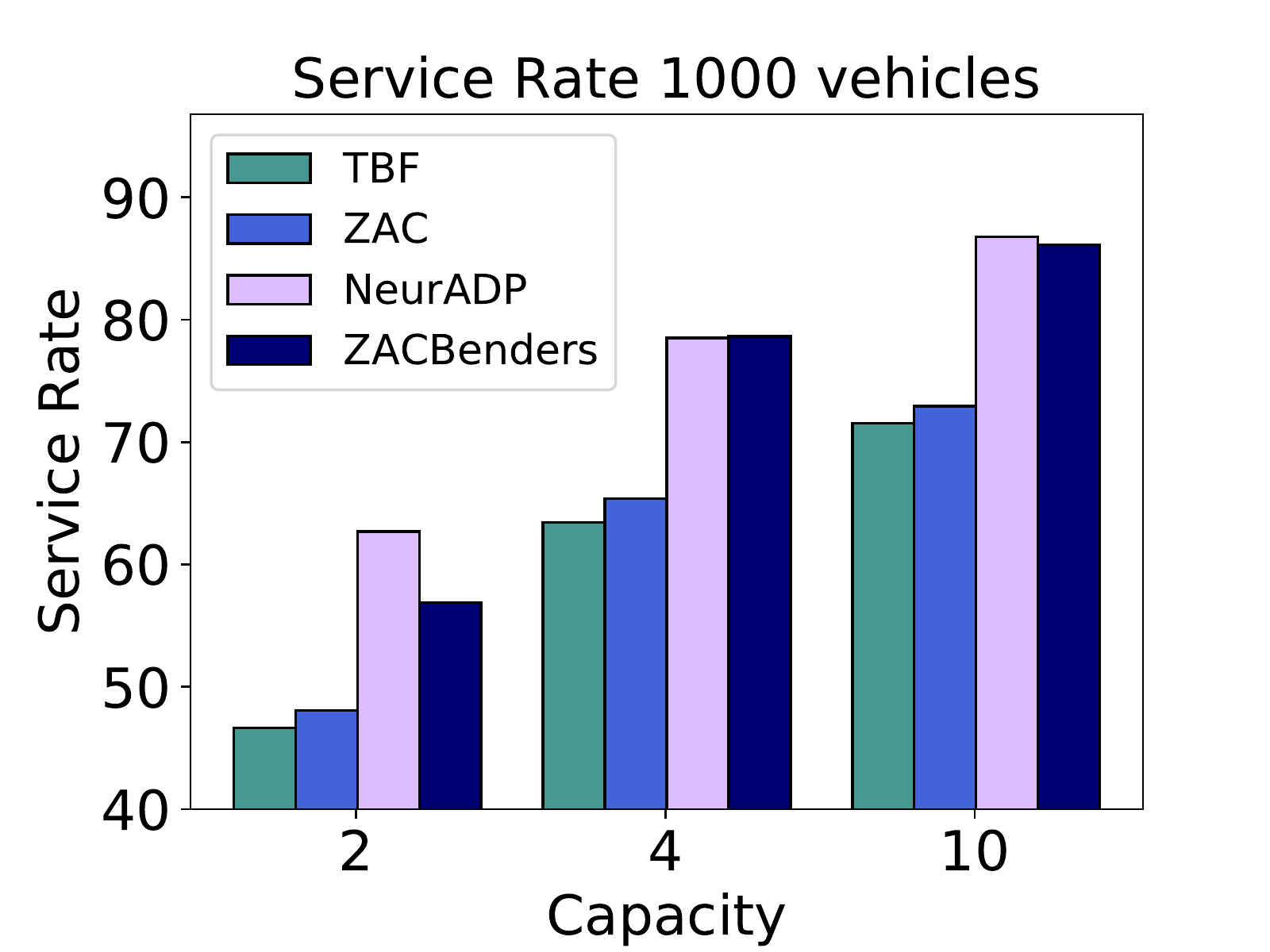}}
        \subfloat[]{\label{fig:figtaxichange}
    \includegraphics[width=0.33\textwidth,height=1.4in]{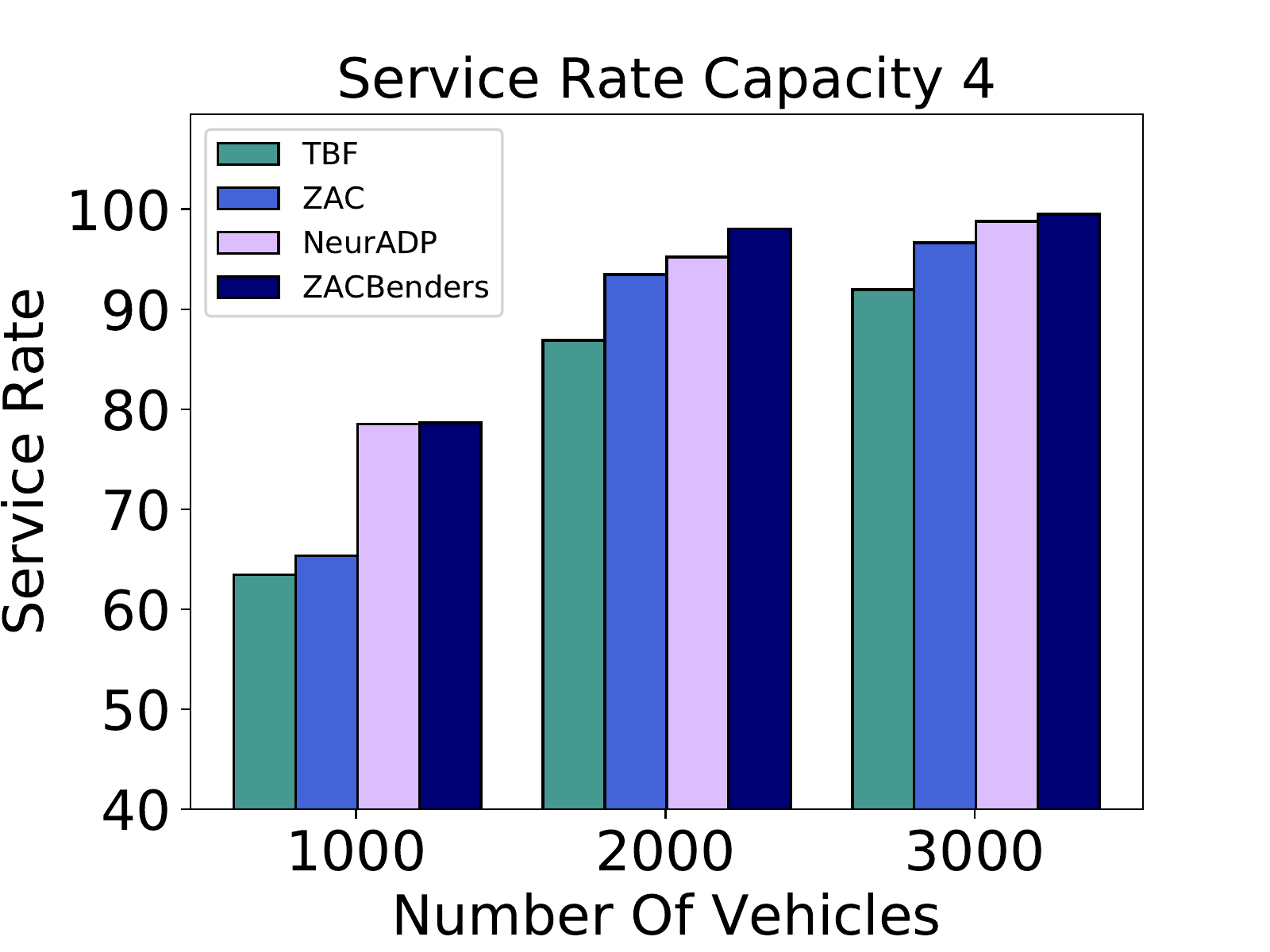}}
      \caption{{Comparison of service rate on NYDataset. (a) (b) Number of vehicles =1000 (b)(c) $\tau =300$,$\lambda=600$}}
  \label{fig:fignadpcompare}
\end{figure*}
\squishlist
\item For 1000 vehicles of capacity 4 with $\tau=300$ and $\lambda=600$ seconds, ZACBenders obtains 14.7\% improvement over TBF over 24 hours. 
\item ZACBenders and NeurADP outperform myopic approaches TBF and ZAC as shown in Figure \ref{fig:fignadpcompare}. ZACBenders and NeurADP have comparable performances, except in a couple of cases (with a maximum improvement of 6\% for NeurADP).
\squishend

\begin{figure*}[h]
  \centering
    \subfloat[]{\label{fig:figdeltachange1}\includegraphics[width=0.33\textwidth,height=1.4in]{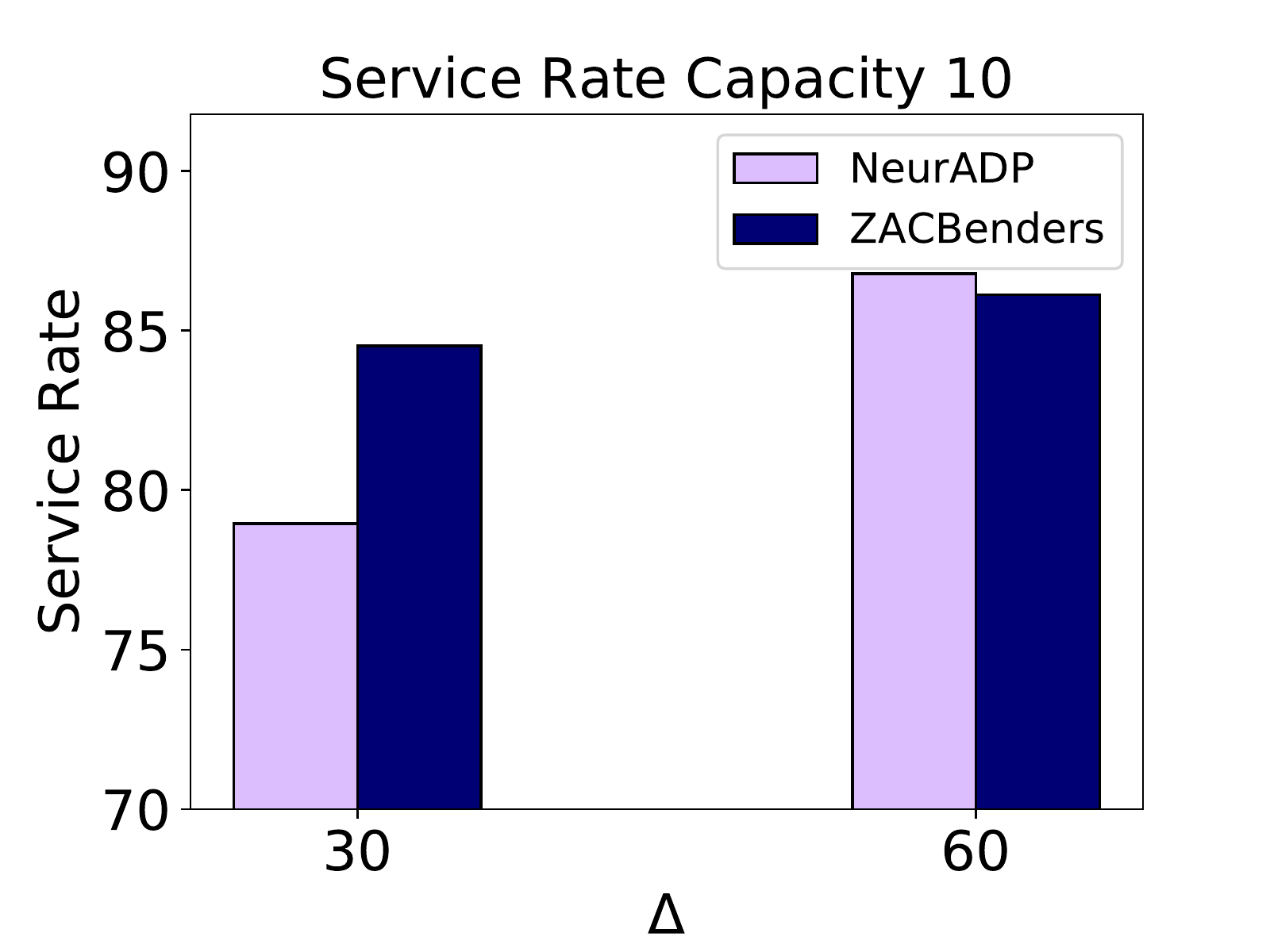}}
        \subfloat[]{\label{fig:figtaxichange1}
    \includegraphics[width=0.33\textwidth,height=1.4in]{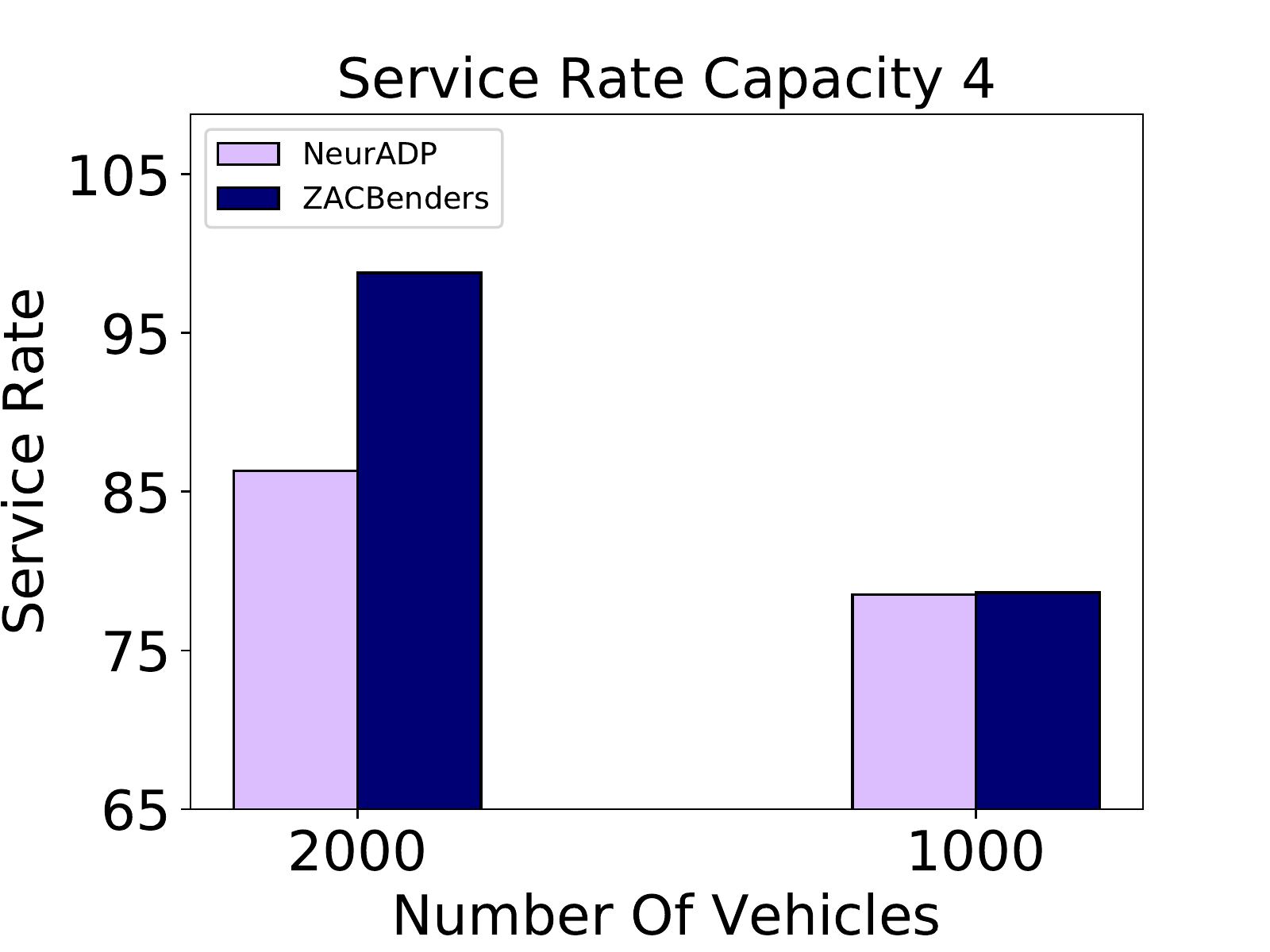}}
        \subfloat[]{\label{fig:figdemandichange1}
    \includegraphics[width=0.33\textwidth,height=1.4in]{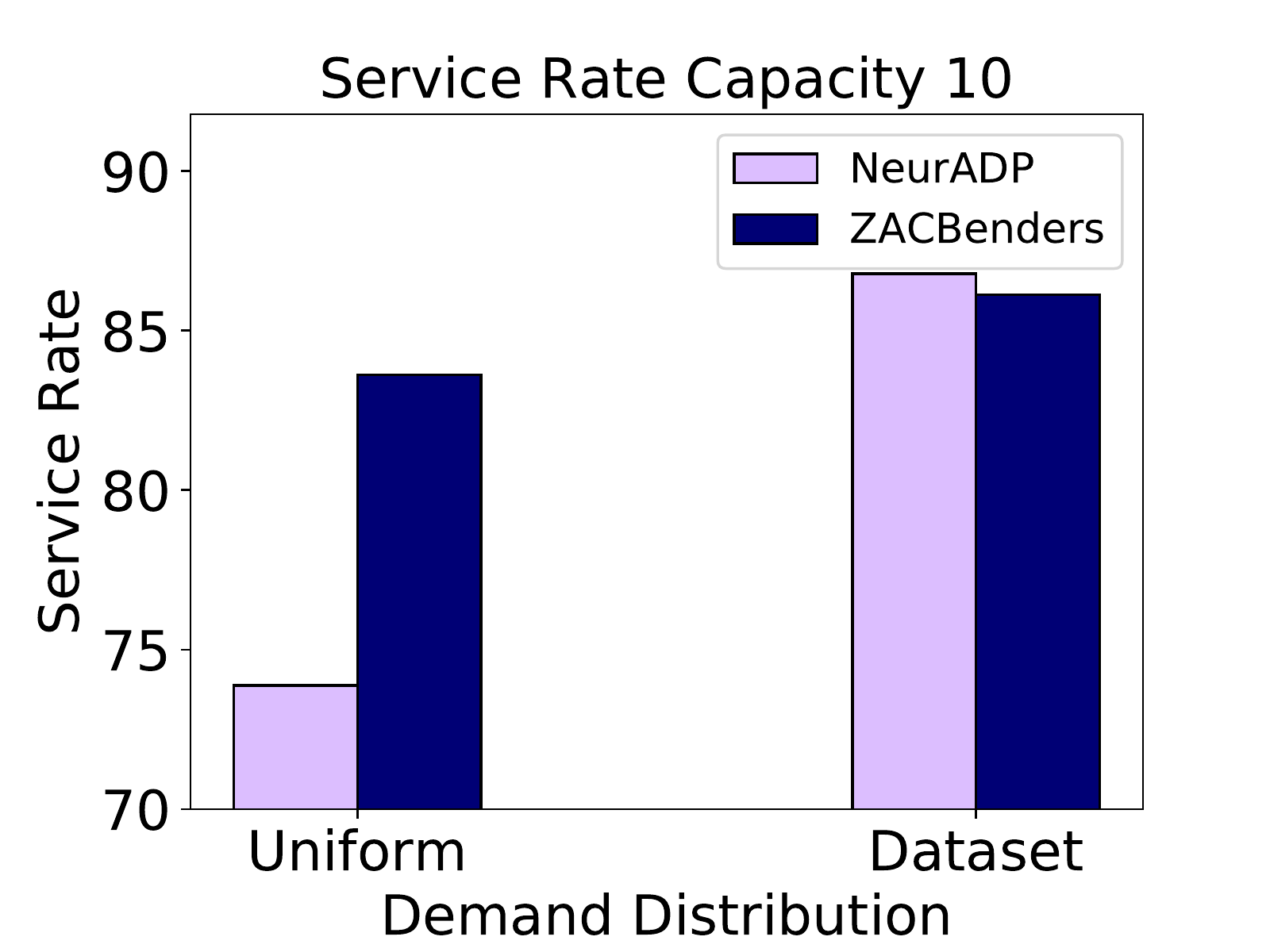}}
      \caption{{Comparison of service rate on NYDataset. For these results NeurADP uses the model trained for 1000 vehicles $\Delta =60$ seconds, $\tau=300$ seconds, $\lambda=600$ seconds and the appropriate capacity (4 or 10)  (a)(c) Number of vehicles =1000, (a)(b)(c) $\tau =300$,$\lambda=600$, (a) (c) Capacity = 10, (b) Capacity = 4}}
  \label{fig:fignadpcompare1}
\end{figure*}

\noindent \textbf{ Generalizability :} Learning a new model for every possible parameter configuration and every dataset is not scalable nor sustainable for real datasets. For instance, the model training is a time-consuming process and it takes around one week to learn for NeurADP. Unlike NeurADP, ZACBenders does not require training for every parameter setting. Therefore, we now evaluate generalizability of NeurADP in comparison to ZACBenders on test settings across three dimensions:
\begin{enumerate}
\item \textbf{Change in Decision Epoch Duration ($\Delta$):} If there is a requirement to reduce the decision epoch duration to enhance the user experience, it will not be possible to adopt this change using NeurADP without training a new model and optimizing the hyper parameters efficiently for this case. On the other hand, ZACBenders can be used readily for different values of $\Delta$. As shown in Figure \ref{fig:figdeltachange1}, if we take the model trained for $\Delta=60$ seconds and use it for $\Delta=30$ seconds, the performance of NeurADP is significantly affected. In this case, ZACBenders obtains 5.5\% improvement in service rate over NeurADP. 
\item \textbf{Change in Number of Vehicles:} If the ridesharing company decides to increase the fleet size (number of vehicles) and uses the NeurADP model trained for smaller fleet size, NeurADP is significantly outperformed by ZACBenders. When we take the model trained for 1000 vehicles and use it in a scenario with 2000 vehicles, ZACBenders obtains 12.48\% improvement in service rate over NeurADP. 
\item \textbf{Change in Demand Distribution:} There is going to be an event organized in the city which will cause demand patterns to deviate from the historical patterns. The change in the demand distribution can be predicted and we can get different samples from this predicted demand distribution. To use NeurADP approach in this scenario, we need to use these samples in NeurADP simulator and train the neural network model using that. So, it will not be possible to use NeurADP immediately. On the other hand, we can provide these samples to ZACBenders and start executing it. In Figure \ref{fig:figdemandichange1}, we show that instead of taking customer requests from the dataset, we evaluate the approaches by taking customer requests from a uniform distribution~\footnote{It can be shown using any distribution, we took uniform distribution as an example.}, ZACBenders can obtain 9.73\% improvement over NeurADP.
\end{enumerate}

\subsection{Results on Synthetic Dataset}
\label{sect:synthetic}
The real-world taxi datasets can not capture the scenarios for on demand shuttle services~\cite{shotl,beeline,grabshuttle} having a small set of pick-up/drop-off points in a city. These involve scenarios where many requests can be combined at each decision epoch. We represent these scenarios by simulating the case of first and last mile transportation in the synthetic network (details provided in experimental setup), where there are multiple requests at each decision epoch with either identical pick-up location and nearby drop-off locations or identical drop-off locations and nearby pick-up locations resulting in higher possibility of having large number of request combinations at a decision epoch.  

The gain obtained by ZAC over TBF is even more significant in these scenarios as TBF will not be exploring all relevant combinations while ZAC can explore more combinations by using zone paths. ZACBenders provide a slight improvement over ZAC by using future information but in these scenarios, as the travel times are small and the pick-up and drop-off locations of requests are near each other, the major improvement is obtained by exploring more combinations at a single decision epoch.

We compare the service rate obtained by TBF, ZAC and ZACBenders with different number of vehicles and different capacities and make following observations:\\
\squishlist
\item We observe that with 500 vehicles and capacity 10, ZAC can obtain 20.8\% improvement in service rate over TBF. The gain reduces to 16\% on increasing vehicles to 1000 as when more vehicles are available, it reduces the need of generating all combinations.
\item The service rate obtained by ZACBenders and ZAC is almost the same on this dataset as it is more important to explore more combinations in these scenarios. For 100 vehicles with capacity 10, ZACBenders obtains 2.2\% improvement over ZAC and for 500 vehicles with capacity 4 ZACBenders obtains 2\% improvement over ZAC. 
\squishend

These results demonstrate that ZAC is able to consider significantly more trips than TBF. 

\begin{figure}[htbp]
 \centering
  \subfloat[]{\label{fig:fig100synserv}\includegraphics[width=0.33\textwidth,height=1.4in]{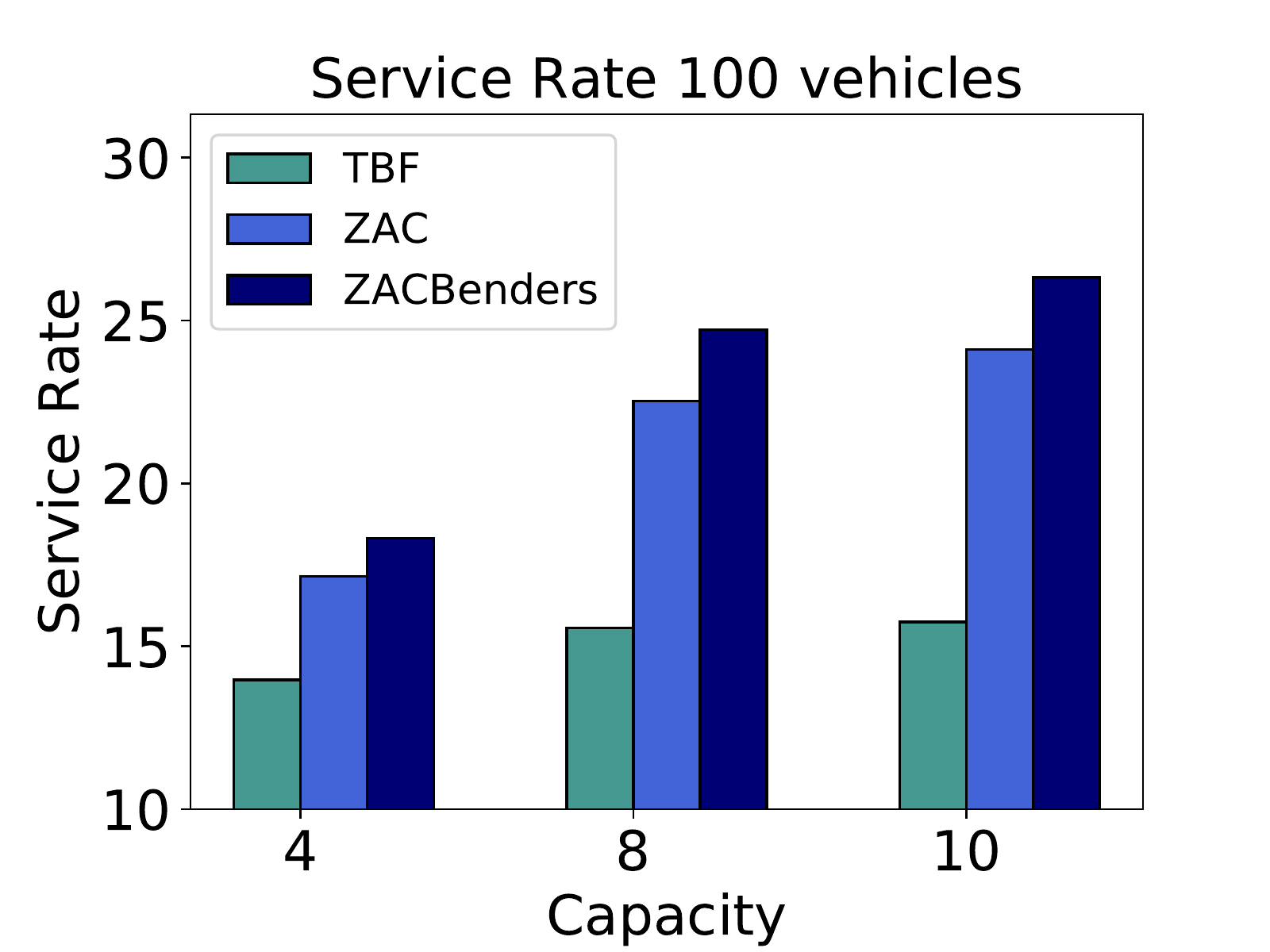}}
  \subfloat{\label{fig:fig500synserv}\includegraphics[width=0.33\textwidth,height=1.4in]{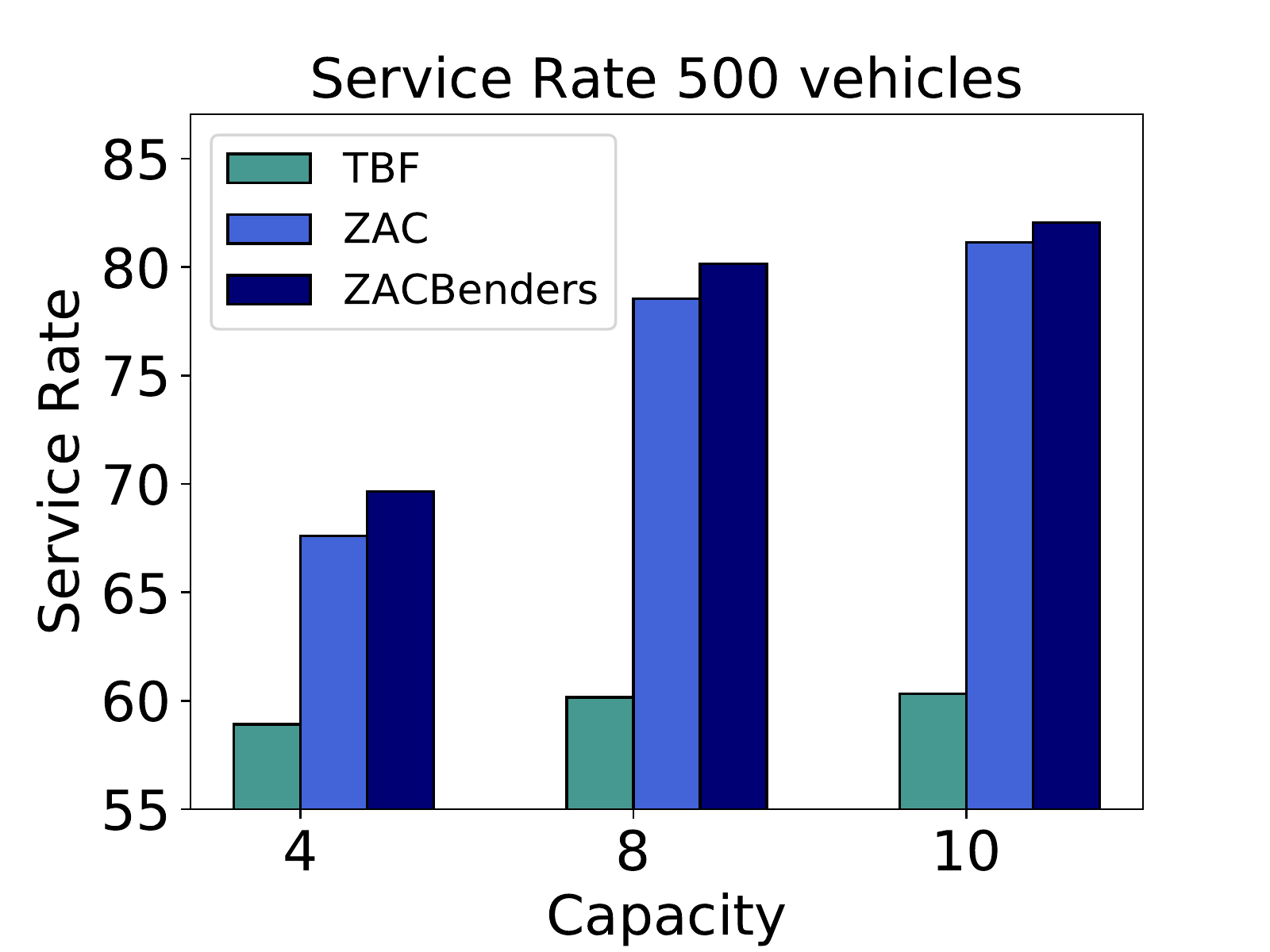}}
  \subfloat{\label{fig:fig1000synserv}\includegraphics[width=0.33\textwidth,height=1.4in]{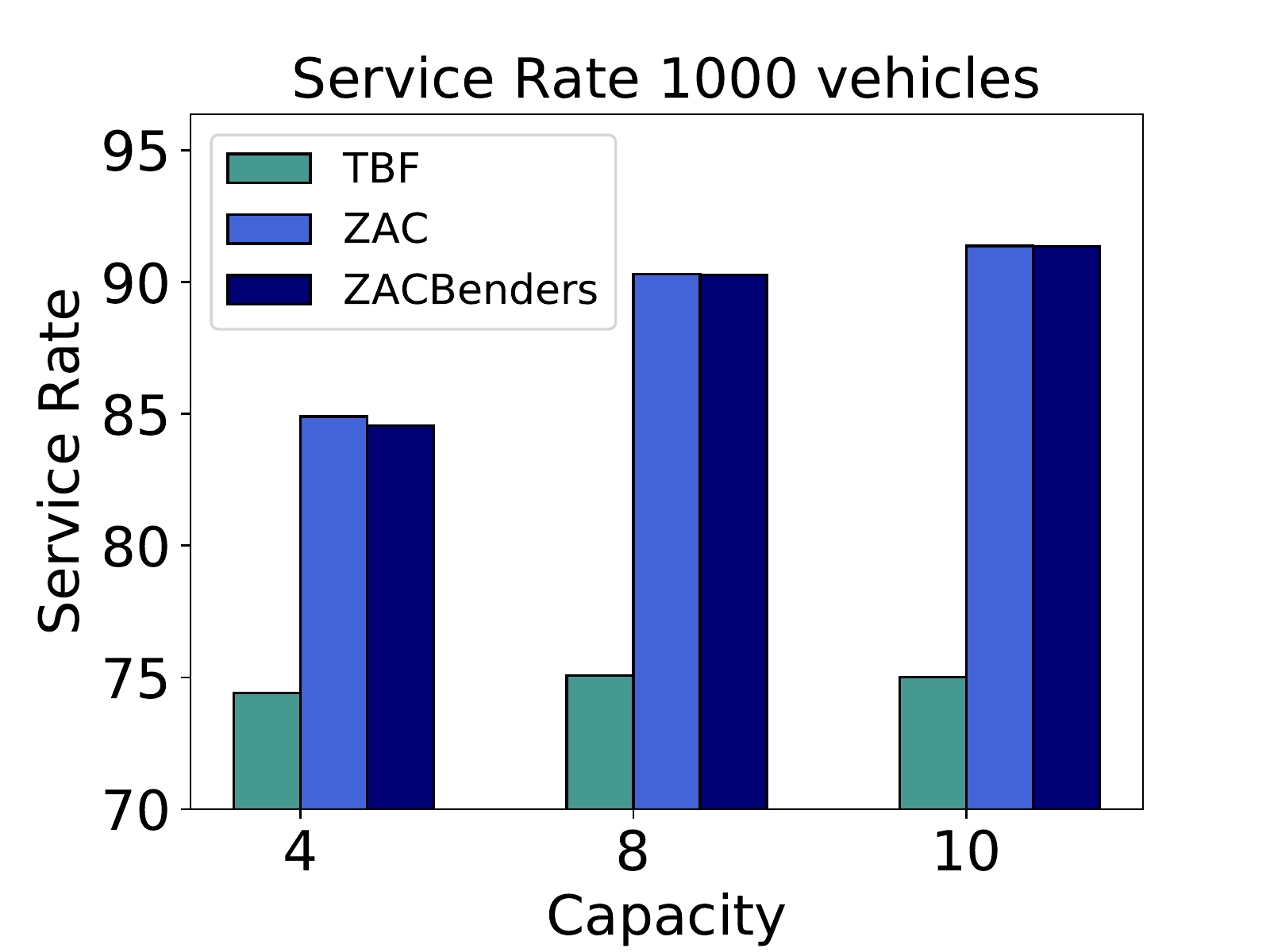}}
 \caption{{Synthetic Dataset with $\tau=120$,$\lambda=240$ and $\Delta=60$ seconds}}
 \label{fig:figsyn}
\end{figure}

\subsection{Justification for values of algorithmic parameter settings}
\label{sect:justifyparam}
In this section, we show the reason for using the fixed algorithmic parameter values (used in previous sections) for ZAC and ZACBenders. 

\begin{figure*}[h]
  \centering
    \subfloat[]{\label{fig:figclustservrate}\includegraphics[width=0.25\textwidth,height=1.4in]{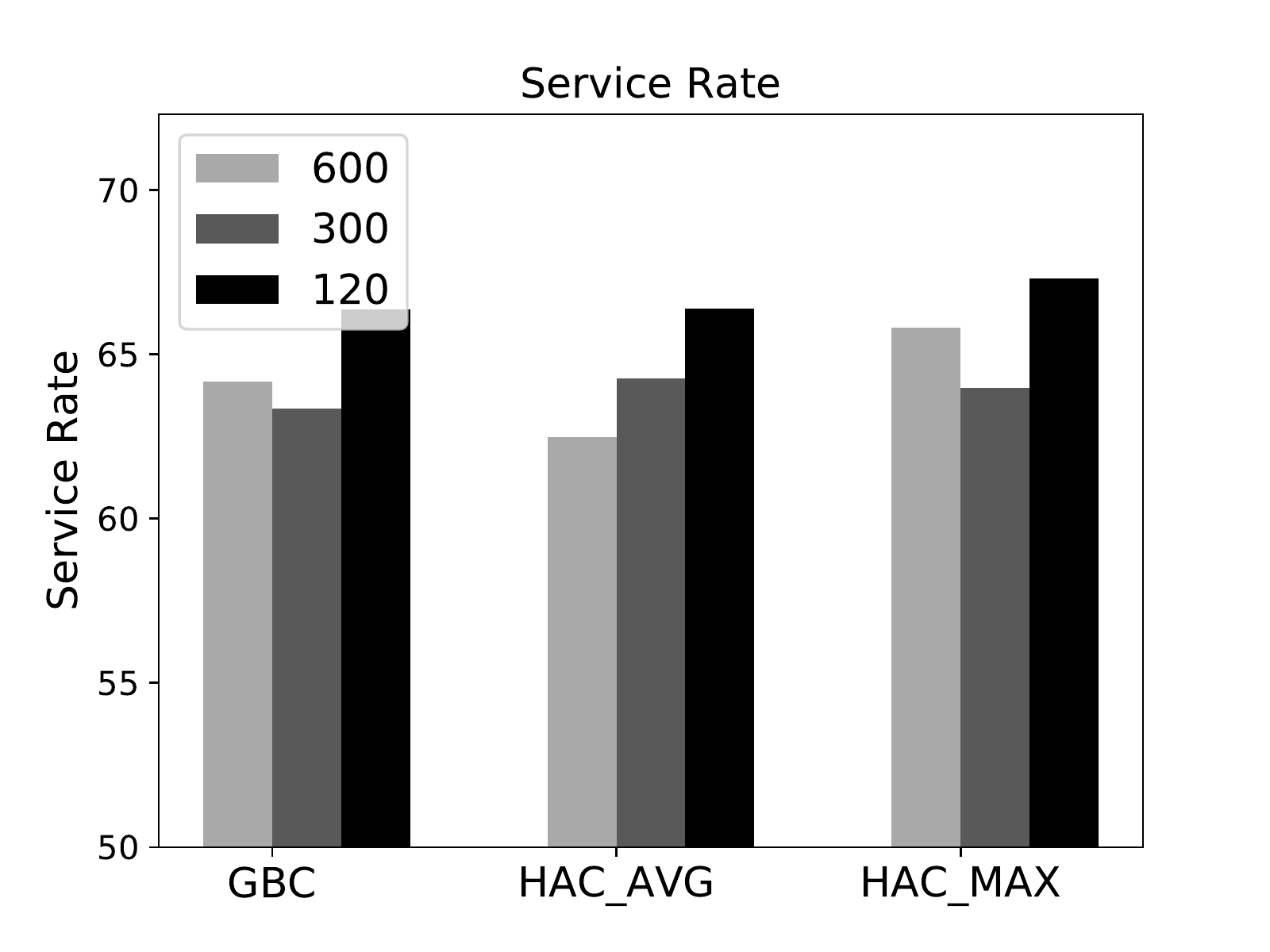}}
    \subfloat[]{\label{fig:figclustrtime}\includegraphics[width=0.25\textwidth,height=1.4in]{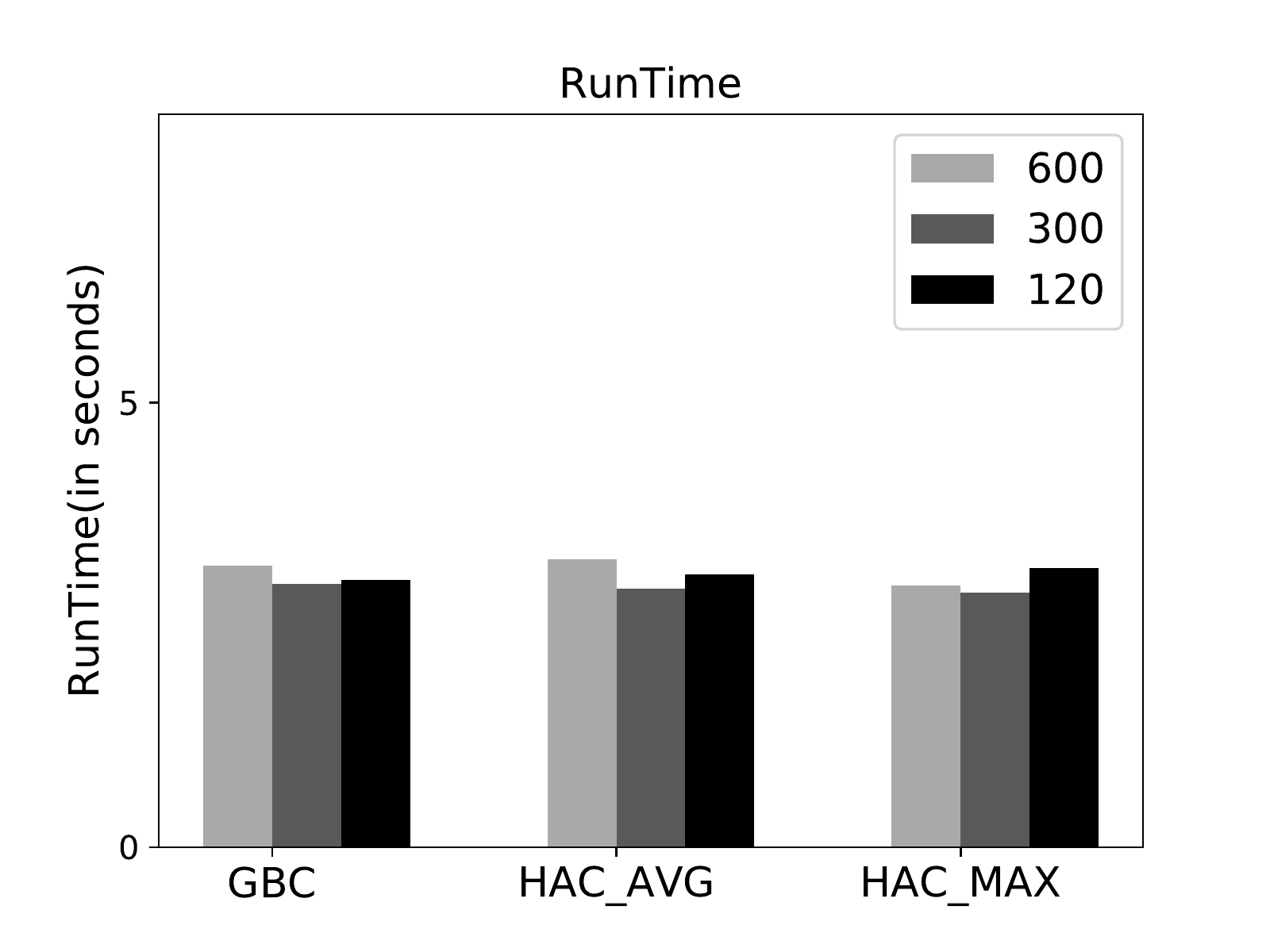}}
    \subfloat[]{\label{fig:figclustpercentreq}\includegraphics[width=0.25\textwidth,height=1.4in]{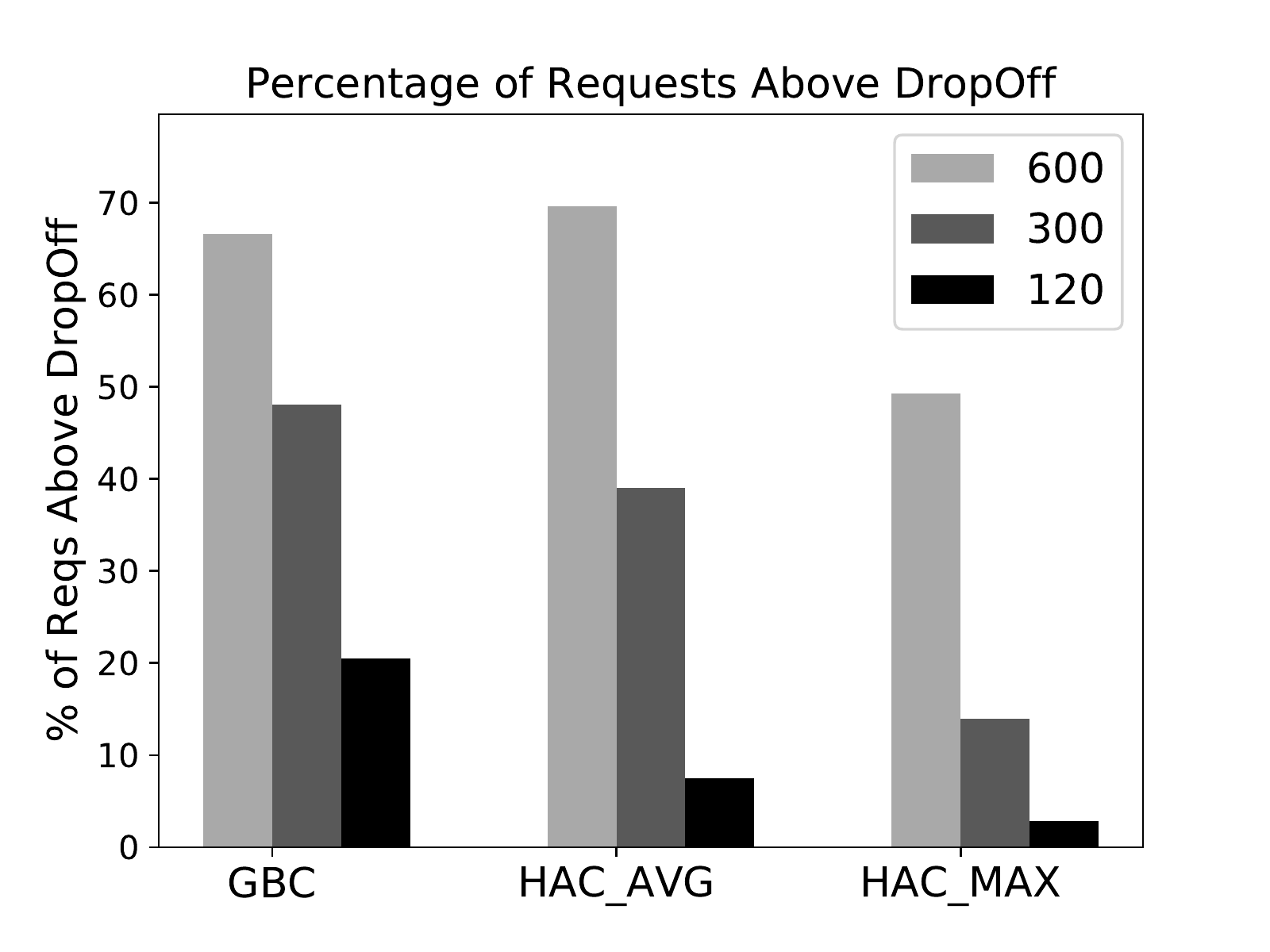}}
    \subfloat[]{\label{fig:figclustmaxdropoff}\includegraphics[width=0.25\textwidth,height=1.4in]{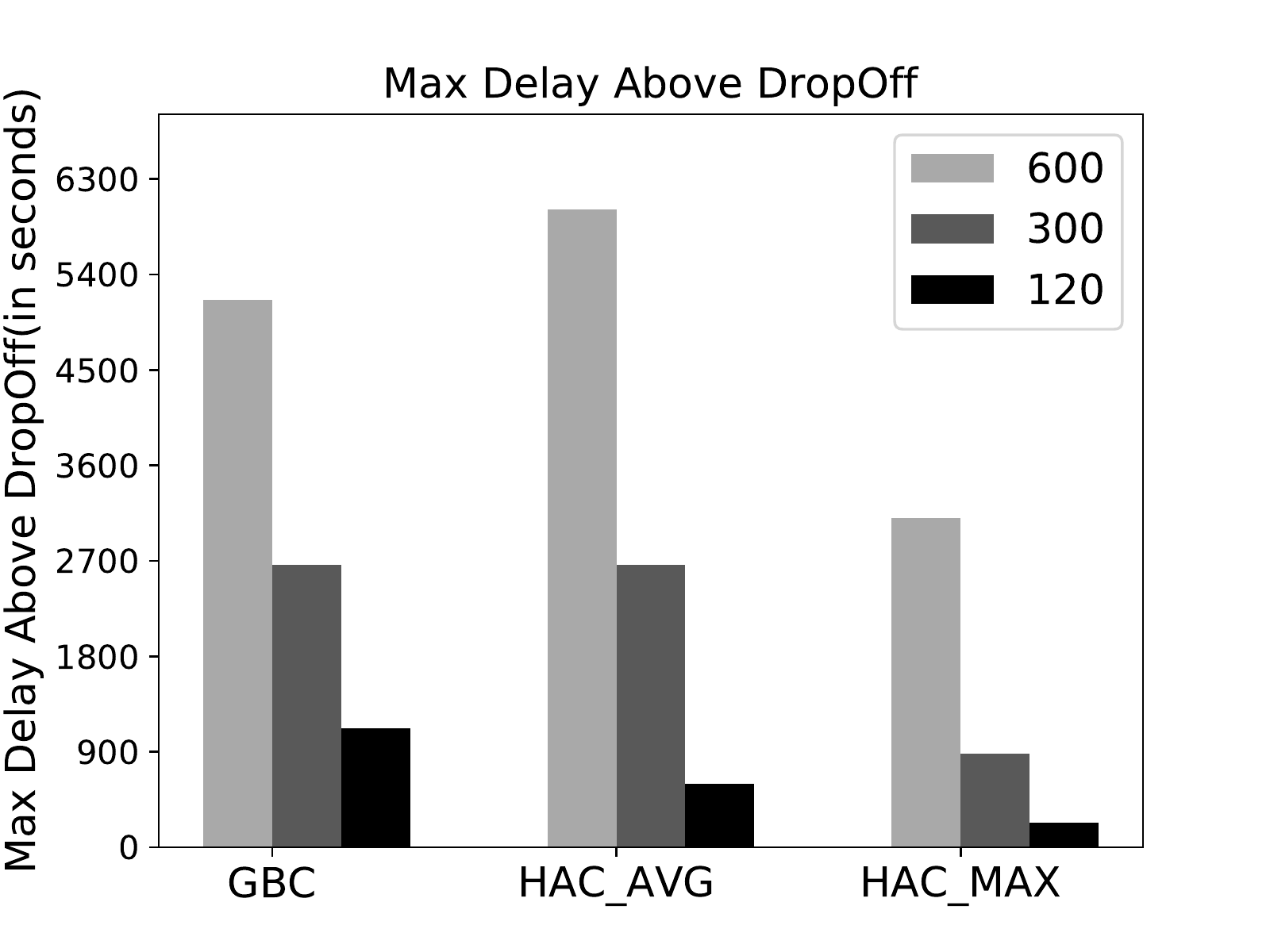}}
  \caption{{Comparison of service rate, runtime and abstraction error with different clustering methods and zone sizes for $M=1$, number of vehicles = 1000, capacity = 10, $\tau$ =300, $\lambda$ = 600 seconds}}
  \label{fig:figcluster}
\end{figure*}

\begin{figure*}[h]
  \centering
    \subfloat[]{\label{fig:figstaticdynamicservrate}\includegraphics[width=0.25\textwidth,height=1.4in]{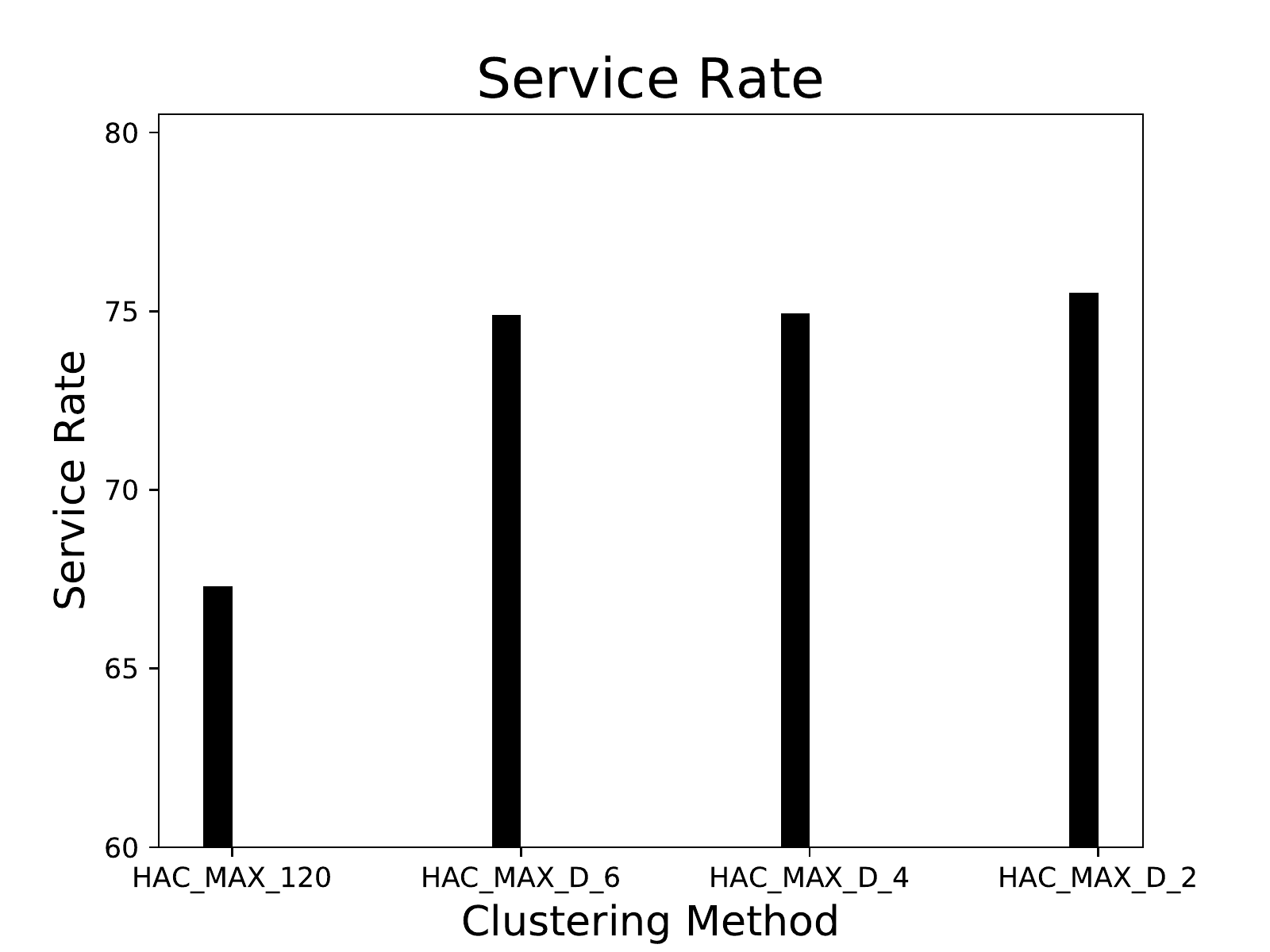}}
    \subfloat[]{\label{fig:figstaticdynamicrtime}\includegraphics[width=0.25\textwidth,height=1.4in]{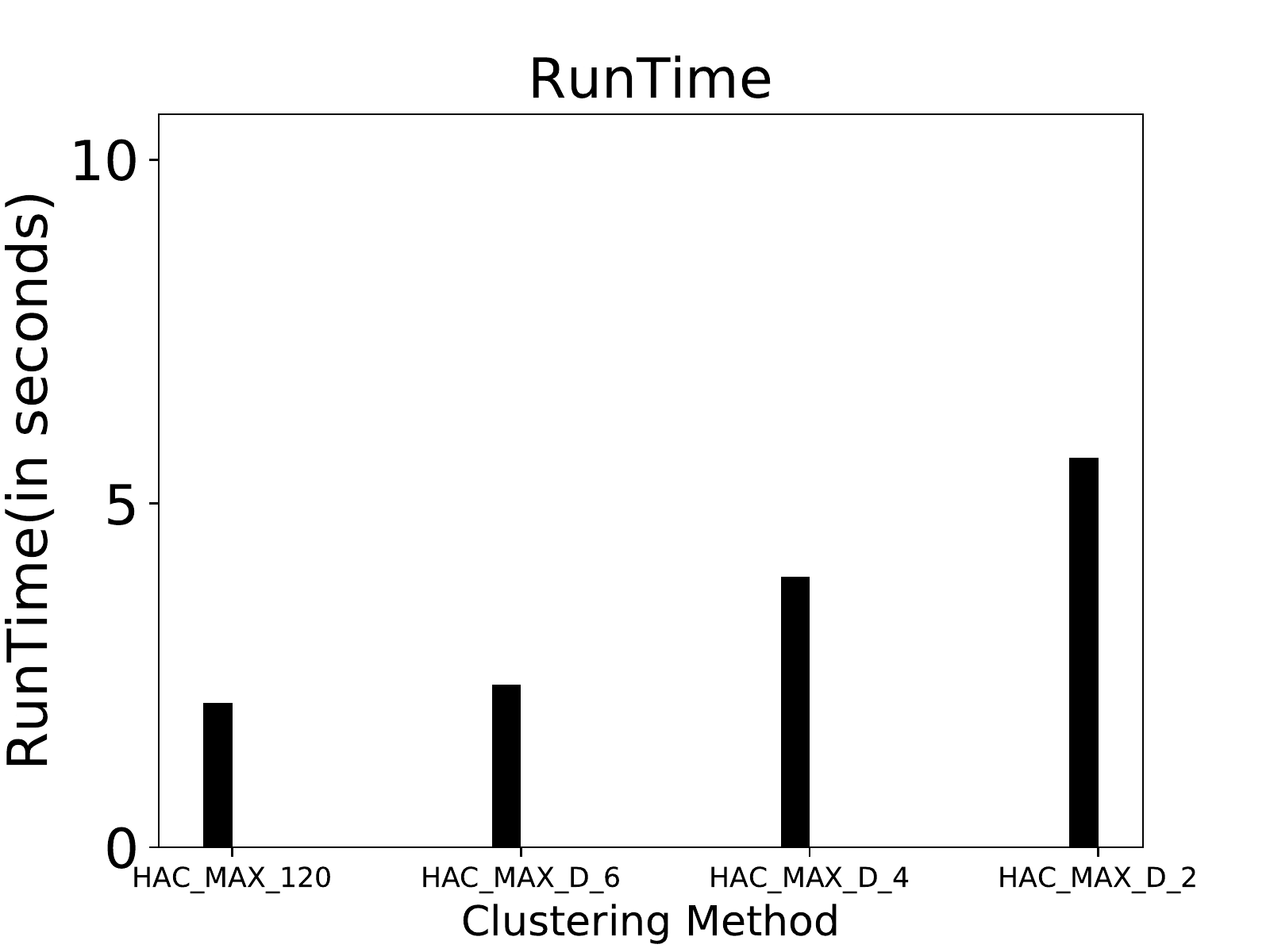}}
    \subfloat[]{\label{fig:figstaticdynamicpercent}\includegraphics[width=0.25\textwidth,height=1.4in]{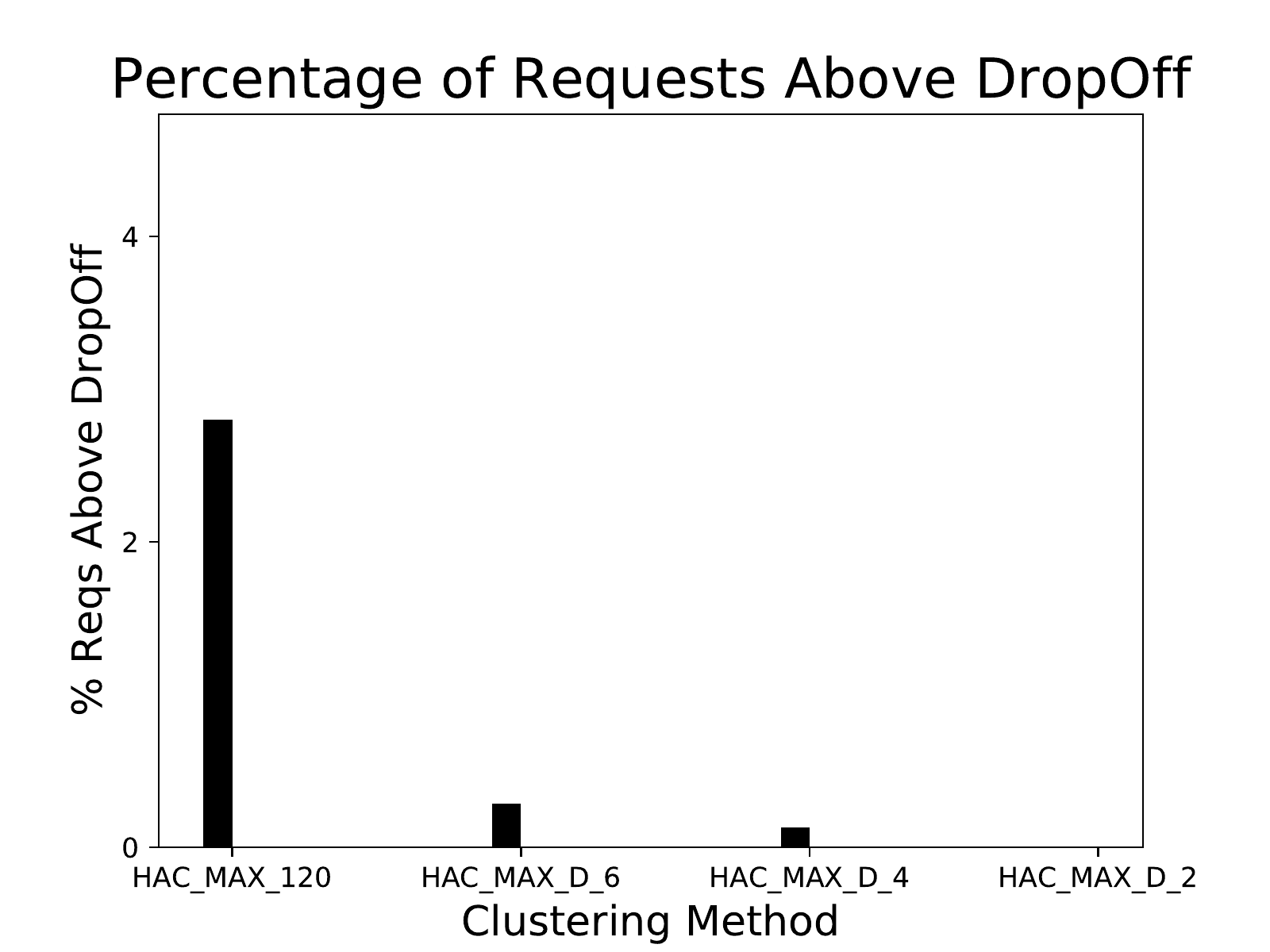}}
    \subfloat[]{\label{fig:figstaticdynamicmaxdrop}\includegraphics[width=0.25\textwidth,height=1.4in]{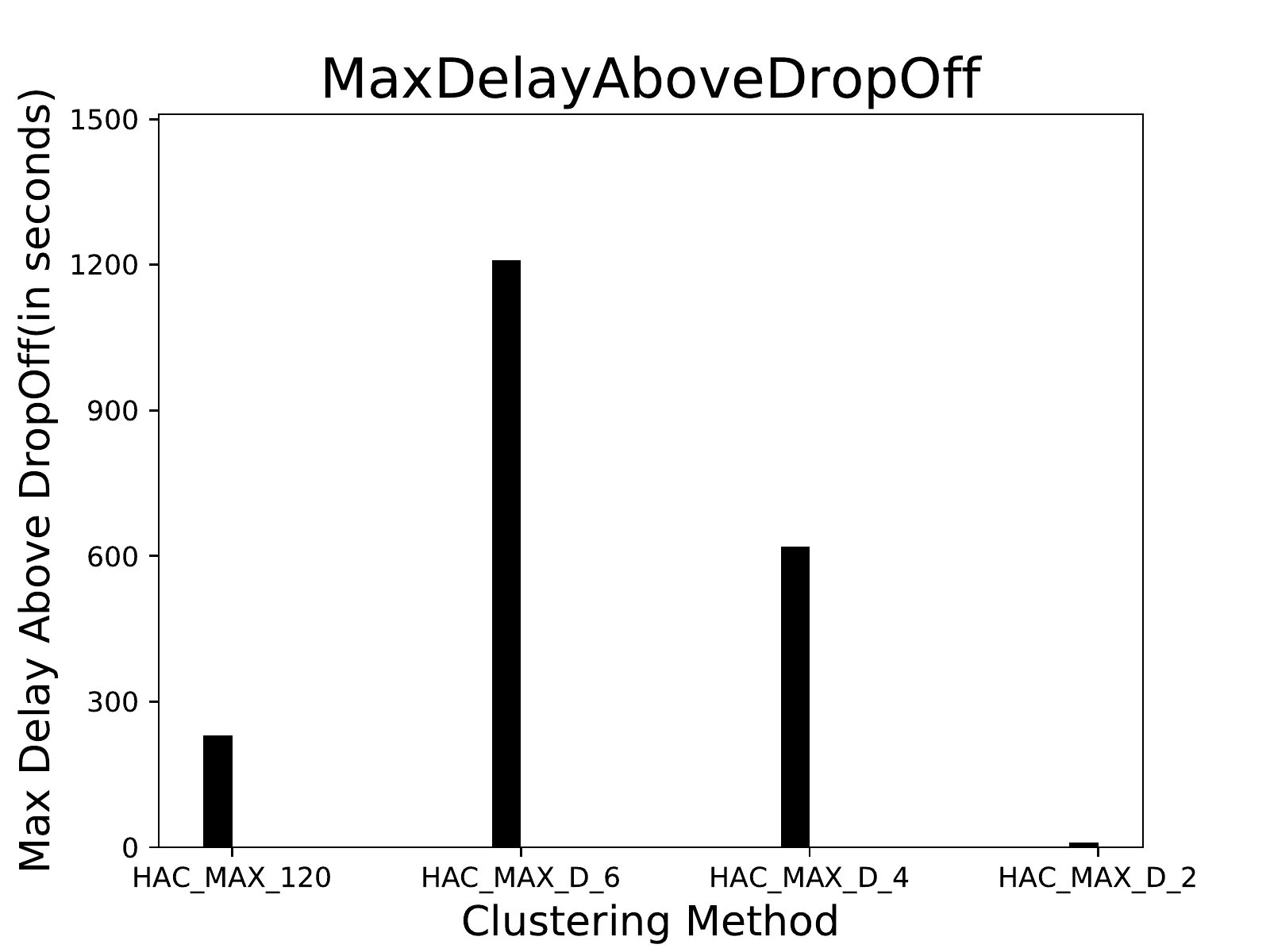}}
  \caption{{ Comparison of service rate, runtime and abstraction error with different values of $M$ and zone sizes for NYDataset, number of vehicles = 1000, capacity 10, $\tau$ = 300, $\lambda$ = 600 seconds}}
  \label{fig:figstaticdynamic}
\end{figure*}
\begin{figure*}[h]
  \centering
    \subfloat[]{\label{fig:figsamplesservrate}\includegraphics[width=0.45\textwidth,height=1.4in]{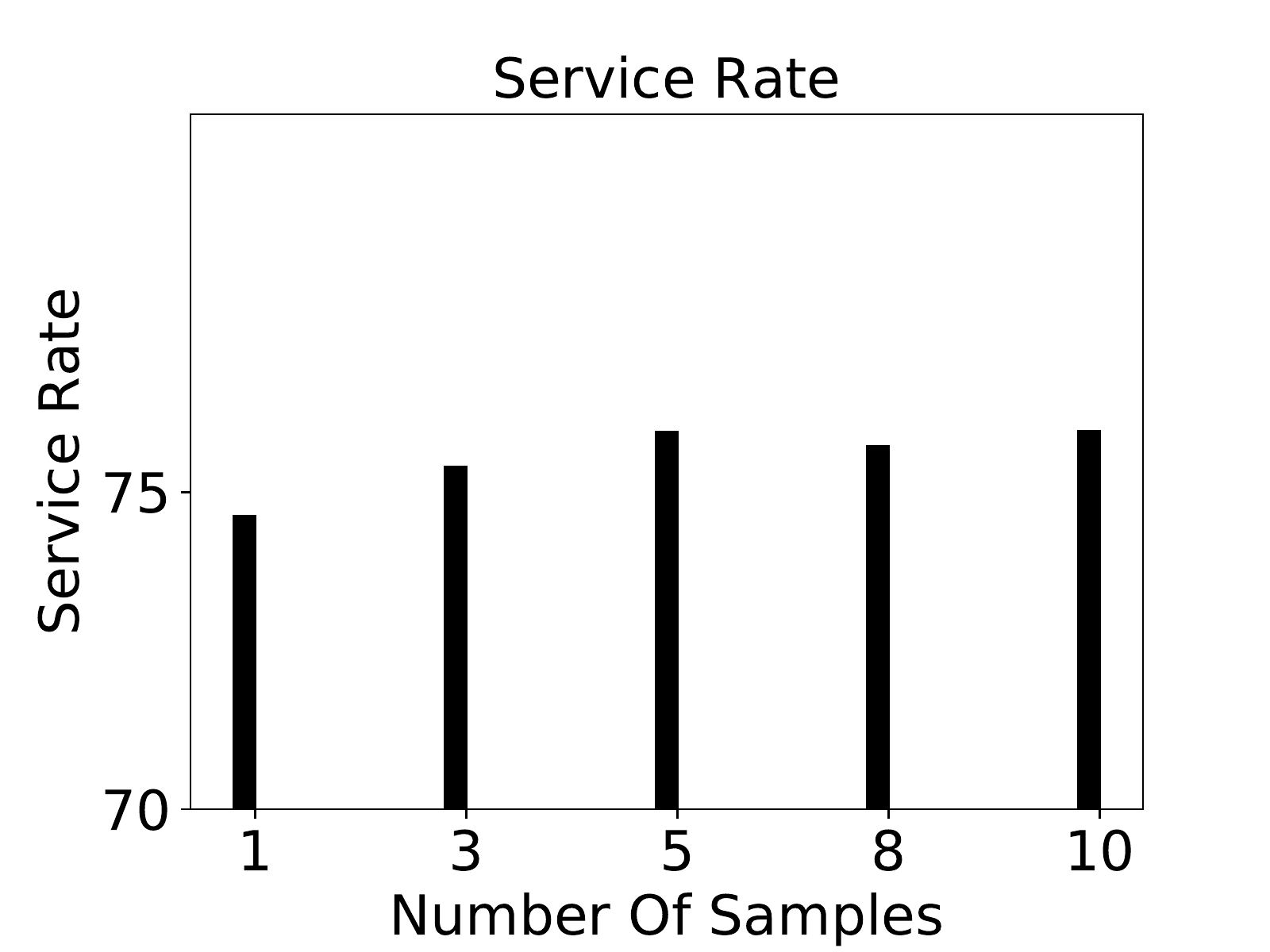}}
    \subfloat[]{\label{fig:figlookaheadservrate}\includegraphics[width=0.45\textwidth,height=1.4in]{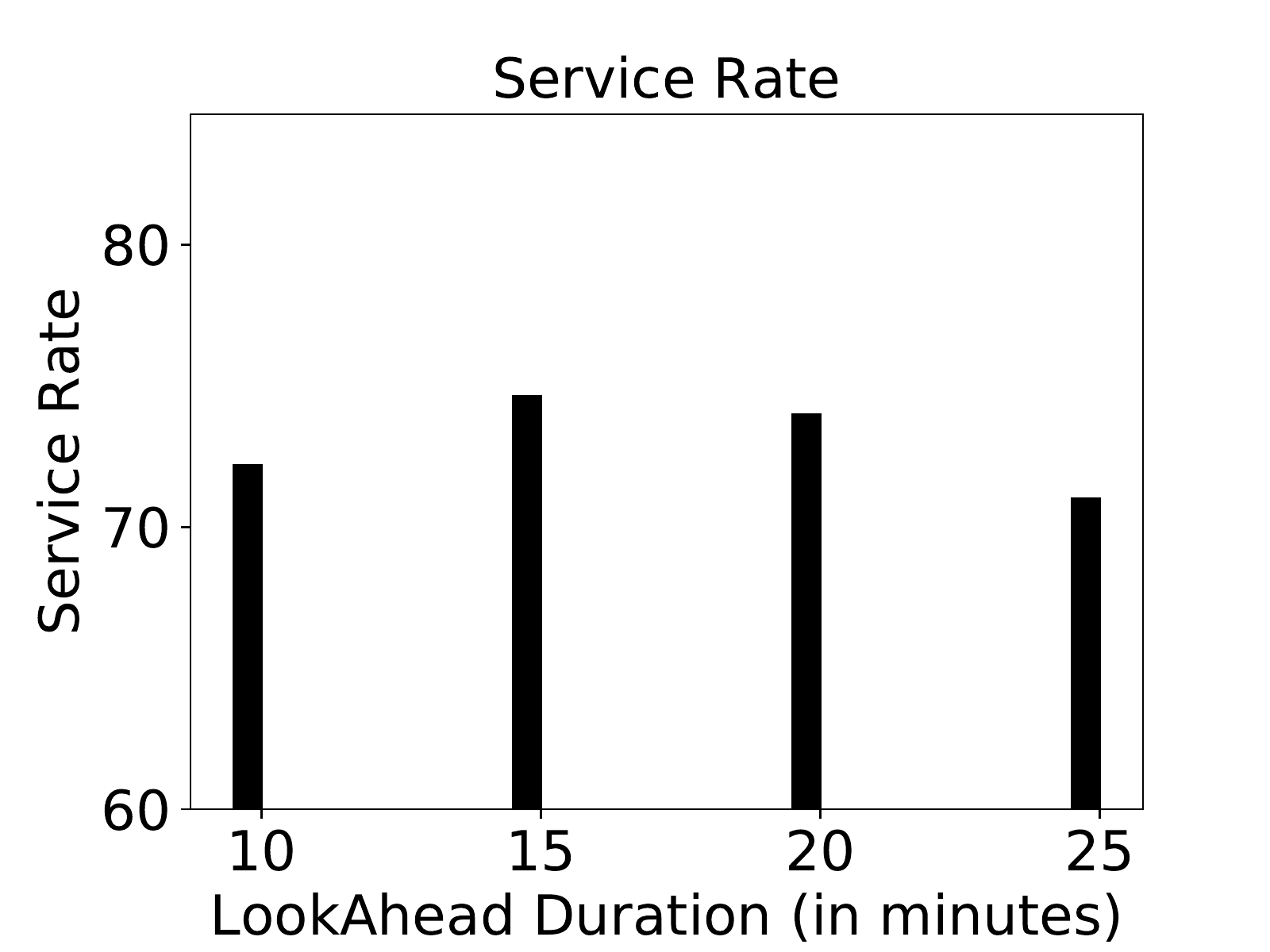}}
    %\subfloat[]{\label{fig:figlookaheadrtime}\includegraphics[width=0.45\textwidth,height=1.5in]{varyLookAhead_1000_4_rtime_ny.pdf}}\\
   % \subfloat[]{\label{fig:figsamplesrtime}\includegraphics[width=0.33\textwidth,height=1.5in]{varySamples_1000_4_rtime_ny.pdf}}
  \caption{{Comparison of service rate for different number of samples and lookahead duration number of vehicles = 1000, capacity = 4, $\tau$ =300, $\lambda$ = 600 seconds}}
  \label{fig:figsampleslookahead}
\end{figure*}

\subsubsection{Identification of Right Clustering Method}
We first conduct experiments by using different clustering methods, with $M=1$, by varying the zone sizes. Zone size is taken as the intra zone travel time (in seconds). Figure \ref{fig:figcluster} shows the comparison of GBC, HAC\_MAX and HAC\_AVG on NYDataset. We compare the service rate, runtime and abstraction error with different clustering methods and different zone sizes for ZAC. We measure abstraction error by computing the percentage of requests having delay above $\lambda$ and maximum delay obtained by any request that is above $\lambda$. We can observe that with HAC\_MAX more requests can be served while keeping the error due to abstraction minimal. We also observe that as the zone size decreases, the number of requests served increases, error due to abstraction decreases with a slight increase in runtime. Based on these results, we use HAC\_MAX as the clustering method for our next set of experiments.
%\subsection{Justification of 
\subsubsection{Identification of Right Value of $M$}
Our next set of experiments compare the service rate, runtime and abstraction error obtained using different values of $M$. Based on the observations made earlier, for $M=1$, we use HAC\_MAX with zone size 120. For $M>1$, the clustering method used is HAC\_MAX and we run the experiments with different values of $M$. We use the zone sizes as 0, 60, 120, 300, 480, 600. The zone size of 0 means that the actual locations in the street network are used. Zone size of 60 means that the intra zone travel time is 60 seconds and so on. For $M=2$, zone sizes used are 0 and 60, for $M=4$ zone sizes used are 0, 60, 120 and 300 and for $M=6$, zone sizes used are 0, 60, 120, 300, 480, 600.

We show the comparison of service rate and runtime with $M=1$ (with zone size 120) and different values of $M$ in Figure \ref{fig:figstaticdynamic}. {HAC\_MAX\_120} is used to denote that $M=1$ with zone size 120 is used. HAC\_MAX\_D\_$<m>$ denotes that value of $M$ used is $m$. From the Figure \ref{fig:figstaticdynamic} we can observe that we can serve more requests when $M > 1$, as compared to using fix large size zones. The abstraction error also reduces significantly by using $M > 1$. As the value of $M$ is reduced, quality of solution improves with the increase in runtime. With $M=2$ (for zone sizes 0 and 60) , the abstraction error is almost 0 but runtime also increases. With $M=4$, the abstraction error is less than 1\%.

%Please note that with $M=1$ and zone size as 0, we will get optimal result for the current decision epoch, but it is not possible to execute it due to computational complexity. The comparison with $M=1$ for zone size $>$ 0, shows that instead of using a fix zone size $>$ 0, using multiple zone sizes with 0 as one of the zone sizes can provide the right tradeoff between computational complexity and solution quality.

From these experiments on NYDataset, we obtain that by clustering locations into zones using HAC\_MAX and using $M$=4 (with zone sizes 0,60,120,300), we get the right trade-off between computational complexity and solution quality. Therefore, we use this configuration for ZAC and ZACBenders. We now identify the right number of samples and lookahead duration for the ZACBenders algorithm. 

\subsubsection{Number of Samples:} We compare the service rate and runtime by varying number of samples from 1 to 10 as shown in Figure \ref{fig:figsamplesservrate}. We can observe from the figure, the service rate obtained by using a single sample is 74.6\% and increases to 75.9\% on using 5 samples. The service rate obtained by using 10 samples is 75.96\% which is only 0.06\% more than the service rate obtained by 5 samples. As the improvement beyond 5 samples is not much and comes at the cost of extra average runtime, we use 5 samples for the ZACBenders algorithm.  

\subsubsection{LookAhead Duration:}
We compare the service rate by varying the look ahead duration from 10 minutes to 25 minutes for a single sample as shown in Figure \ref{fig:figlookaheadservrate}. We can observe from the figure that the service rate obtained by using look ahead duration of 10 minutes is 72.16\% and increases to 74.6\% on using look ahead of 15 minutes. The service rate obtained by using higher lookahead is less as with higher look ahead more requests are present at future decision epochs which increases the complexity of the problem and so the number of benders decomposition iterations which can be completed within the maximum time limit reduces affecting the performance of ZACBenders. Moreover, due to the approximations used in the ZACBenders algorithm, it is not necessary that the performance will improve on using higher look ahead duration. Based on these experimental results, we choose a look ahead of 15 minutes for the ZACBenders algorithm.  

\section{Related Work}
\label{sect:related}
Given the practical and environmental benefits of ridesharing systems, there has always been a lot of interest in developing algorithms for performing matching in these systems. The ridesharing problem is related to the Online Multi-Vehicle Pick-up and Delivery problems which typically represent problems where there are multi-capacity vehicles that transport multiple resources/loads from their origins to destinations. When vehicles are used to move people instead of resources, the problem is referred to as dial-a-ride problem~\cite{feuerstein2001line,lipmann2002line,bonifaci2006online} and when all the origins or all the destinations are located at a depot, the problem is referred to as vehicle routing problem~\cite{ritzinger2016survey}. The general representation of the dial-a-ride problems is ideally suited to represent problems faced by companies such as super shuttle (transports people from an airport to different locations in the city), uber pooling (transports customers from near by start locations to near by destination locations). But these problems are hard to solve and the traditional approaches for these problems can solve only very small instances of 96 requests and 8 vehicles~\cite{ropke2007models}.
 
The integer programming formulation, without any spatial or temporal aggregation~\cite{ropke2007models}, is difficult to solve and is not scalable to large scale problems and online decision making even for unit-capacity. Therefore, in recent times, many heuristic approaches have been proposed to solve the real-time taxi ridesharing problem. As shown in Figure \ref{fig:relwork}, the existing work on ridesharing systems can be categorized along three dimensions of capacity, consideration of requests and the nature of assignment (whether it is myopic or takes future demand into account for making current assignments). 

In case of unit-capacity ridesharing systems, vehicles need to be assigned to at most one request at a time. Greedy and randomized ranking~\cite{karp1990optimal} algorithms have also been used in the literature to compute myopic matching when requests are considered sequentially. The myopic matching for the batch case is also trivial in this case and can be achieved by performing a bipartite matching between vehicles and requests~\cite{agatz2011dynamic}. To improve the performance of these myopic algorithms in the batch case, there has been research on providing approximate dynamic programming ~\cite{simao2009approximate}, reinforcement learning~\cite{xu2018large,Lin:2018} and multi-stage stochastic optimization approaches which consider multiple samples of future demand to estimate the expected future value of current assignments. While we also consider the batch case and use stochastic optimization approaches, our work is different from this thread as we focus on high capacity ridesharing where computing a myopic assignment in itself is a challenging problem. To further include the future demand samples and solve the optimization in real-time, we need to propose multiple approximations. 

For multi-capacity ridesharing, due to the complexity of finding a myopic batch assignment, most of the existing works consider sequential (i.e., one by one) assignment. Widdows {\em et al.}~\cite{grabshare} and Tang {\em et al.}~\cite{grabshare1} propose an approach which allows 2 passengers to travel in the vehicle at the same time. It takes one request at a time and generates all feasible driver paths by inserting the pick-up and drop-off of the request in the existing driver paths. Pelzer {\em et al.}~\cite{pelzer2015partition} also allow 2 passengers to share the ride. They divide the road network into multiple partitions and limit the search space within the partition to find the match for the incoming request. Ma {\em et al.}~\cite{ma2013t} propose a myopic sequential matching algorithm for high capacity ridesharing. They propose a taxi searching algorithm which uses a spatio-temporal index to quickly retrieve candidate taxis. It then uses a scheduling algorithm which after comparing the current request with each candidate taxi, insert into the schedule of taxi which minimizes additional incurred distance for the request. Other works~\cite{huang2014large,tong2018unified,chen2018ptrider,cheng2017utility} also provide approaches where insertion operation is widely utilized, i.e., for each request, they find the best place to insert in a taxi's path. 

While the sequential solution is faster to compute, the quality of solution obtained is typically poor, therefore, there have been works on finding a myopic batch solution. Most of the works in this case have focussed on low capacity vehicles. Zheng {\em et al.}~\cite{zheng2018order} consider batch assignment but they only consider grouping at most two requests in a vehicle. They propose different approximation and apply matching and optimization based approaches to assign vehicles to the combination of two requests. Dutta~\cite{dutta2018hashing} use a locally sensitive hashing technique to efficient group two requests together but they do not consider assignment of vehicles to requests. 
Brown {\em et al.}~\cite{lyftline} propose exhaustively generating the combinations of at most three requests from all the available requests. For capacity two vehicles, Yu {\em et al.}~\cite{yu2019integrated} propose an approximate dynamic programming approach which is non-myopic. They use a linear value function approximation to approximate the future effect of assignment and use spatial and temporal aggregation to group different parts of road network into a small number of regions. Their approach is not scalable to large number of locations and higher capacity vehicles. 

A leading approach for high capacity ridesharing was provided by Alonso {\em et al.}~\cite{alonso2017demand}. This is a myopic batch assignment approach which as discussed in Section \ref{sect:intro} employs different heuristics for online execution. The existing non-myopic batch assignment approaches for high-capacity ridesharing~\cite{alonso2017predictive,neuradp} use the myopic approach by Alonso {\em et al.}~\cite{alonso2017demand} as a base approach. These approaches have drawbacks where either they can not be executed in real-time or can not be easily adapted to different settings and parameters. 

The non-myopic approach by Alonso {\em et al.}~\cite{alonso2017predictive} is a minor extension of their myopic approach where they randomly sample 200 or 400 requests for next 30 minutes and then use those requests along with currently available requests to generate the assignments. Typically, there are 300 requests per minute so randomly sampling 200 or 400 requests for next 30 minutes does not help in improving the quality of solution. This is reflected in their results as well, where the service rate remains approximately same as the service rate of the myopic approach. But they observe a minor decrease in the average delay experienced by the passengers. The sampled requests also increase the computational complexity of the approach and the runtime of the approach after adding sampled requests is more than the time available for assignment(duration over which requests are batched). As a result, it is not possible to use the approach for real-time assignments. The non-myopic approach by Shah {\em et al.}~\cite{neuradp} uses similar approach as Alonso {\em et al.}~\cite{alonso2017demand} to generate the feasible trips and then use a neural network based value function approximation to estimate the future effect of an current assignment of vehicle to trips. While the approach greatly outperforms the myopic approaches, due to the need of training a separate network model for each dataset and each change of input parameter, it is not easily adaptable to different settings. 

The zone path construction based approaches proposed in this work, overcome these limitations of existing work by providing an offline-online method to generate request combinations efficiently by employing zone-paths. The future value of assignment to these zone paths is computed by considering multiple samples of future demand.  

\section{Conclusion}
In this paper, we presented zone path construction based approaches that can efficiently perform ridesharing for higher capacity vehicles. The experimental comparison on real-world and synthetic datasets show that our approach can outperform the current best myopic approach (used even by taxi and car aggregation companies like Grab and Lyft) in terms of both runtime and solution quality. Our non-myopic approach further improves the performance by using multiple future demand samples and outperforms the state of the art approaches. 

\nocite{chen2018ptrider,Lin:2018,pelzer2015partition,cheng2017utility,hasan2018community,lowalekar2018online}

\section{Acknowledgements}
This work was partially supported by the Singapore National Research Foundation through the Singapore-MIT Alliance for Research and Technology (SMART) Centre for Future Urban Mobility (FM). We thank Sanket Shah for providing valuable comments that greatly improved the paper.  

\appendix
\section{Zone Creation}
\label{appendix:zonecreation}
\begin{figure}[htbp]
 \centering
\subfloat[]{\label{fig:figz1}\includegraphics[width=0.3\textwidth,height=2.3in]{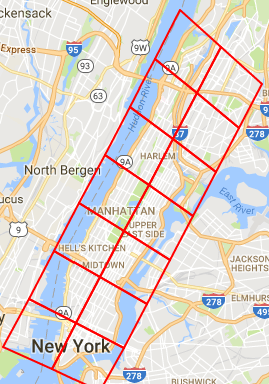}}
\subfloat[]{\label{fig:figz2}\includegraphics[width=0.3\textwidth,height=2.3in]{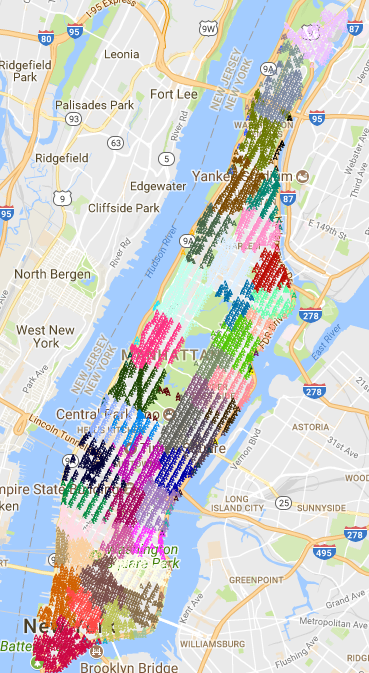}}
 \vspace{-0.1in}
 \caption{Abstraction of Locations into Zones (a) Grid based clustering (b) Clusters formed by Hierarchical Clustering}
 \label{fig:zones}
 \vspace{-0.2in}
\end{figure}

We cluster the set of locations ${\cal L}$ into zones. In this work, we mainly explored the following methods to cluster locations into zones:
\begin{enumerate}
\item \textit{Grid Based Clustering (GBC):} As shown in Figure \ref{fig:figz1}, the city can be divided into different parts using square grid cells. Each square grid represents a zone. The size of square grid can be used to determine the time taken to move from one zone to another zone or within a zone~\footnote{Presence of one way streets can make the computation of actual time taken a bit challenging.}. It provides a simple method to divide city of any size into multiple zones. A very similar grid based clustering is used by Ma {\em et.al.} \cite{ma2013t}.
\item \textit{Hierarchical Agglomerative Clustering (HAC):} We also employed Hierarchical Agglomerative clustering~\cite{rocach2005clustering} to cluster locations into zones (Figure \ref{fig:figz2}). The algorithm starts by including each location in a different cluster/zone and the distance between two locations is measured in terms of the time taken to travel between them. At each iteration of the algorithm, the two closest clusters are merged to form a single cluster. The process continues until the minimum inter-cluster travel time between any two clusters is more than a given threshold. We define the inter cluster travel time using the following two linkage criterion:\\
\noindent (1) \textit{Complete Linkage (HAC\_MAX): }The time taken to travel between two clusters is the maximum time required to travel between two locations of different clusters, i.e.,
{
$$T(X,Y) = \max_{x \in X, y\in Y} t(x,y)$$
}
where $X$ and $Y$ denote the two clusters and $x$ and $y$ denote the locations. $t(x,y)$ denotes the time taken to travel from location $x$ to $y$ and $T(X,Y)$ denotes the time taken to travel from cluster $X$ to cluster $Y$. Complete linkage tends to find compact clusters of approximately equal diameters.\\
\noindent (2) \textit{Mean Linkage (HAC\_AVG):} The time taken to travel between two clusters is the average of the time required to travel between any two locations of different clusters, i.e.,
{
$$T(X,Y) = \frac{1}{|X||Y|}\sum_{x \in X,}\sum_{y\in Y} t(x,y)$$
}
\end{enumerate}

As mentioned before, we use these methods as they do not require prior knowledge about the number of clusters and have been used in earlier works on similar problems~\cite{ma2013t,hasan2018community}. We perform experiments comparing these different zone creation methods and use the method which provides right trade-off between computational complexity and solution quality. (Please refer to Section ~\ref{sect:justifyparam} for detailed results).

\section{Offline Partial Path Generation: Data driven Approach}
\label{appendix:offline}
While for smaller networks and for smaller values of $\tau$ ($\leq$ 180 seconds) for larger networks, we can generate and store all possible paths, but due to the exponential increase in the number of paths with increasing value of $\tau$ for large networks, it is necessary to limit the number of paths. One possible way to reduce the number of paths is by using zones instead of locations in the offline path generation. However, this can lead to additional delay during pick-up, which is not preferred. Therefore, we use the following data driven approach to reduce the number of paths for larger $\tau$ values.

The key idea in the data driven approach is to only keep the paths that have a high likelihood of serving large number of requests with the assumption that future requests will follow the historical demand pattern. 

When the value of $\tau$ is large, we break $\tau$ into multiple smaller time intervals $\tau_1, \tau_2, ...\tau_{n}$, such that $\tau = \sum_{i} \tau_i$. The value $\tau_{i}, \forall i$ is taken such that we can generate and store all possible paths of $\tau_{i}$ duration. 

We process all possible paths for $\tau_{1}$, $\tau_{2}$,..,$\tau_{n}$ duration against the requests available in historical data. We ignore the paths for which the average number of requests served per minute is less than a threshold $\gamma_{i}$ (learned experimentally) and then combine only those paths for which number of requests served is greater than $\gamma_{i}$. This process ensures that we keep the paths which are more likely to serve high number of requests. 

By following the above process, there is a chance that no path gets recorded between two locations that can be reached in $\tau$ seconds. For such location pairs, we keep the shortest path to travel between them. 

As an example, in our experiments, for $\tau=300$ seconds, we consider $\tau = \tau_{1} + \tau_{2}$, where $\tau_{1}$ = 120 seconds and $\tau_{2}$ = 180 seconds and we take $\gamma_{1}$ and $\gamma_{2}$ as 1. The paths of $\tau_{1}$ and $\tau_{2}$ duration are processed against 20 days of historical data.

\section{Pseudocode for GetPathsForVehicle function used in ProcessOfflinePartialPaths}
In this section, we describe the pseudocode for GetPathsForVehicle function used in the Algorithm \ref{alg:processoffline} in main paper. 
In step \ref{step:v66}, we store the destination location of the requests previously assigned to vehicles. A vehicle should deviate from its current path only if it can be assigned to a new request, therefore, in step \ref{step:v6}, we consider only those paths which can pick at least one of the newly available requests. 
Steps \ref{step:v9}-\ref{step:v10} ensure that we consider only those paths which can potentially satisfy all the previously assigned requests for a vehicle. This is because a vehicle will be assigned to a path if and only if it can serve all previously assigned requests. 
\begin{algorithm}
\caption{GetPathsForVehicle($i,q_i,{\cal R}[k],{\cal R}_{p}[k],{\cal P}_{off})$}
{\small
\begin{algorithmic}[1]
\STATE{plist=\{\}}
\STATE{${\cal R}^{i}$=[],${\cal R}^{i}_{p}$=[]}
\FOR{ $r \in q_{i}$}
\IF{$r$ is already picked}
\STATE{ ${\cal P}_{off}^{h,r} = GetPathsFromIndex({\cal P}_{off}^{h},\nu_{i},\omega_{i},\omega_{i})$}
\ELSE
\STATE{ ${\cal P}_{off}^{h,r} = GetPathsFromIndex({\cal P}_{off}^{h},o_{r},a_{r}-t,a_{r}-t+\tau)$}
\ENDIF
\FOR{each path $k \in {\cal P}_{off}^{h,r}$}
\STATE{plist[k]+=1}
\IF{$|{\cal R}[k]| > 0$}\label{step:v6}
\STATE{$lb_{j} = a_{j}-t+{\cal T}(o_{j},d_{j}), ub_{j} = lb_{j} + \lambda$}
\STATE{${\cal R}^{i}[k].add(d_{j},lb_{j},ub_{j})$}\label{step:v66}
\STATE{${\cal R}^{i}_{p}[k].add(o_{j})$}
\IF{$lb_{j} < \tau$} \label{step:v7}
\IF{$k$ visits $d_{j}$}
\STATE{${\cal R}_{p}[k].add(d_{j})$}
\ELSIF{$ub_{j} < \tau$}
\STATE{${\cal R}[k].remove(d_{j},(lb_{j},ub_{j}))$}\label{step:v8}
\ENDIF
\ENDIF
\ENDIF
\ENDFOR 
\ENDFOR
\FOR{each path $k \in $ plist}
\IF{$|plist[k]| == |q_{i}|$} \label{step:v9}
\STATE{${\cal R}[k].addAll({\cal R}^{i}[k])$}
\STATE{${\cal R}_{p}[k].addAll({\cal R}^{i}_{p}[k])$}\label{step:v10}
\ENDIF
%\ENDIF
\ENDFOR
\RETURN{${\cal R}[k],{\cal R}_{p}[k]$}
\end{algorithmic}
\label{alg:processofflinevehicle}}
\end{algorithm}

%\section{Example showing different steps of ZAC}
%Figure \ref{fig:example} shows all the steps of ZAC. In the first image, we show one of the offline partial path. There is a vehicle present at the location marked as Start. This offline partial path is processed against incoming requests and the pickup locations of the incoming requests is marked with green marker in the next image. In the next, step we consider the drop-off locations of all the requests which have pickup along the path. These drop-off locations are grouped together in zones/Clusters by using appropriate zone size. The offline partial path is then completed by starting exhaustive search at the last green node. As shown in the figure, the online completion of one offline partial path results in providing multiple zone paths. In the last figure, using arrows, we highlight one of the zone path.

\section{Re-balancing of unassigned vehicles}
Similar to Alonso {\em et al.}~\cite{alonso2017demand}, we perform a re-balancing of unassigned vehicles to high demand areas by using the same method as them. We make similar assumptions 
\begin{enumerate}
\item Unserved customers may request again.
\item At future timesteps, more customer requests may originate from the areas where we could not serve requests at current timestep.  
\end{enumerate}

%\begin{figure}
% \centering
%\includegraphics[width=0.95\textwidth,height=3.0in]{images/example.pdf}
 %\vspace{-0.1in}
 %\caption{Example showing different steps of ZAC. (a) One of the offline partial paths (b) Online processing of Offline Partial Path - green markers represent the pick up location of the requests grouped along the path. (c) Online Partial Path Completion - The blue markers represent the dropoff locations of the requests which are grouped along the path based on their pickup location. (d) Online Partial Path Completion - The drop off locations are grouped together based on the different zones (e) 
%Online Path Completion - The path is completed by starting exhaustive search at the last green node. The exhaustive search generates these different paths shown in different colors. (f) The path represented by arrows shows one of the complete zone path.  
%}
% \label{fig:example}
% \vspace{-0.2in}
%\end{figure} 
Therefore, to rebalance the unassigned vehicles, after each batch assignment, the unassigned vehicles are assigned to unserved requests to minimize the sum of travel times, with the constraint that either all unserved requests or all of the unassigned vehicles are assigned. The linear program is provided in table \ref{table:rebal_opt}. Let ${\cal V}_{u}$ denotes the set of unassigned vehicles and ${\cal D}_u$ denote the set of unserved customer requests. $m_{ij}$ is a binary variable indicating that vehicle $i$ is moving towards customer request $j$.

\begin{table}[htbp]
\center
    \begin{tabular}{|r|}
    \hline

    \begin{minipage}{0.45\textwidth}
        \vspace{0.05in}
\textbf{RebalanceVehicles():}
{\small
\begingroup
\addtolength{\jot}{-2pt}
\begin{align}
\min \quad & \sum_{j \in {\cal D}_u} \sum_{i \in {\cal V}_{u}} {\cal T}(\nu_{i},o_{j})*m_{ij} \\
subject ~ to \quad & \sum_{i \in {\cal V}_{u}} m_{ij} \leq 1 ::: \forall j \in {\cal D}_u \label{cons:r2}\\
& \sum_{j \in {\cal D}_u} x_{ij} \leq 1 ::: \forall i \in {\cal V}_u \label{cons:r3}\\
& \sum_{j \in {\cal D}_u}\sum_{i \in {\cal V}_u}  m_{ij} = min(|{\cal V}_u|,|{\cal D}_u|) \label{cons:r4}\\
&0 \leq  m_{ij} \leq 1 ::: \forall i,j
\end{align}
\vspace{-8pt}
\endgroup }
\end{minipage} \\
    \hline
    \end{tabular}
    \vspace{-0.1in}
    \caption{Optimization Formulation for Rebalancing unassigned vehicles}
    \label{table:rebal_opt}
    \vskip -10pt
    \end{table}

\section{Complexity Analysis for Offline-Online Generation of Zone Paths}
We analyse the computational complexity of the Offline-Online Generation of Zone Paths. We look at the complexity of each step.
\begin{enumerate}
\item \textbf{Offline Partial Path Generation:} This step requires generating all paths in the network of ${\cal G}$ duration $\tau$. The network ${\cal G}$ has ${\cal L}$ nodes and ${\cal E}$ edges. As the graph is not fully connected, we assume a branching factor of $b$ (average number of neighbors of any vertex $l \in {\cal L}$). Let the maximum length of a duration $\tau$ path is $k$. Therefore, the complexity of this step is the complexity of generating all paths of length $k$ in a network with ${\cal L}$ nodes, $E$ edges and branching factor $b$. 
The total possible number of paths is $|{\cal L}|*b^k$. The total nodes explored in path generation are $|V|*(1+b^1+b^2+...+b^k)$ = $|V|*\frac{b^{k+1} -1}{b-1}$. 
\item \textbf{Online Processing of Offline Partial Paths:} The offline partial paths are processed using the current demand ${\cal D}$. Suppose $\alpha$ denotes the fraction of paths in which the pick-up location of any demand element is present. Then, the total complexity of online processing of offline partial paths is $O(|{\cal D}|*\alpha*|{\cal L}|*b^k)$.
\item \textbf{Online Completion of offline partial paths:} Let the total number of unique partial paths after online processing step is $\omega*|{\cal L}|*b^{k}$. Now suppose each path is associated with maximum $\beta$ drop-off locations after considering the appropriate zone size out of the $M$ available zone sizes. As we need to perform exhaustive search on these $\beta$ locations, the complexity of this step is $O((\beta!)*\omega*|{\cal L}|*b^k)$. 
\end{enumerate}

As the network ${\cal G}$ is generally very sparse and each of the individual steps can be parallelized, the actual computation time is much less allowing to execute this in real-time.  
\bibliography{TaxiRef}
\bibliographystyle{theapa}

\end{document}